\documentclass[10pt,twocolumn]{IEEEtran}
\usepackage{graphicx}
\usepackage{epstopdf}
\usepackage{amssymb}
\usepackage{amsmath}
\usepackage{subfigure}
\usepackage{color}
\usepackage{comment}
\usepackage{widetext}
\usepackage{enumitem}
\usepackage{wrapfig}
\usepackage{algpseudocode}
\usepackage{algorithm}

\usepackage{epstopdf}
\usepackage{fancyhdr}
\usepackage{bbm}
\newtheorem{definition}{Definition}

\newtheorem{corollary}{Corollary}

\begin{document}
\title{Understanding Convolutional Neural Networks with Information Theory: An Initial Exploration}
%

\author{Shujian~Yu,~\IEEEmembership{Student Member,~IEEE,} Kristoffer Wickstr{\o}m,\\ Robert Jenssen,~\IEEEmembership{Member,~IEEE,} and Jos\'{e} C. Pr\'{i}ncipe,~\IEEEmembership{Life Fellow,~IEEE.}%
\thanks{Shujian Yu and Jos\'{e} C. Pr\'{i}ncipe are with the Department of Electrical and Computer Engineering, University of Florida, Gainesville, FL 32611,
  USA. (email: yusjlcy9011@ufl.edu; principe@cnel.ufl.edu)}%
\thanks{Kristoffer Wickstr{\o}m and Robert Jenssen are with the Department of Physics and Technology at UiT - The Arctic University of Norway, Troms{\o} 9037, Norway.
(email: \{kwi030,~robert.jenssen\}@uit.no)}}%
\maketitle

\begin{abstract}
A novel functional estimator for R{\'e}nyi's $\alpha$-entropy and its multivariate extension was recently proposed in terms of of the normalized eigenspectrum of a Hermitian matrix of the projected data in a reproducing kernel Hilbert space (RKHS). However, the utility and possible applications of these new estimators are rather new and mostly unknown to practitioners. In this brief, we first show that our estimators enable straightforward measurement of information flow in realistic convolutional neural networks (CNNs) without any approximation. Then, we introduce the partial information decomposition (PID) framework and develop three quantities to analyze the synergy and redundancy in convolutional layer representations. Our results validate two fundamental data processing inequalities and reveal some fundamental properties concerning CNN training.
\end{abstract}

%
\begin{IEEEkeywords}
Convolutional Neural Networks, Data Processing Inequality, Multivariate Matrix-based R{\'e}nyi's $\alpha$-entropy, Partial Information Decomposition.
\end{IEEEkeywords}

\graphicspath{{figures/}}
\section{Introduction}


There has been a growing interest in understanding deep neural networks (DNNs) mapping and training using information theory~\cite{tishby2015deep,achille2017emergence,tax2017partial}. According to Schwartz-Ziv and Tishby~\cite{shwartz2017opening}, a DNN should be analyzed by measuring the information quantities that each layer's representation $T$ preserves about the input signal $X$ with respect to the desired signal $Y$ (i.e., $\mathbf{I}(X;T)$ with respect to $\mathbf{I}(T;Y)$, where $\mathbf{I}$ denotes mutual information), which has been called the Information Plane (IP). Moreover, they also empirically show that the common stochastic gradient descent (SGD) optimization undergoes two separate phases in the IP: an early ``fitting" phase, in which both $\mathbf{I}(X;T)$ and $\mathbf{I}(T;Y)$ increase rapidly along with the iterations, and a later ``compression" phase, in which there is a reversal such that $\mathbf{I}(X;T)$ and $\mathbf{I}(T;Y)$ continually decrease. 
However, the observations so far have been constrained to a simple multilayer perceptron (MLP) on toy data, which were later questioned by some counter-examples in~\cite{michael2018on}. 


In our most recent work~\cite{yu2018understanding}, we use a novel matrix-based R{\'e}nyi's $\alpha$-entropy functional estimator~\cite{giraldo2015measures} to analyze the information flow in stacked autoencoders (SAEs). We observed that the existence of a ``compression" phase associated with $\mathbf{I}(X;T)$ and $\mathbf{I}(T;Y)$ in the IP is predicated to the proper dimension of the bottleneck layer size $S$ of SAEs: if $S$ is larger than the intrinsic dimensionality $d$~\cite{camastra2016intrinsic} of training data, the mutual information values start to increase up to a point and then go back approaching the bisector of the IP; if $S$ is smaller than $d$, the mutual information values increase consistently up to a point, and never go back. 


Despite the great potential of earlier works~\cite{shwartz2017opening,michael2018on,yu2018understanding}, there are several open questions when it comes to the applications of information theoretic concepts to convolutional neural networks (CNNs). These include but are not limited to:
\begin{enumerate}[fullwidth,itemindent=2em]
	\item The accurate and tractable estimation of information quantities in CNNs. Specifically, in the convolutional layer, the input signal $X$ is represented by multiple feature maps, as opposed to a single vector in the fully connected layers. Therefore, the quantity we really need to measure is the \emph{multivariate mutual information} (MMI) between a single variable (e.g., $X$) and a group of variables (e.g., different feature maps). Unfortunately, the reliable estimation of MMI is widely acknowledged as an intractable or infeasible task in machine learning and information theory communities~\cite{brown2012conditional}, especially when each variable is in a high-dimensional space.
	\item A systematic framework to analyze CNN layer representations. By interpreting a feedforward DNN as a Markov chain, the existence of data processing inequality (DPI) is a general consensus~\cite{shwartz2017opening,yu2018understanding}. However, it is necessary to identify more inner properties on CNN layer representations using a principled framework, beyond DPI.
\end{enumerate}

In this brief, we answer these two questions and make the following contributions:
\begin{enumerate}[fullwidth,itemindent=2em]
\item By defining a multivariate extension of the matrix-based R{\'e}nyi's $\alpha$-entropy functional~\cite{yu2018multivariate}, we show that the information flow, especially the MMI, in CNNs can be measured without knowledge of the probability density function (PDF).
\item By capitalizing on the partial information decomposition (PID) framework~\cite{williams2010nonnegative} and on our sample based estimator for MMI, we develop three quantities that bypass the need to estimate the synergy and redundancy amongst different feature maps in convolutional layers. Our result sheds light on the determination of network depth (number of layers) and width (size of each layer). It also gives insights on network pruning.
\end{enumerate}


\section{Information Quantity Estimation in CNNs} \label{sec2}

In this section we give a brief introduction to the recently proposed matrix-based R{\'e}nyi's $\alpha$-entropy functional estimator~\cite{giraldo2015measures} and its multivariate extension~\cite{yu2018multivariate}. Benefiting from the novel definition, we present a simple method to measure MMI between any pairwise layer representations in CNNs.


\subsection{Matrix-based R{\'e}nyi's $\alpha$-entropy functional and its multivariate extension}

In information theory, a natural extension of the well-known Shannon's entropy is R{\'e}nyi's $\alpha$-order entropy~\cite{renyi1961measures}. For a random variable $X$ with probability density function (PDF) $f(x)$ in a finite set $\mathcal{X}$, the $\alpha$-entropy $\mathbf{H}_\alpha(X)$ is defined as:
\begin{equation}
\mathbf{H}_{\alpha}(f)=\frac{1}{1-\alpha}\log\int_\mathcal{X}f^\alpha(x)dx.\label{eq3}
\end{equation}

R{\'e}nyi's entropy functional evidences a long track record of usefulness in machine learning and its applications~\cite{principe2010information}. Unfortunately, the accurate PDF estimation impedes its more widespread adoption in data driven science. To solve this problem, \cite{giraldo2015measures,yu2018multivariate} suggest similar quantities that resembles quantum R{\'e}nyi's entropy~\cite{muller2013quantum} in terms of the normalized eigenspectrum of the Hermitian matrix of the projected data in RKHS. The new estimators avoid evaluating the underlying probability distributions, and estimate information quantities directly from data. For brevity, we directly give the following definitions. The theoretical foundations for \emph{Definition~1} and \emph{Definition~2} are proved respectively in~\cite{giraldo2015measures} and~\cite{yu2018multivariate}.

\begin{definition} Let $\kappa:\mathcal{X}\times\mathcal{X}\mapsto\mathbb{R}$ be a real valued positive definite kernel that is also infinitely divisible~\cite{bhatia2006infinitely}. Given $X=\{x^1,x^2,...,x^n\}$ and the Gram matrix $K$ obtained from evaluating a positive definite kernel $\kappa$ on all pairs of exemplars, that is $(K)_{ij}=\kappa(x^i,x^j)$, a matrix-based analogue to R{\'e}nyi's $\alpha$-entropy for a normalized positive definite (NPD) matrix $A$ of size $n\times n$,  such that $\mathrm{tr}(A)=1$, can be given by the following functional:
\begin{equation}
\mathbf{S}_\alpha(A)=\frac{1}{1-\alpha}\log_2\left(\mathrm{tr}(A^\alpha)\right)=
\frac{1}{1-\alpha}\log_2\big[\sum_{i=1}^n\lambda_i(A)^\alpha\big], \label{eq5}
\end{equation}
where $A_{ij}=\frac{1}{n}\frac{K_{ij}}{\sqrt{K_{ii}K_{jj}}}$ and $\lambda_i(A)$ denotes the $i$-th eigenvalue of $A$.
\end{definition}

\begin{definition}
Given a collection of $n$ samples $\{s_i=(x_1^i,x_2^i,\cdots, x_C^i)\}_{i=1}^n$, where the superscript $i$ denotes the sample index, each sample contains $C$ ($C\geq2$) measurements $x_1\in \mathcal{X}_1$, $x_2\in \mathcal{X}_2$, $\cdots$, $x_C\in \mathcal{X}_C$ obtained from the same realization, and the positive definite kernels $\kappa_1:\mathcal{X}_1\times \mathcal{X}_1\mapsto\mathbb{R}$, $\kappa_2:\mathcal{X}_2\times \mathcal{X}_2\mapsto\mathbb{R}$, $\cdots$, $\kappa_C:\mathcal{X}_C\times \mathcal{X}_C\mapsto\mathbb{R}$, a matrix-based analogue to R{\'e}nyi's $\alpha$-order joint-entropy among $C$ variables can be defined as:
\begin{equation}
\mathbf{S}_\alpha(A_1,A_2,\cdots,A_C)=\mathbf{S}_\alpha\left(\frac{A_1\circ A_2\circ\cdots\circ A_C}{\mathrm{tr}(A_1\circ A_2\circ\cdots\circ A_C)}\right), \label{eq6}
\end{equation}
where $(A_1)_{ij}=\kappa_1(x_1^i,x_1^j)$, $(A_2)_{ij}=\kappa_2(x_2^i,x_2^j)$, $\cdots$, $(A_C)_{ij}=\kappa_k(x_C^i,x_C^j)$, and $\circ$ denotes the Hadamard product.
\end{definition}

The following corollary (see proof in~\cite{yu2018multivariate}) serve as a foundation for our~\emph{Definition~2}. Specifically, the first inequality indicates that the joint entropy of a set of variables is greater than or equal to the maximum of all of the individual entropies of the variables in the set, whereas the second inequality suggests that the joint entropy of a set of variable is less than or equal to the sum of the individual entropies of the variables in the set.

\begin{corollary}
Let $A_1$, $A_2$, $\cdots$, $A_C$ be $C$ $n\times n$ positive definite matrices with trace $1$ and nonnegative entries, and $(A_1)_{ii}=(A_2)_{ii}=\cdots=(A_C)_{ii}=\frac{1}{n}$, for $i=1,2,\cdots,n$. Then the following two inequalities hold:
\begin{equation} \label{corollary2.2}
\mathbf{S}_\alpha\left(\frac{A_1\circ A_2\circ\cdots\circ A_C}{\mathrm{tr}(A_1\circ A_2\circ\cdots\circ A_C)}\right)\geq\max
[\mathbf{S}_\alpha(A_1),\mathbf{S}_\alpha(A_2),\cdots,\mathbf{S}_\alpha(A_C)].
\end{equation}
\begin{equation} \label{corollary2.1}
\mathbf{S}_\alpha\left(\frac{A_1\circ A_2\circ\cdots\circ A_C}{\mathrm{tr}(A_1\circ A_2\circ\cdots\circ A_C)}\right)\leq
\mathbf{S}_\alpha(A_1)+\mathbf{S}_\alpha(A_2)+\cdots+\mathbf{S}_\alpha(A_C),
\end{equation}
\end{corollary}

\subsection{Multivariate mutual information estimation in CNNs}


\begin{wrapfigure}{r}{0.25\textwidth} 
    \centering
    \includegraphics[width=0.25\textwidth]{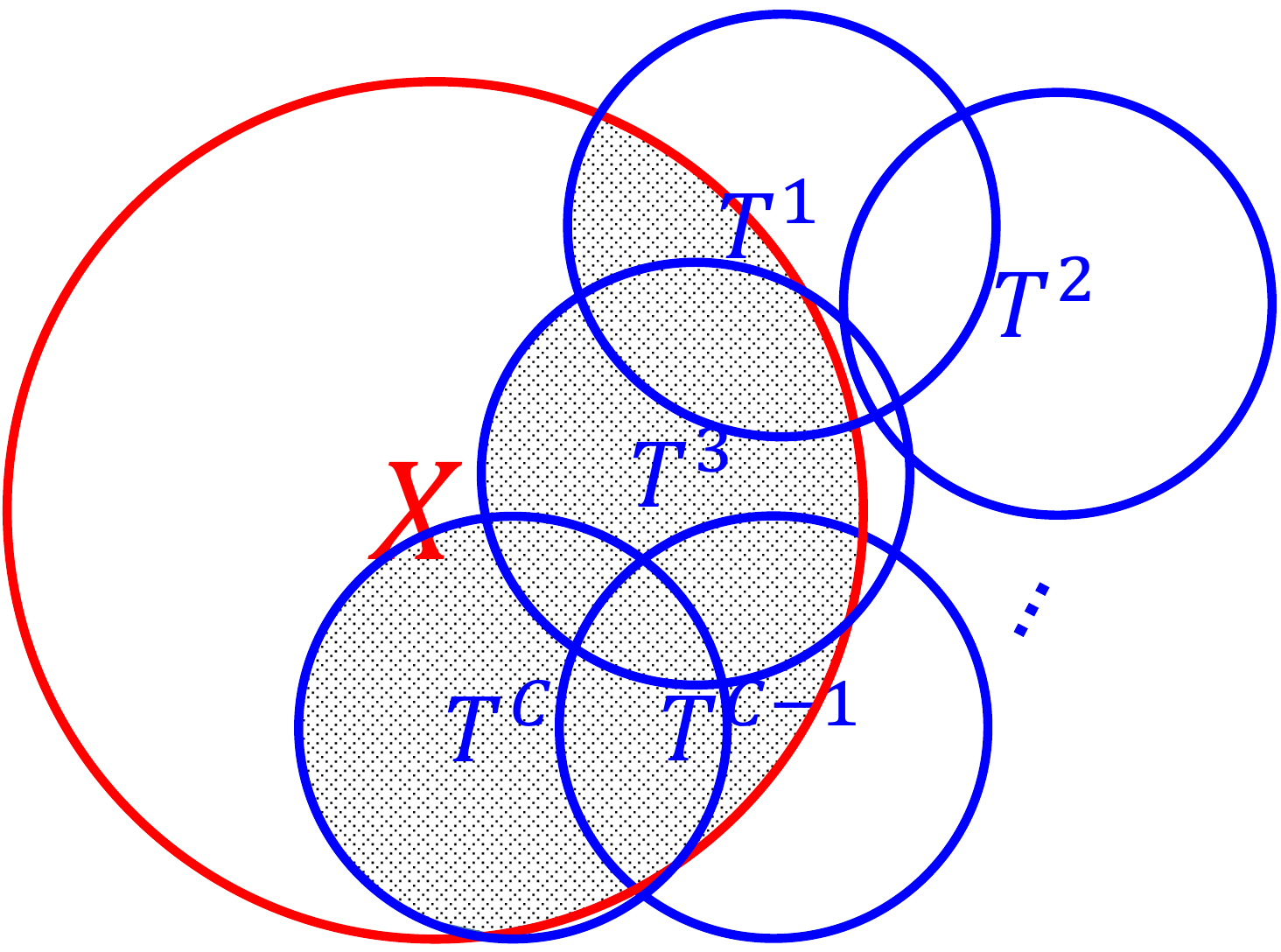}
    \caption{Venn diagram of $\mathbf{I}(X;\{T^1,T^2,\cdots,T^C\})$.}\label{venn_diagram_MMI}
\end{wrapfigure}

Suppose there are $C$ filters in the convolutional layer, then an input image is therefore represented by $C$ different feature maps. Each feature map characterizes a specific property of the input. This suggests that the amount of information that the convolutional layer gained from input $X$ is preserved in $C$ information sources $T^1$, $T^2$, $\cdots$, $T^C$. The Venn Diagram for $X$ and $T^1$, $\cdots$, $T^C$ is demonstrated in Fig.~\ref{venn_diagram_MMI}. Specifically, the red circle represents the information contained in $X$, each blue circle represents the information contained in each feature map. The amount of information in $X$ that is gained from $C$ feature maps (i.e., $\mathbf{I}(X;\{T^1,T^2,\cdots,T^C\})$) is exactly the shaded area. By applying the inclusion-exclusion principle~\cite{yeung1991new}, this shaded area can be computed by summing up the area of the red circle (i.e., $\mathbf{H}(X)$) with the area occupied by all blue circles (i.e., $\mathbf{H}(T^1,T^2,\cdots,T^C)$), and then subtracting the total joint area occupied by the red circle and all blue circles (i.e., $\mathbf{H}(X,T^1,T^2,\cdots,T^C)$). Formally speaking, this indicates that:
\begin{multline}
\mathbf{I}(X;\{T^1,T^2,\cdots,T^C\})=\mathbf{H}(X)+\mathbf{H}(T^1,T^2,\cdots,T^C)\\-\mathbf{H}(X,T^1,T^2,\cdots,T^C), \label{eq7}
\end{multline}
where $\mathbf{H}$ denotes entropy for a single variable or joint entropy for a group of variables.


Given Eqs.~(\ref{eq5}), (\ref{eq6}) and (\ref{eq7}), $\mathbf{I}(X;\{T^1,T^2,\cdots,T^C\})$ in a mini-batch of size $n$ can be estimated with:
\begin{multline} \label{eq8}
\small
\mathbf{I}_\alpha(B;\{A_1,A_2,\cdots,A_C\})=\mathbf{S}_\alpha(B)+\mathbf{S}_\alpha\left(\frac{A_1\circ A_2\circ\cdots\circ A_C}{\mathrm{tr}(A_1\circ A_2\circ\cdots\circ A_C)}\right)\\
-\mathbf{S}_\alpha\left(\frac{A_1\circ A_2\circ\cdots\circ A_C\circ B}{\mathrm{tr}(A_1\circ A_2\circ\cdots\circ A_C\circ B)}\right).
\end{multline}

Here, $B$, $A_1,\cdots, A_C$ denote Gram matrices evaluated on the input tensor and $C$ feature maps tensors, respectively. For example, $A_p$ ($1\leq p\leq C$) is evaluated on $\{x_p^i\}_{i=1}^n$, in which $x_p^i$ refers to the feature map generated from the $i$-th input sample using the $p$-th filter. Obviously, instead of estimating the joint PDF on $\{X,T^1,T^2,\cdots,T^C\}$ which is typically unattainable, one just needs to compute $(C+1)$ Gram matrices using a real valued positive definite kernel that is also infinitely divisible~\cite{bhatia2006infinitely}.

\section{Main Results} \label{sec3}
This section presents three sets of experiments to empirically validate the existence of two DPIs in CNNs, using the novel nonparametric information theoretic (IT) estimators put forth in this work. Specifically, Section~\ref{sec3.1} validates the existence of two DPIs in CNNs. In Section~\ref{sec3.2}, we illustrate, via the application of the PID framework in the training phase, some interesting observations associated with different CNN topologies. Following this, we present implications to the determination of network depth and width motivated by these results. We finally point out, in Section~\ref{sec3.4}, an advanced interpretation to the information plane (IP) that deserve more (theoretical) investigations.
Four real-world data sets, namely MNIST~\cite{lecun1998gradient}, Fashion-MNIST~\cite{xiao2017fashion}, HASYv2~\cite{thoma2017hasyv2}, and Fruits~360~\cite{murecsan2018fruit}, are selected for evaluation. The characteristics of each data set are summarized in Table~\ref{tab:data_character}. Note that, compared with the benchmark MNIST and Fashion-MNIST, HASYv2 and Fruits~360 have significantly larger number of classes as well as higher intraclass variance. For example, in Fruits~360, the apple class contains different varieties (e.g., Crimson Snow, Golden, Granny Smith), and the images are captured with different viewpoints and varying illumination conditions. Due to page limitations, we only demonstrate the most representative results in the rest of this paper. Additional experimental results are available in Appendix~\ref{appendix_C}. 

\vspace{-0.3cm}
\begin{table}[!h]
\setlength{\abovecaptionskip}{-0.0cm}
\setlength{\belowcaptionskip}{-0.0cm}
\centering
\caption{The number of classes ($\#$ Class), the number of training samples ($\#$ Train), the number of testing samples ($\#$ Test), and the sample size of selected data sets.}\label{tab:data_character}
\begin{tabular}{ccccccc}\hline
 & $\#$ Class & $\#$ Train & $\#$ Test & Sample Size \\\hline
$\text{MNIST}$~\cite{lecun1998gradient} & $10$ & $60,000$ & $10,000$ & $28\times28$ \\
$\text{Fashion-MNIST}$~\cite{xiao2017fashion} & $10$ & $60,000$ & $10,000$ & $28\times28$ \\
$\text{HASYv2}$~\cite{thoma2017hasyv2} & $369$ & $151,406$ & $16,827$ & $32\times32$ \\
$\text{Fruits~360}$~\cite{murecsan2018fruit} & $111$ & $56,781$ & $19,053$ & $100\times100$ \\\hline
\end{tabular}
\end{table}
\vspace{-0.2cm}



For MNIST and Fashion-MNIST, we consider a LeNet-$5$~\cite{lecun1998gradient} type network which consists of $2$ convolutional layers, $2$ pooling layers, and $2$ fully connected layers. For HASYv2 and Fruits~360, we use a smaller AlexNet~\cite{krizhevsky2012imagenet} type network with $4$ convolutional layers (but fewer filters in each layer) and $3$ fully connected layers. We train the CNN with SGD with momentum $0.95$ and mini-batch size $128$. In MNIST and Fashion-MNIST, we select learning rate $0.1$ and $10$ training epochs. In HASYv2 and Fruits~$360$, we select learning rate $0.01$ and $15$ training epochs. Both ``sigmoid" and ``ReLU" activation functions are tested. For the estimation of MMI, we fix $\alpha=1.01$ to approximate Shannon's definition, and use the radial basis function (RBF) kernel $\kappa(x_i,x_j)=\exp(-\frac{\|x_i-x_j\|^2}{2\sigma^2})$ to obtain the Gram matrices. The kernel size $\sigma$ is determined based on the Silverman's rule of thumb~\cite{silverman1986density} $\sigma=h\times n^{-1/(4+d)}$, where $n$ is the number of samples in the mini-batch ($128$ in this work), $d$ is the sample dimensionality and $h$ is an empirical value selected experimentally by taking into consideration the data's average marginal variance. In this paper, we select $h=5$ for the input signal forward propagation chain and $h=0.1$ for the error backpropagation chain.


\subsection{Experimental Validation of Two DPIs} \label{sec3.1}

We expect the existence of two DPIs in any feedforward CNNs with $K$ hidden layers, i.e., $\mathbf{I}(X,T_1)\geq \mathbf{I}(X,T_2)\geq \cdots\geq \mathbf{I}(X,T_K)$ and $\mathbf{I}(\delta_K,\delta_{K-1})\geq \mathbf{I}(\delta_K,\delta_{K-2})\geq \cdots \geq \mathbf{I}(\delta_K,\delta_1)$, where $T_1$, $T_2$, $\cdots$, $T_K$ are successive
hidden layer representations from the first hidden layer to the output layer and $\delta_K$, $\delta_{K-1}$, $\cdots$, $\delta_1$ are errors from the output layer to the first hidden layer. This is because both $X\rightarrow T_1\rightarrow\cdots\rightarrow T_K$ and $\delta_K\rightarrow \delta_{K-1}\rightarrow\cdots\rightarrow \delta_1$ form a Markov chain~\cite{shwartz2017opening,yu2018understanding}.

\begin{figure}[!htbp]
\centering
\begin{tabular}{ccc}
\subfigure[initial iteration] {\includegraphics[width=.16\textwidth,height=1.5cm]{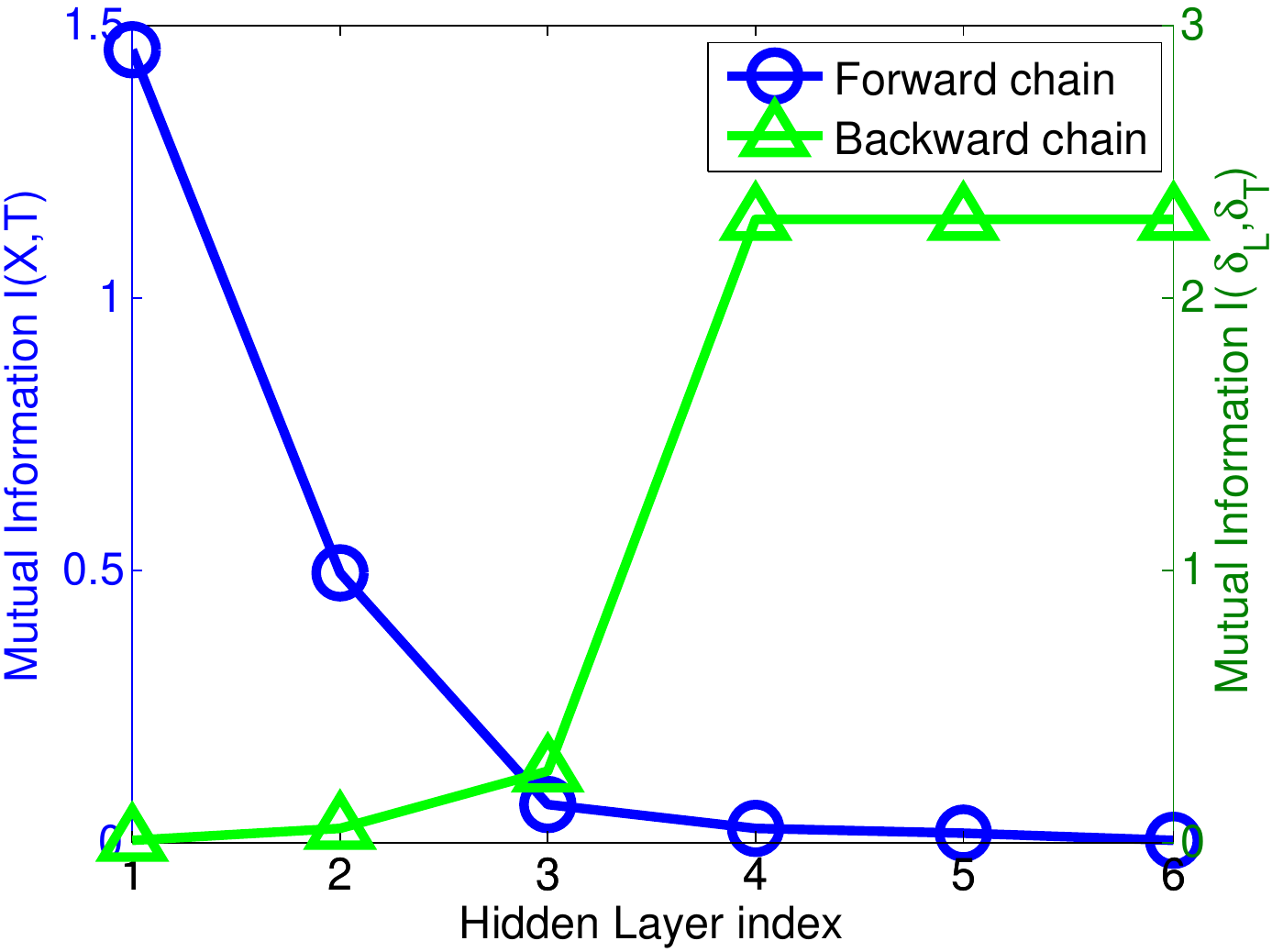}}
\subfigure[1 epochs later] {\includegraphics[width=.16\textwidth,height=1.5cm]{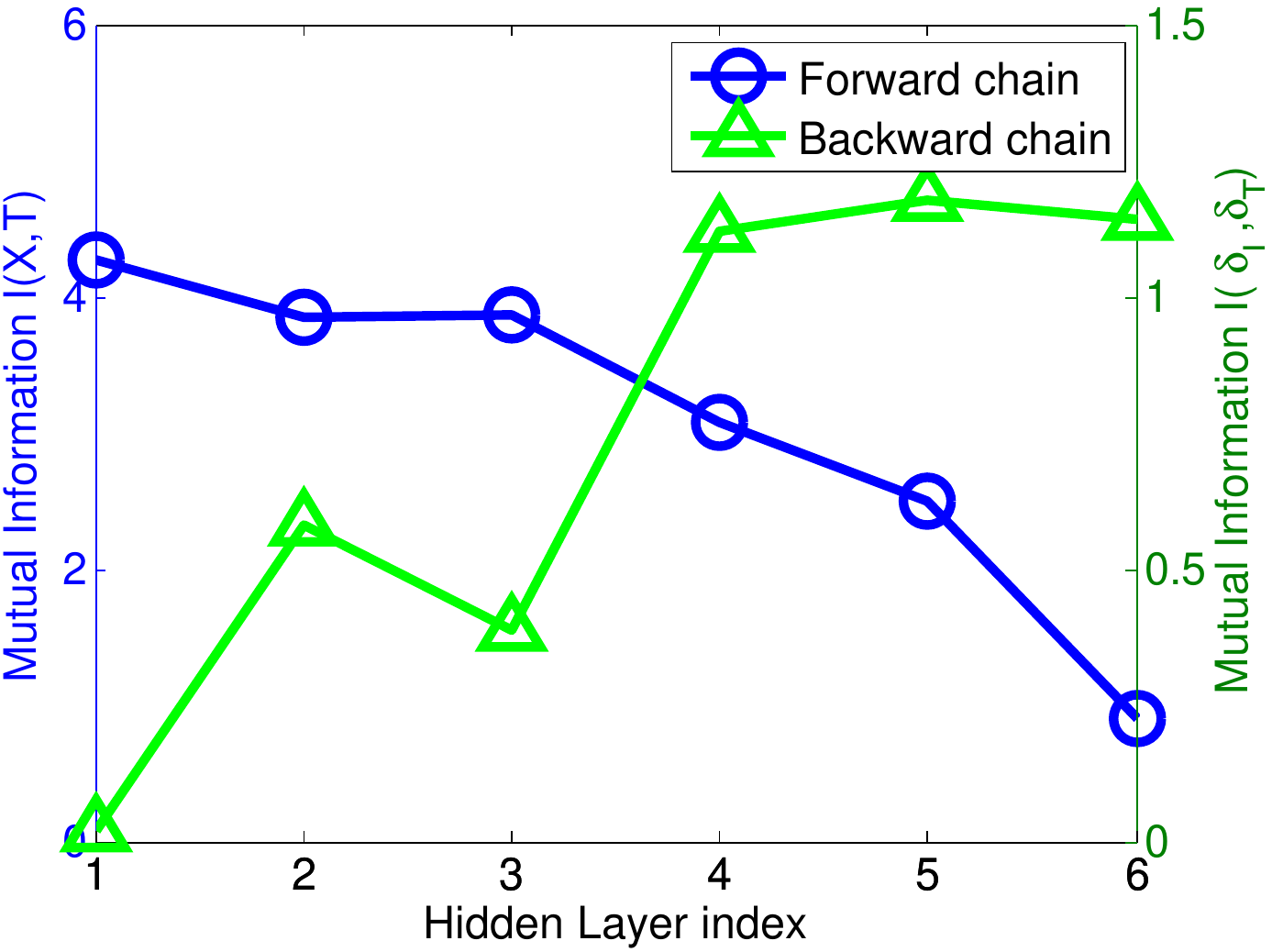}}
\subfigure[10 epochs later] {\includegraphics[width=.16\textwidth,height=1.5cm]{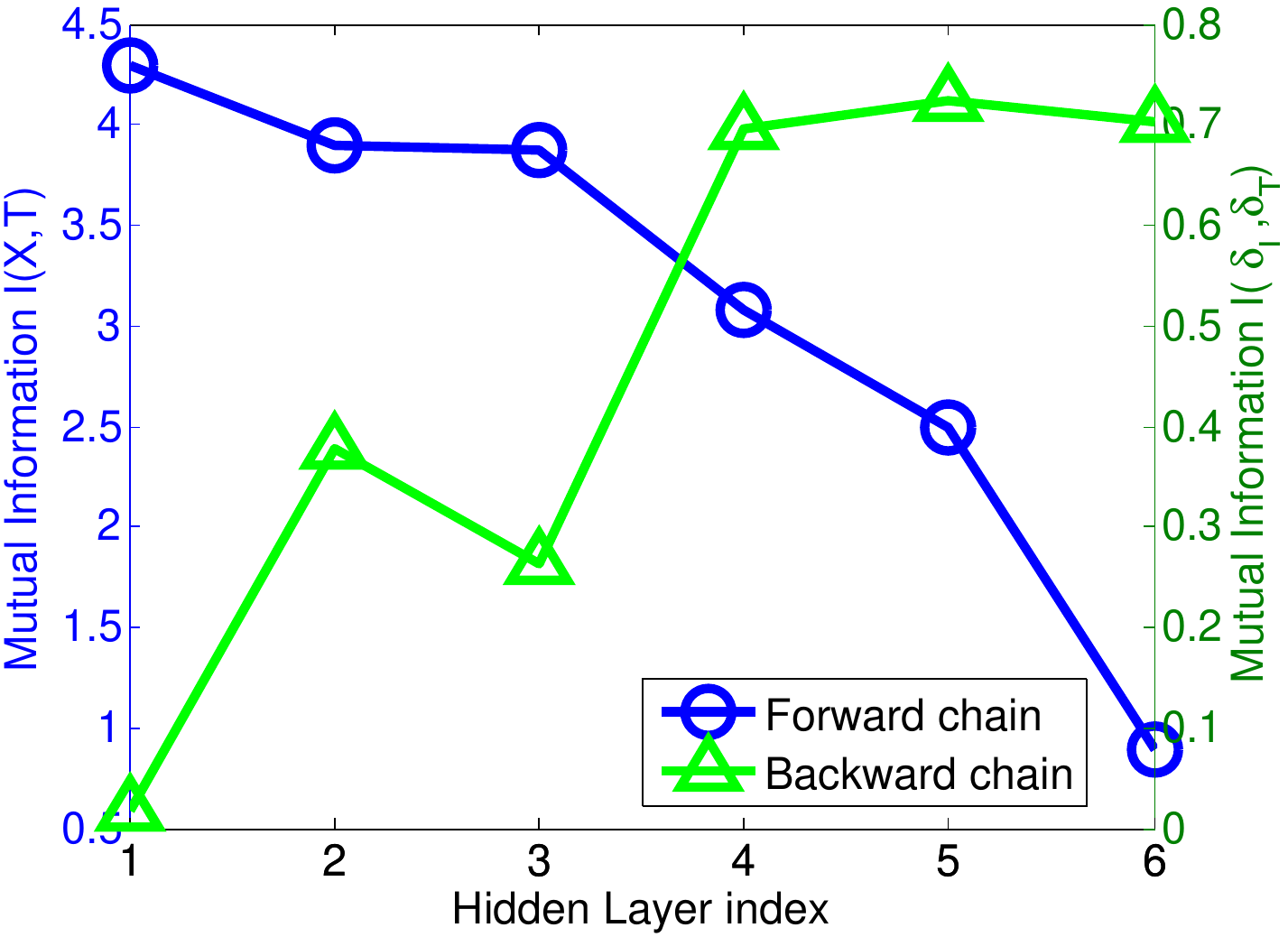}} & \\
\subfigure[initial iteration] {\includegraphics[width=.16\textwidth,height=1.5cm]{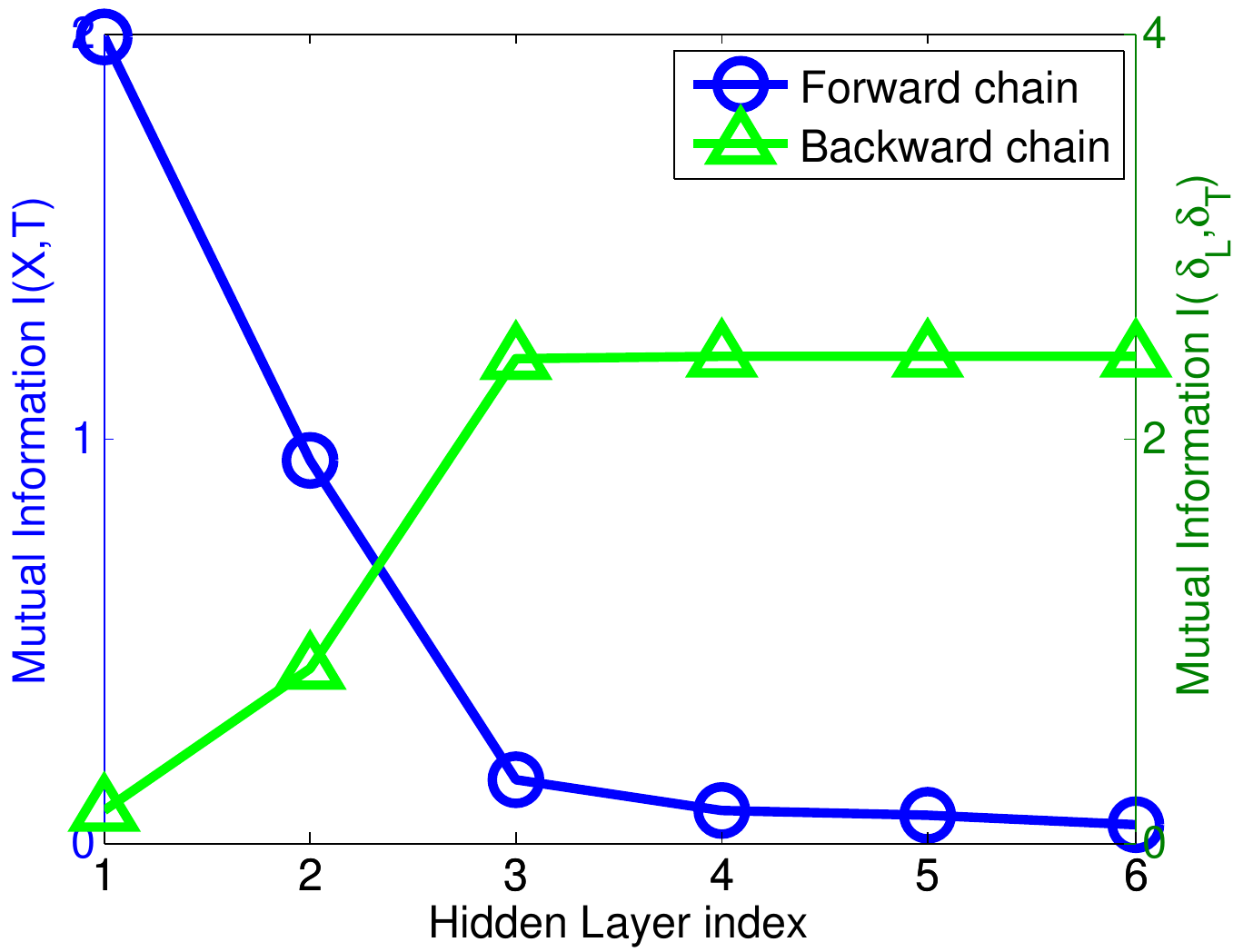}}
\subfigure[1 epochs later] {\includegraphics[width=.16\textwidth,height=1.5cm]{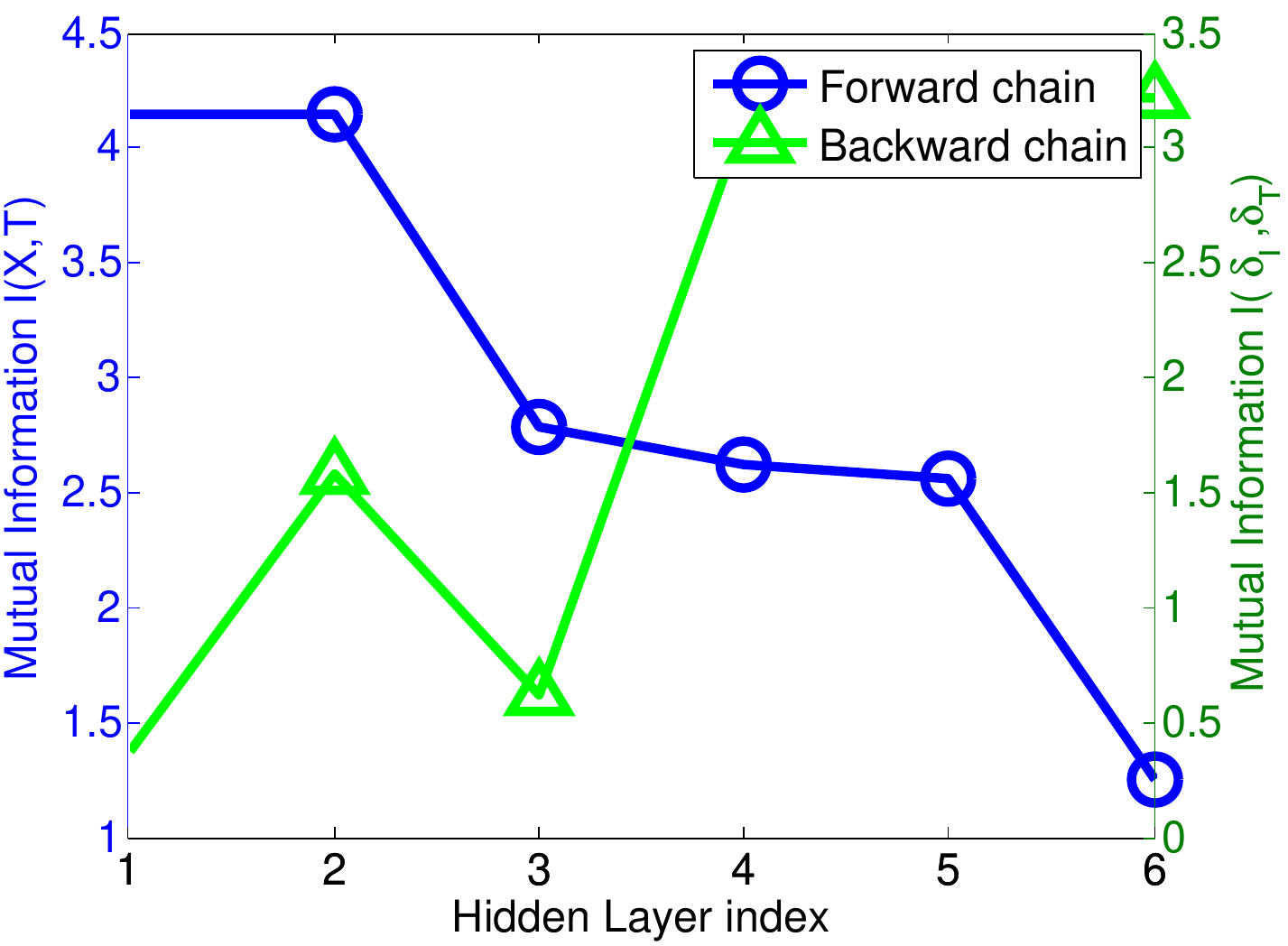}}
\subfigure[10 epochs later] {\includegraphics[width=.16\textwidth,height=1.5cm]{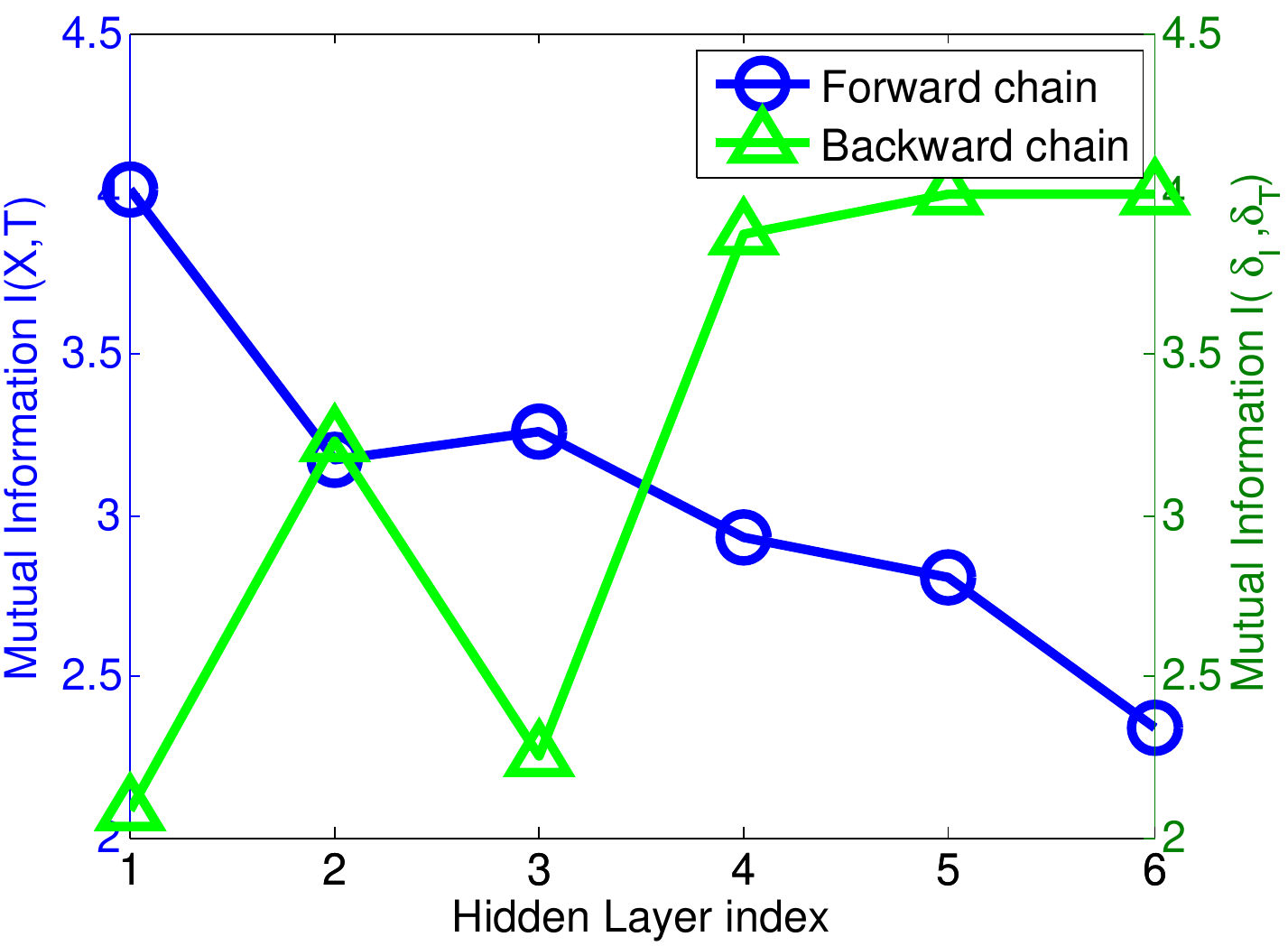}} & \\
\subfigure[initial iteration] {\includegraphics[width=.16\textwidth,height=1.5cm]{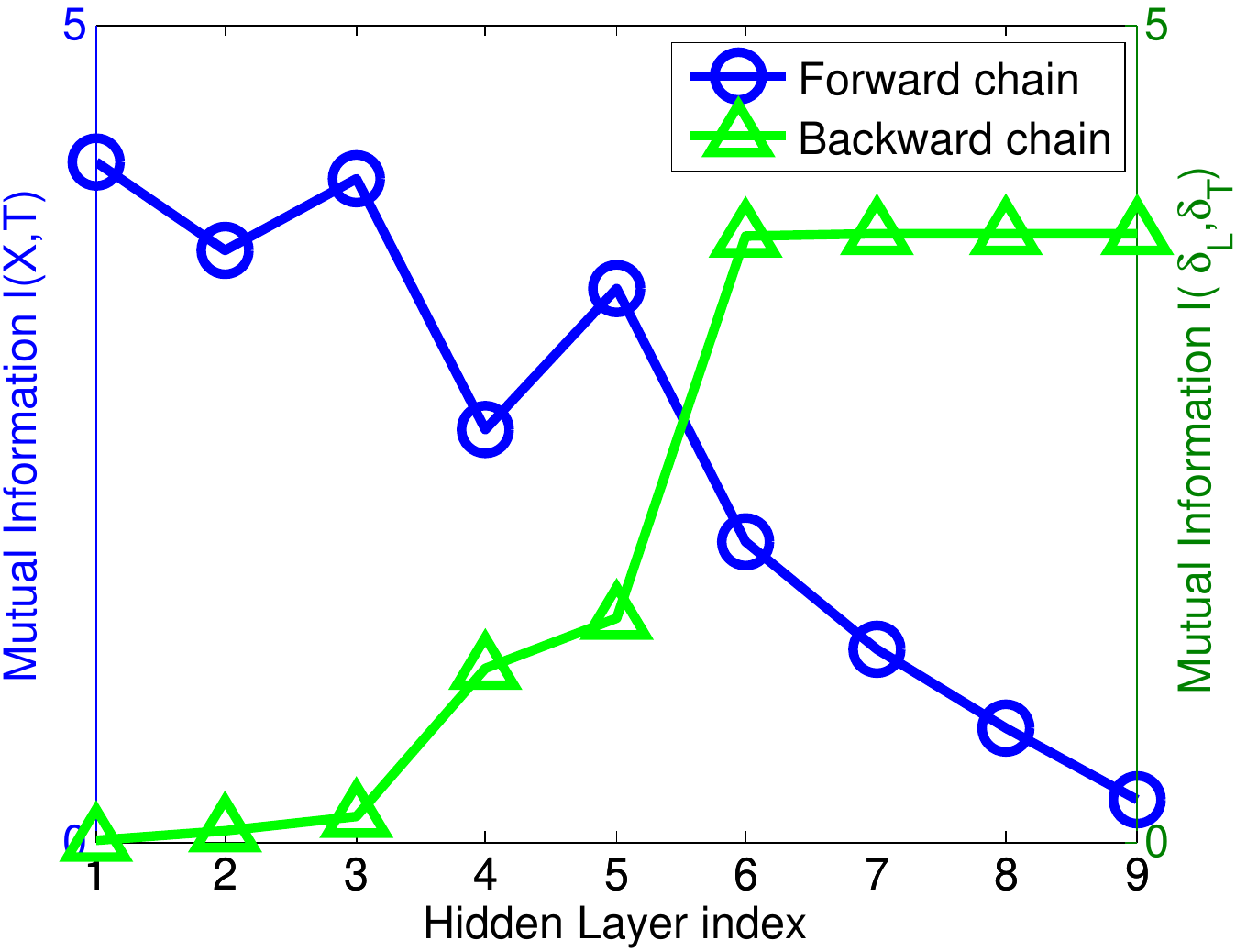}}
\subfigure[1 epoch later] {\includegraphics[width=.16\textwidth,height=1.5cm]{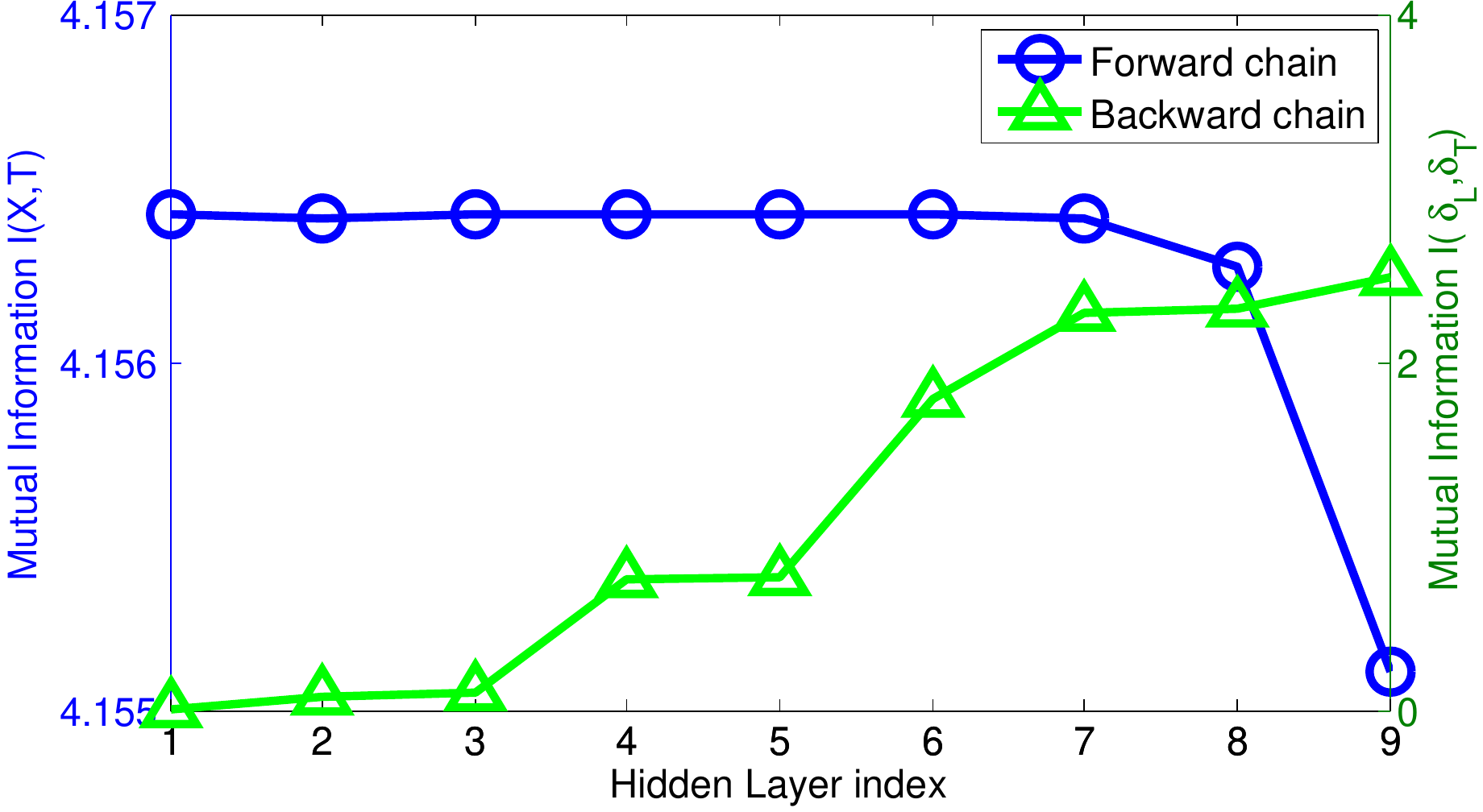}}
\subfigure[15 epochs later] {\includegraphics[width=.16\textwidth,height=1.5cm]{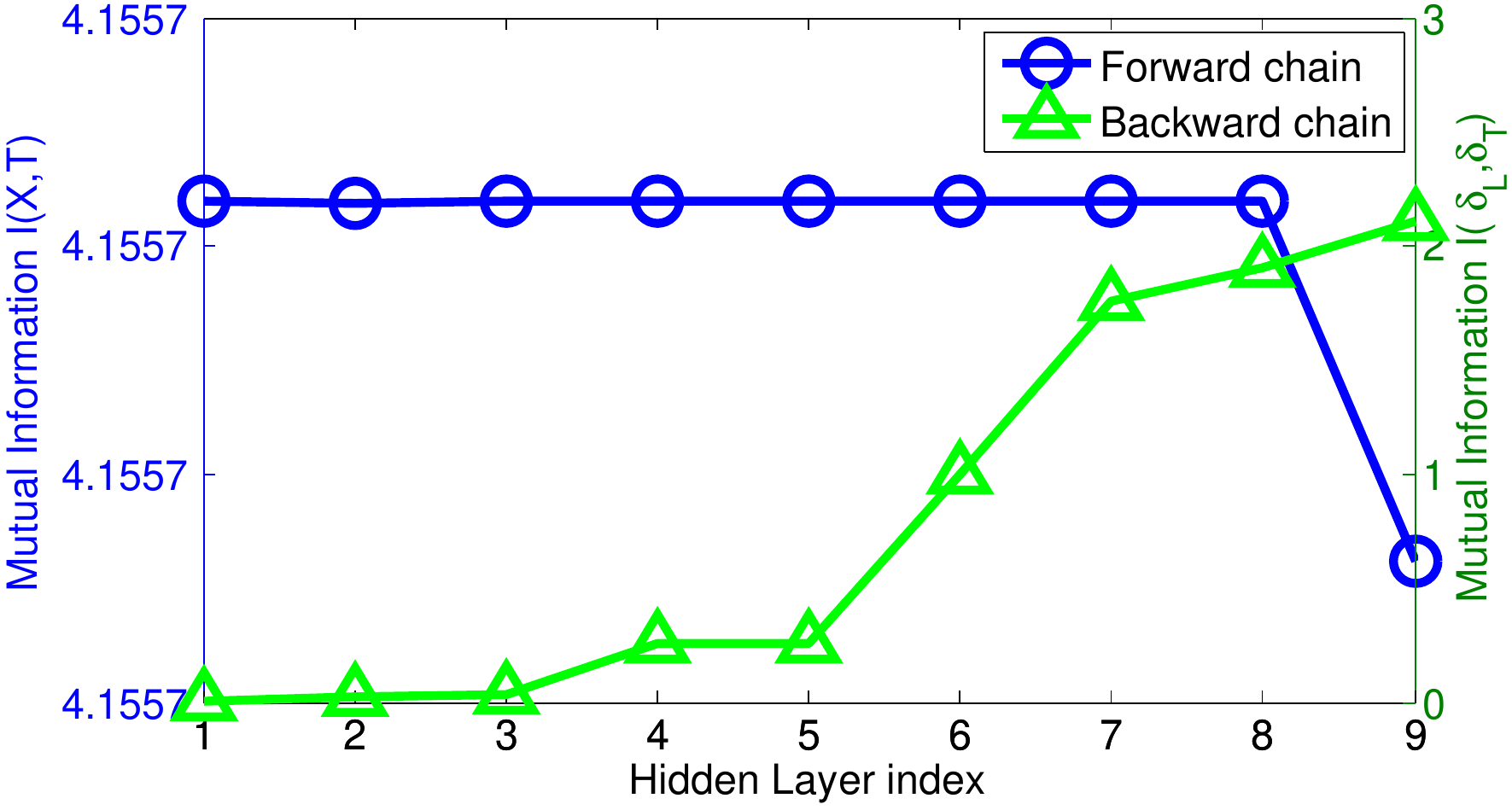}} & \\
\subfigure[initial iteration] {\includegraphics[width=.16\textwidth,height=1.5cm]{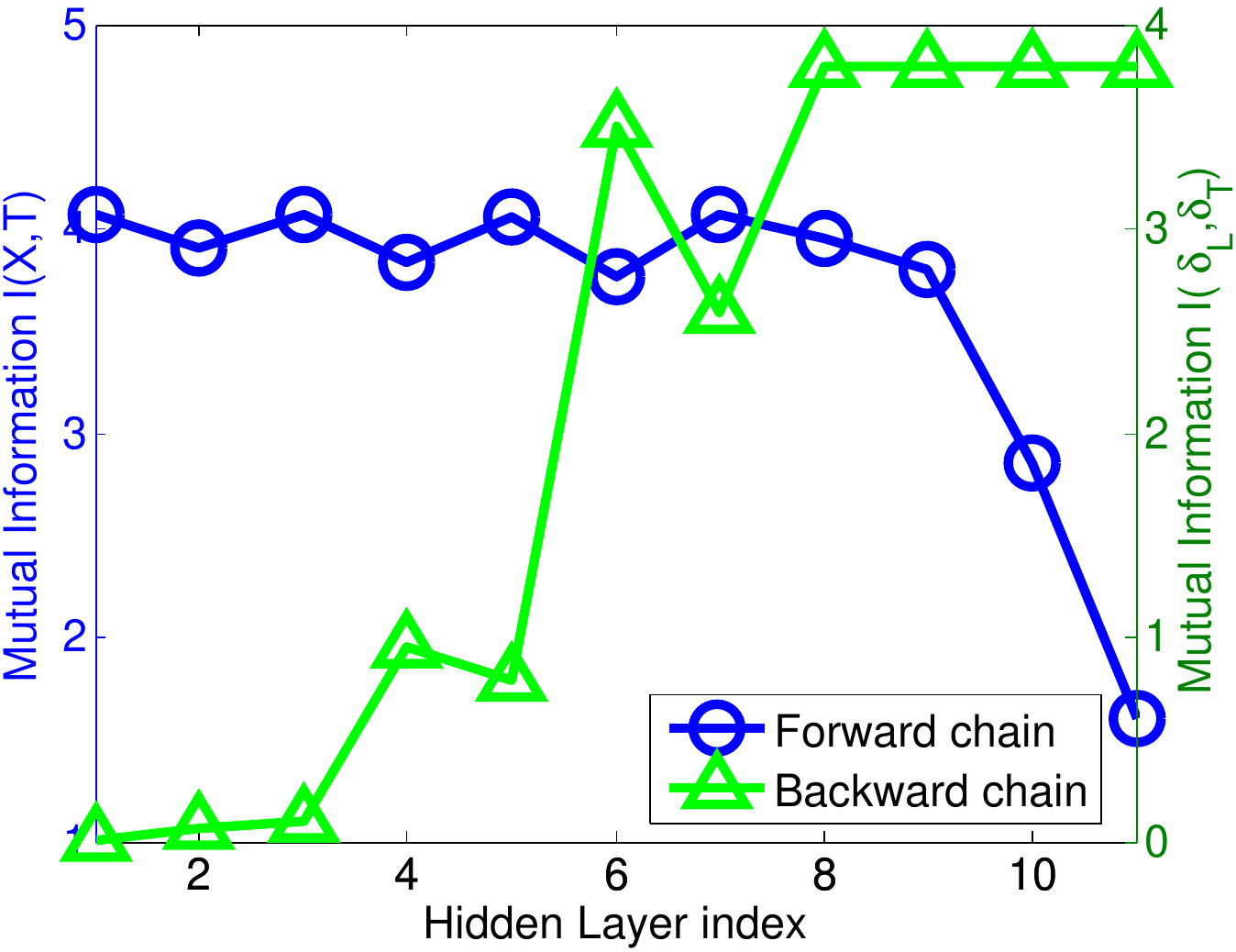}}
\subfigure[1 epoch later] {\includegraphics[width=.16\textwidth,height=1.5cm]{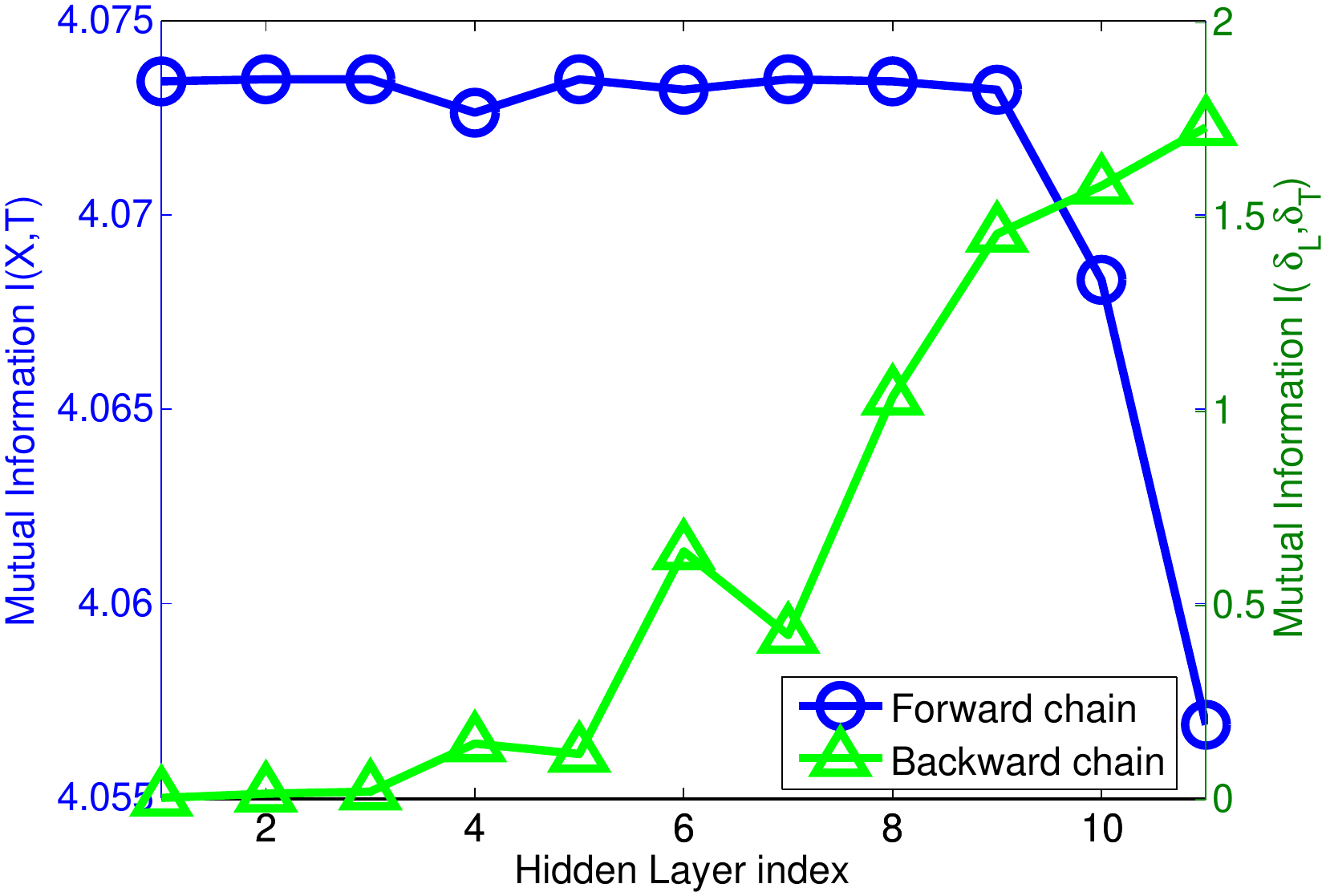}}
\subfigure[15 epochs later] {\includegraphics[width=.16\textwidth,height=1.5cm]{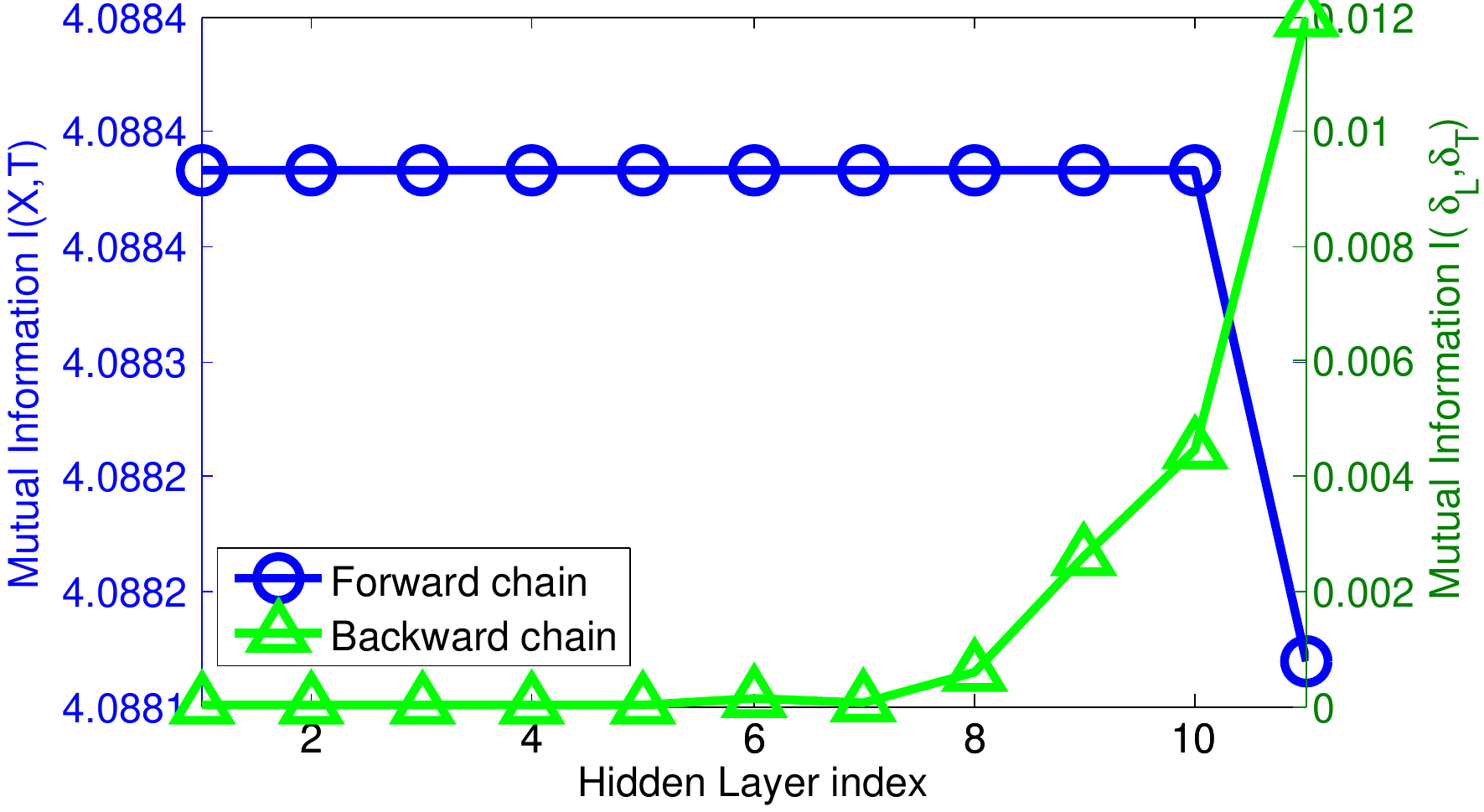}} & \\
\end{tabular}
\caption{Two DPIs in CNNs. The first row shows the validation results, on MNIST data set, obtained by a CNN with $6$ filters in the $1$st convolutional layer (denote Conv.~$1$) and $12$ filters in the $2$nd convolutional layer (denote Conv.~$2$); the second row shows the validation results, on Fashion-MNIST data set, obtained by the same CNN architecture as in the MNIST; the third row shows the validation results, on HASYv2 data set, obtained by a CNN with $30$ filters in Conv.~$1$, $60$ filters in Conv.~$2$ and $90$ filters in Conv.~$3$; whereas the fourth row shows the validation results, on Fruits~360 data set, obtained by a CNN with $16$ filters in Conv.~$1$, $32$ filters in Conv.~$2$, $64$ filters in Conv.~$3$ and $128$ filters in Conv.~$4$. In each subfigure, the blue curves show the MMI values between input and different layer representations, whereas the green curves show the MMI values between errors in the output layer and different hidden layers. 
\vspace{-0.5cm}}
\label{fig:DPI}
\end{figure}

Fig.~\ref{fig:DPI} shows the DPIs at the initial training stage, after $1$ epoch's training and at the final training stage, respectively. As can be seen, DPIs hold in most of the cases. Note that, there are a few disruptions in the error backpropagation chain, because the curves should be monotonic according to the theory. One possible reason is that when training converges, the error becomes tiny such that Sliverman's rule of thumb is no longer a reliable choice to select scale parameter $\sigma$ in our estimator.


\subsection{Redundancy and Synergy in Layer Representations} \label{sec3.2}
In this section, we explore properties of different IT quantities during the training of CNNs, with the help of the PID framework. Particularly, we are interested in determining the redundancy and synergy amongst different feature maps and how they evolve with training in different CNN topologies. Moreover, we are also interested in identifying some upper and lower limits (if they exist) for these quantities. However, the analysis is not easy because the set of information equations is underdetermined as we will show next.

\begin{figure}[!htbp]
\centering
\begin{tabular}{ccc}
\subfigure[$\mathbf{I}(X;\{T^1,T^2\})$] {\includegraphics[width=.23\textwidth]{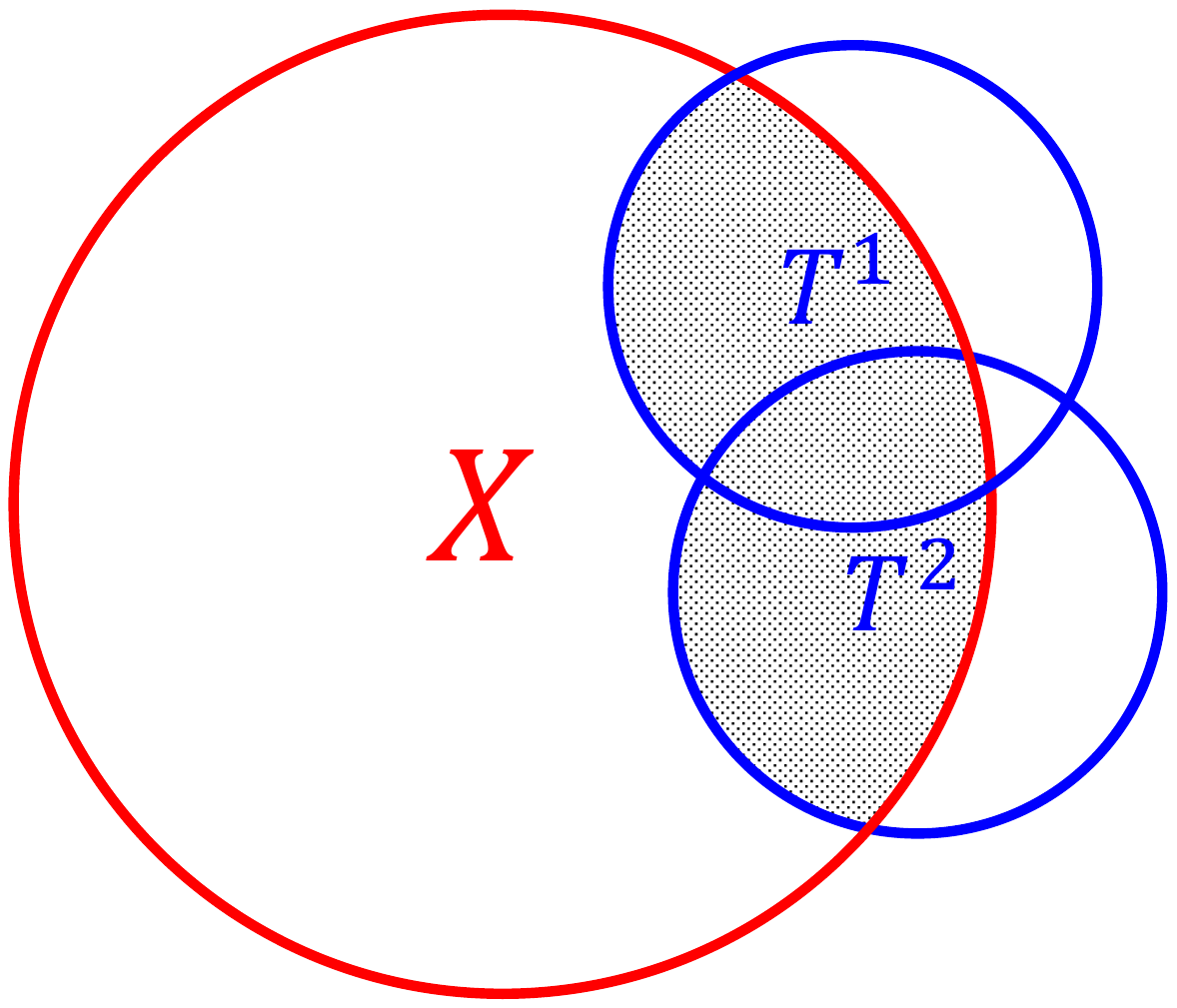}}
\subfigure[PID to $\mathbf{I}(X;\{T^1,T^2\})$] {\includegraphics[width=.23\textwidth]{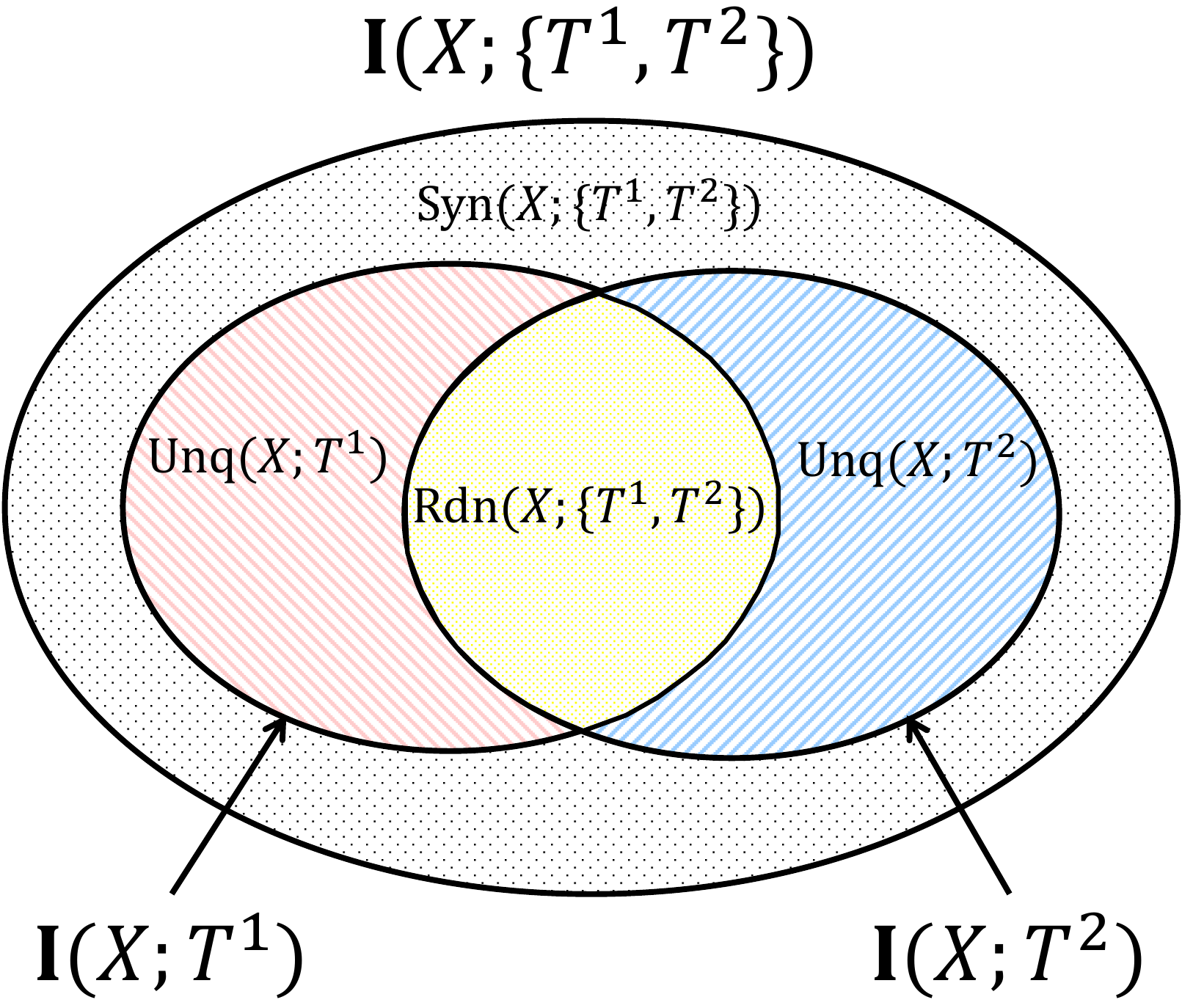}}
\end{tabular}
\caption{Synergy and redundancy amongst different feature maps. (a) shows the interactions between input signal and two feature maps. The shadow area indicates the MMI $\mathbf{I}(X;\{T^1,T^2\})$. (b) shows the PID to $\mathbf{I}(X;\{T^1,T^2\})$.
\vspace{-0.0cm}}
\label{fig:PID}
\end{figure}

Given input signal $X$ and two feature maps $T^1$ and $T^2$, the PID framework indicates that the MMI $\mathbf{I}(X;\{T^1,T^2\})$ can be decomposed into four non-negative IT components: the synergy $\mathbf{Syn}(X;\{T^1,T^2\})$ that measures the information about $X$ provided by the combination of $T^1$ and $T^2$ (i.e., the information that cannot be captured by either $T^1$ or $T^2$ alone); the redundancy $\mathbf{Rdn}(X;\{T^1,T^2\})$ that measures the shared information about $X$ that can be provided by either $T^1$ or $T^2$; the unique information $\mathbf{Unq}(X;T^1)$ (or $\mathbf{Unq}(X;T^2)$) that measures the information about $X$ that can only be provided by $T^1$ (or $T^2$). Moreover, the unique information, the synergy and the redundancy satisfy (see Fig.~\ref{fig:PID}):
\begin{multline}\label{eq_PID1}
\small
\mathbf{I}(X;\{T^1,T^2\}) = \mathbf{Syn}(X;\{T^1,T^2\})+\mathbf{Rdn}(X;\{T^1,T^2\})\\+\mathbf{Unq}(X;T^1)+\mathbf{Unq}(X;T^2);
\end{multline}
\begin{equation}\label{eq_PID2}
\small
\mathbf{I}(X;T^1) = \mathbf{Rdn}(X;\{T^1,T^2\})+\mathbf{Unq}(X;T^1);
\end{equation}
\begin{equation}\label{eq_PID3}
\small
\mathbf{I}(X;T^2) = \mathbf{Rdn}(X;\{T^1,T^2\})+\mathbf{Unq}(X;T^2).
\end{equation}

Notice that we have $4$ IT components (i.e., synergy, redundancy, and two unique information terms), but only $3$ measurements: $\mathbf{I}(X;\{T^1,T^2\})$, $\mathbf{I}(X;T^1)$, and $\mathbf{I}(X;T^2)$. Therefore, we can never determine uniquely the $4$ IT quantities. This decomposition for $\mathbf{I}(X;\{T^1,T^2\})$ can be straightforwardly extended for more than three variables, thus decomposing $\mathbf{I}(X;\{T^1,T^2,\cdots,T^C\})$ into much more components. For example, if $C=4$, there will be $166$ individual non-negative items. Admittedly, the PID diagram (see Appendix~\ref{appendix_A} for more details) offers an intuitive understanding of the interactions between input and different feature maps, and our estimators have been shown appropriate for high dimensional data. However, the reliable estimation of each IT component still remains a bigger challenge, because of the undetermined nature of the problem. In fact, there is no universal agreement on the definition of synergy and redundancy among one-dimensional $3$-way interactions, let alone the estimation of each synergy or redundancy among numerous variables in high-dimensional spaces~\cite{bertschinger2014quantifying,griffith2014quantifying}. To this end, we develop three quantities based on the $3$ measurements by manipulating Eqs.~(\ref{eq_PID1})-(\ref{eq_PID3}) to characterize intrinsic properties of CNN layer representations. The new developed quantities avoid the direct estimation of synergy and redundancy. They are:\\
\noindent
1) $\mathbf{I}(X;\{T^1,T^2,\cdots,T^C\})$, which is exactly the MMI. This quantity measures the amount of information about $X$ that is captured by all feature maps (in one convolutional layer).\\

\noindent
2) $\small{\frac{2}{C(C-1)}\sum_{i=1}^{C}\sum_{j=i+1}^{C}\mathbf{I}(X;T^i)+\mathbf{I}(X;T^j)-\mathbf{I}(X;\{T^i,T^j\})}$, which is referred to redundancy-synergy trade-off. This quantity measures the (average) redundancy-synergy trade-off in different feature maps. By rewriting Eqs.~(\ref{eq_PID1})-(\ref{eq_PID3}), we have:
\begin{multline}
\mathbf{I}(X;T^i)+\mathbf{I}(X;T^j)-\mathbf{I}(X;\{T^i,T^j\})\\=\mathbf{Rdn}(X;\{T^i,T^j\})-\mathbf{Syn}(X;\{T^i,T^j\}).\label{eq_tradeoff}
\end{multline}

Obviously, a positive value of this trade-off implies redundancy, whereas a negative value signifies synergy~\cite{bell2003co}. Here, instead of measuring all PID terms that increase polynomially with $C$, we sample pairs of feature maps, calculate the information quantities for each pair, and finally compute averages over all pairs to determine if synergy dominates in the training phase. Note that, the pairwise sampling procedure has been used in neuroscience~\cite{timme2014synergy} and a recent paper on information theoretic investigation of Restricted Boltzmann Machine (RBM)~\cite{tax2017partial}.\\

\noindent
3) $\small{\frac{2}{C(C-1)}\sum_{i=1}^{C}\sum_{j=i+1}^{C}2\times\mathbf{I}(X;\{T^i,T^j\})-\mathbf{I}(X;T^i)-\mathbf{I}(X;T^j)}$, which is referred to weighted non-redundant information. This quantity measures the (average) amount of non-redundant information about $X$ that is captured by pairs of feature maps. As can be seen, from Eqs.~(\ref{eq_PID1})-(\ref{eq_PID3}),
\begin{multline}
\small
2\times\mathbf{I}(X;\{T^i,T^j\})-\mathbf{I}(X;T^i)-\mathbf{I}(X;T^j)\\=\mathbf{Unq}(X;T^i)+\mathbf{Unq}(X;T^j)+2\times\mathbf{Syn}(X;\{T^i,T^j\}).\label{eq_non_redunt}
\end{multline}

We call this quantity ``weighted" because we overemphasized the role of synergy, but notice that redundancy does not explicitly appear, while the two unique information terms reappear.


One should note that, Eqs.~(\ref{eq_tradeoff}) and (\ref{eq_non_redunt}) are just two of many equations that can be written, but all are going to be a linear combination of more than one IT component. Therefore, we do not introduce any errors in computing Eqs.~(\ref{eq_tradeoff}) and (\ref{eq_non_redunt}), we simply work on the linear projected space of synergy and redundancy. We will now experimentally show how these two pairs of IT components (synergy and redundancy from Eq.~(\ref{eq_tradeoff}), and synergy with the two unique information terms from Eq.~(\ref{eq_non_redunt})) evolve across different CNN layer changes.


\begin{figure}[!htbp]
\centering
\begin{tabular}{ccc}
\subfigure[MMI in MNIST] {\includegraphics[width=.23\textwidth]{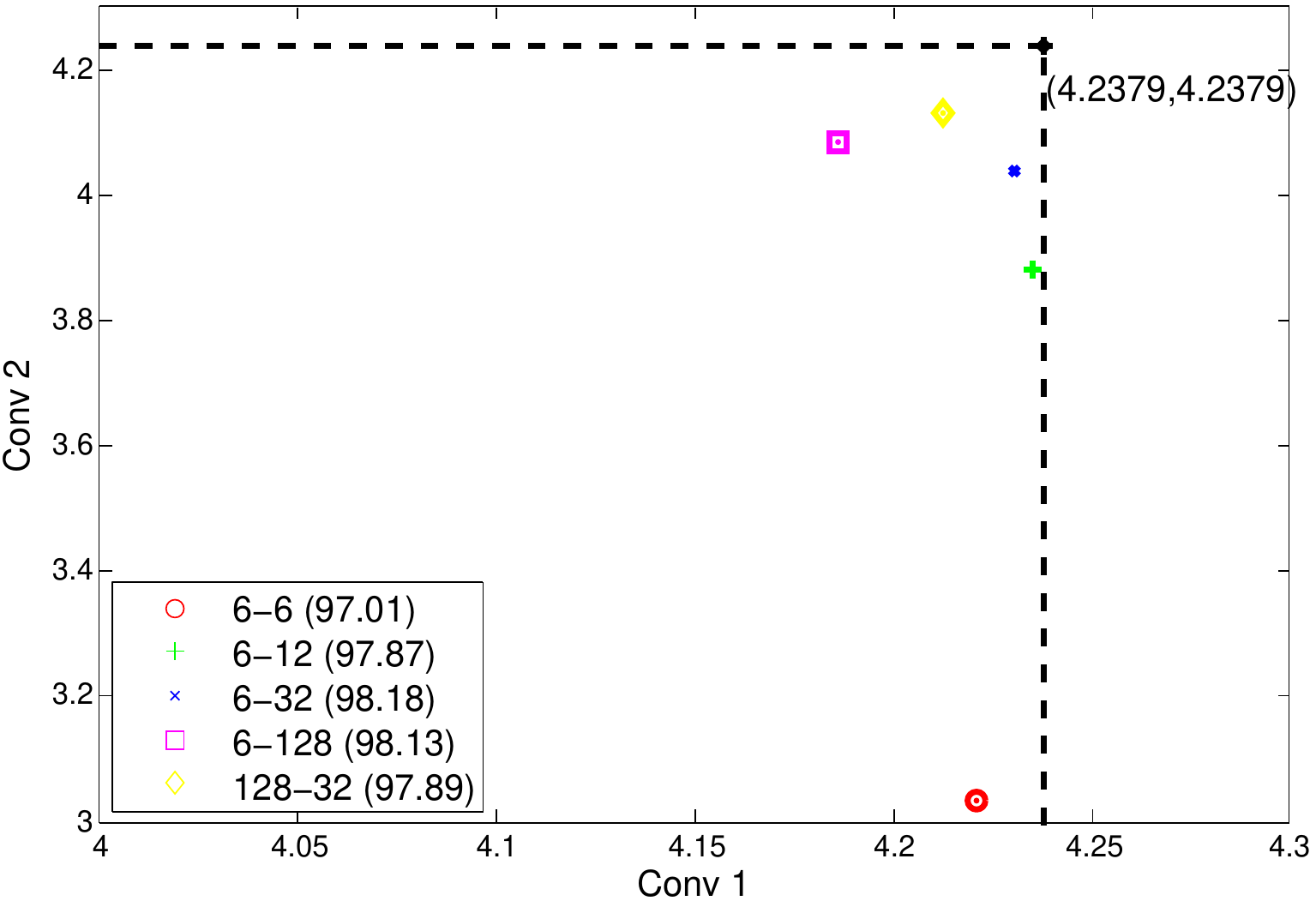}}
\subfigure[MMI in Fruits 360] {\includegraphics[width=.23\textwidth]{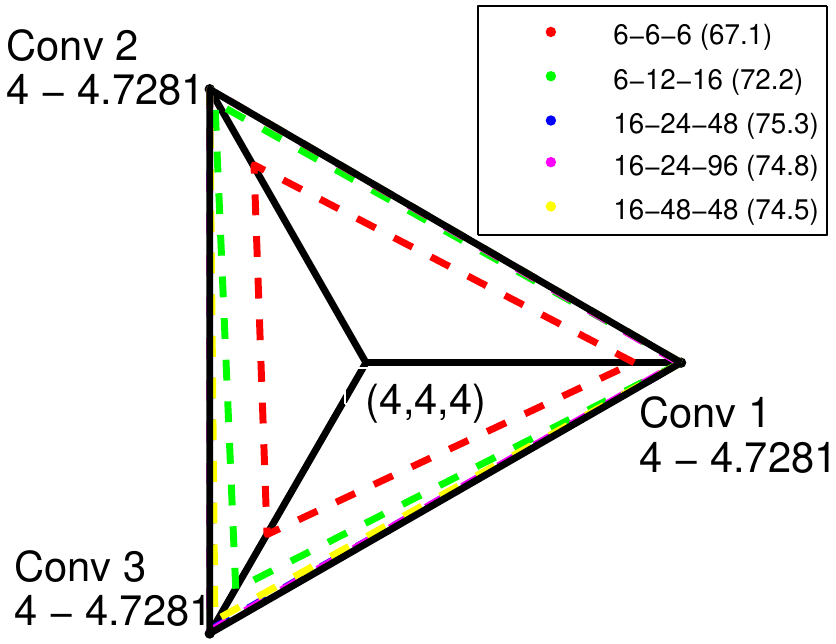}}
\end{tabular}
\caption{The MMI values in (a) Conv. $1$ and Conv. $2$ in MNIST data set; and (b) Conv. $1$, Conv. $2$ and Conv. $3$ in Fruits $360$ data set. The black line indicates the upper bound of MMI, i.e., the average mini-batch input entropy. The topologies of all competing networks are specified in the legend, in which the successive numbers indicate the number of filters in each convolutional layer. We also report their classification accuracies ($\%$) on testing set averaged over $10$ Monte-Carlo simulations in the parentheses.
\vspace{-0.4cm}}
\label{fig:MMI_saturation}
\end{figure}

\begin{figure*}[!htbp]
\centering
\begin{tabular}{ccccc}
\subfigure[Redundancy-Synergy trade-off. The networks differ in the number of filters in Conv.~$1$.] {\includegraphics[width=.23\textwidth]{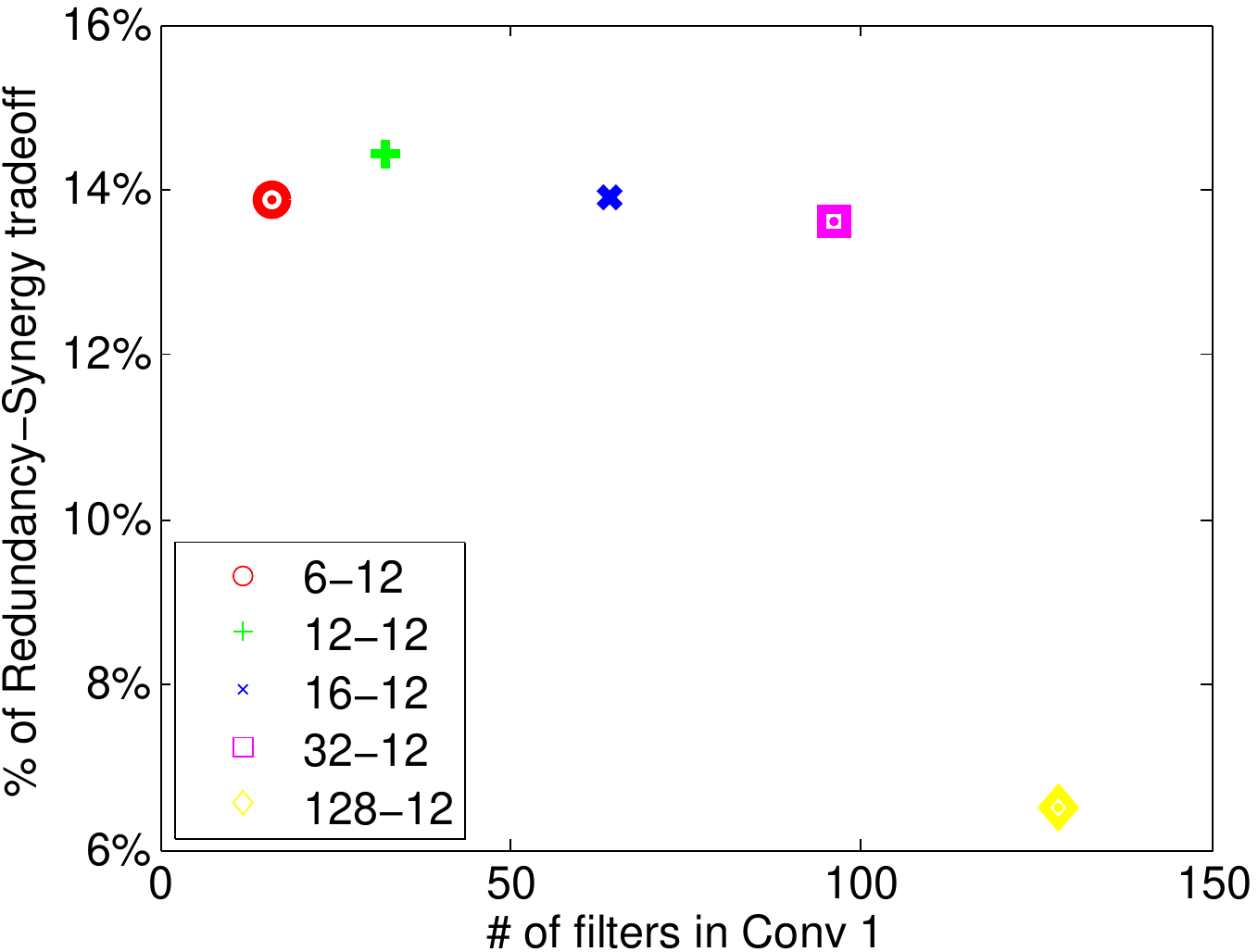}}
\subfigure[Weighted non-redundant information. The networks differ in the number of filters in Conv.~$1$.] {\includegraphics[width=.23\textwidth]{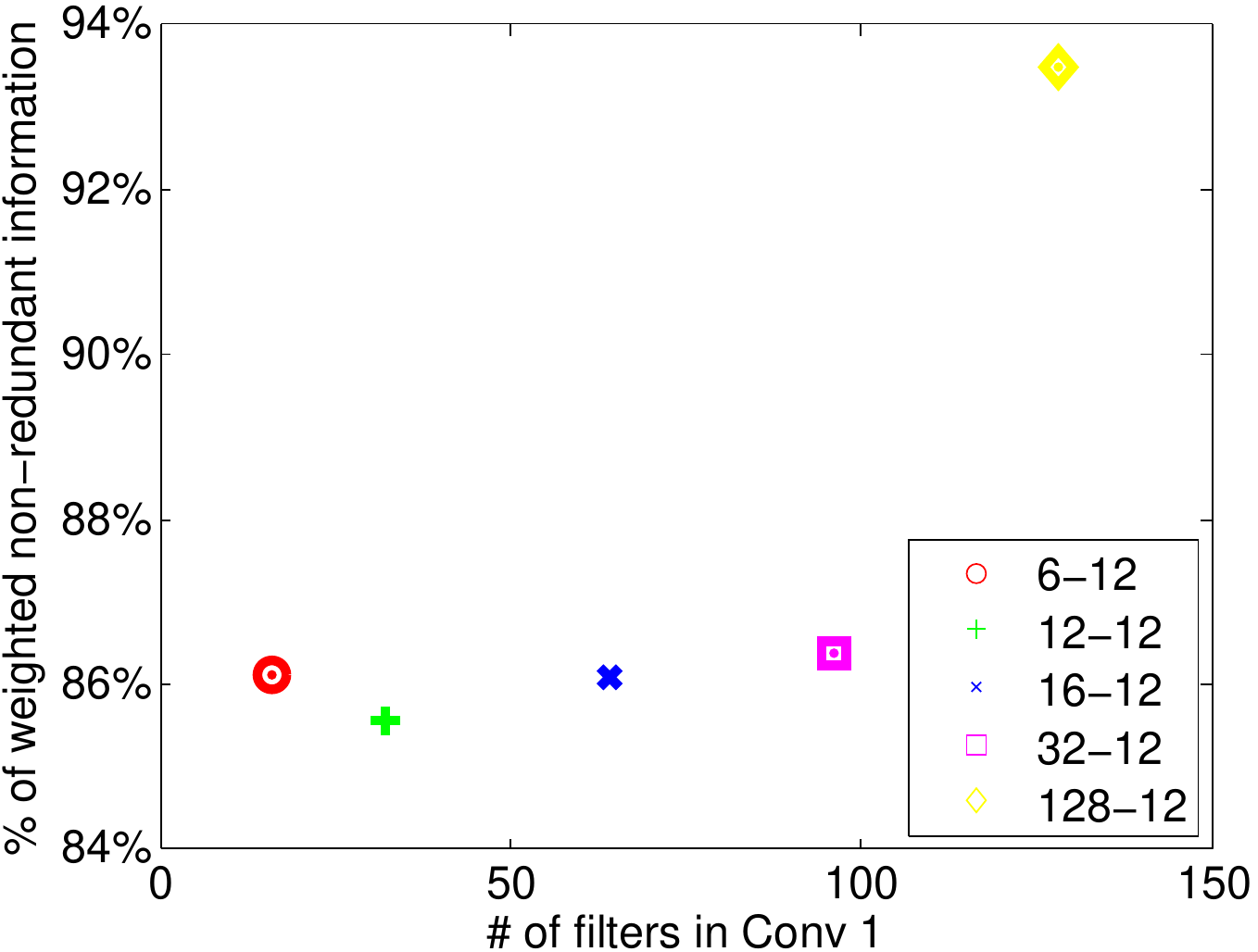}}
\subfigure[Redundancy-Synergy trade-off. The networks differ in the number of filters in Conv.~$2$.] {\includegraphics[width=.23\textwidth]{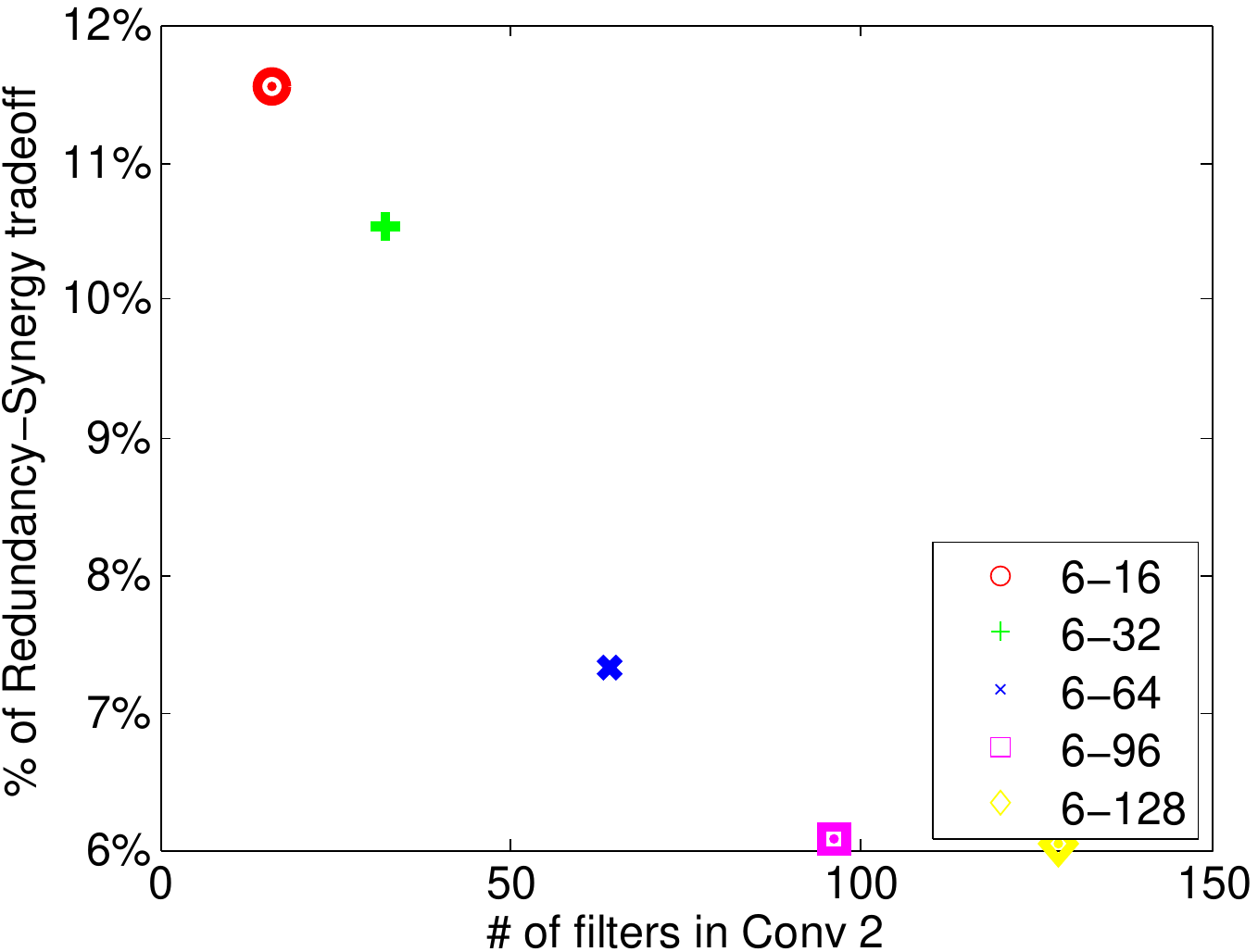}}
\subfigure[Weighted non-redundant information. The networks differ in the number of filters in Conv.~$2$.] {\includegraphics[width=.23\textwidth]{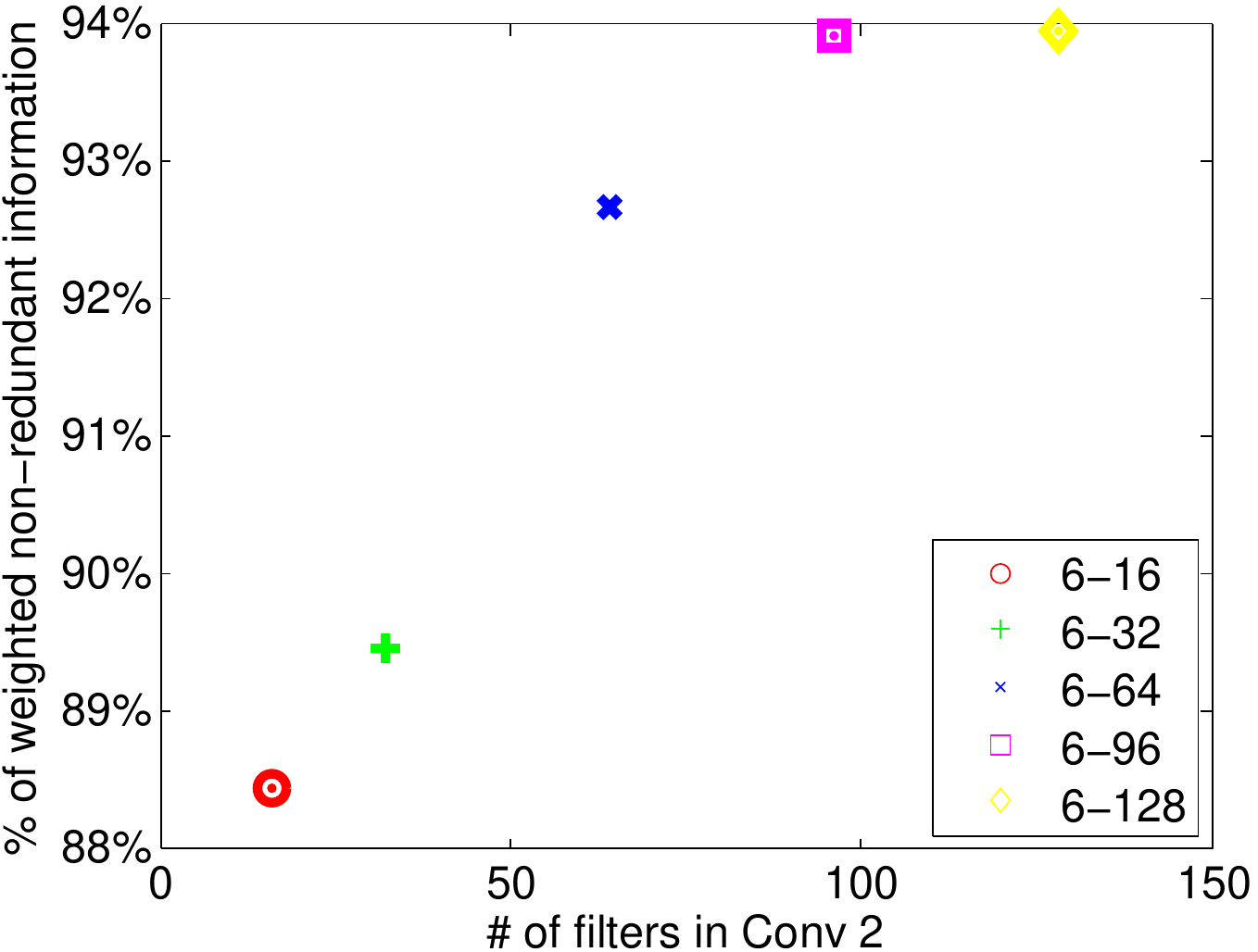}} & \\
\subfigure[Redundancy-Synergy trade-off. The networks differ in the number of filters in Conv.~$1$ and Conv.~$2$.] {\includegraphics[width=.23\textwidth]{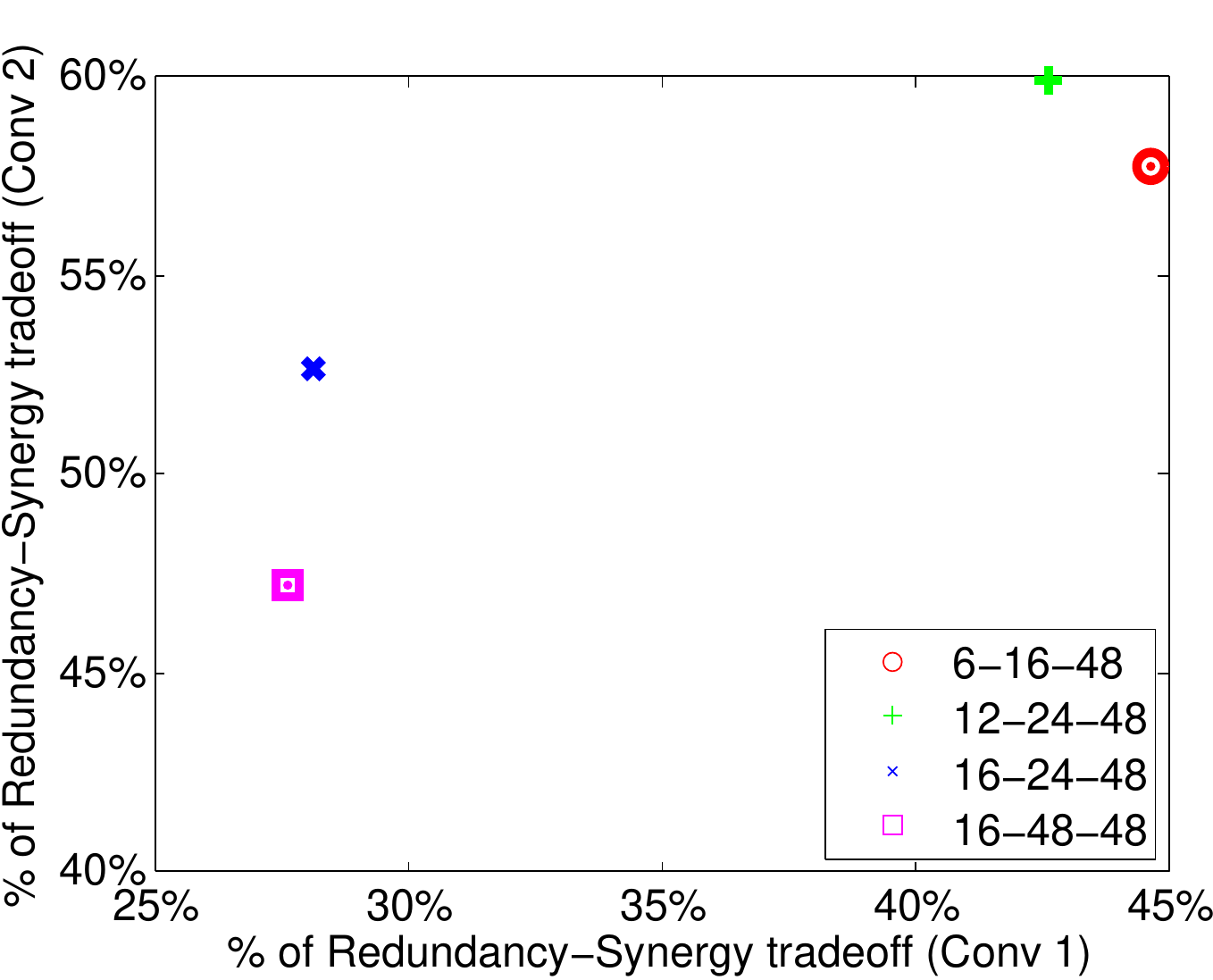}}
\subfigure[Weighted non-redundant information. The networks differ in the number of filters in Conv.~$1$ and Conv.~$2$.] {\includegraphics[width=.23\textwidth]{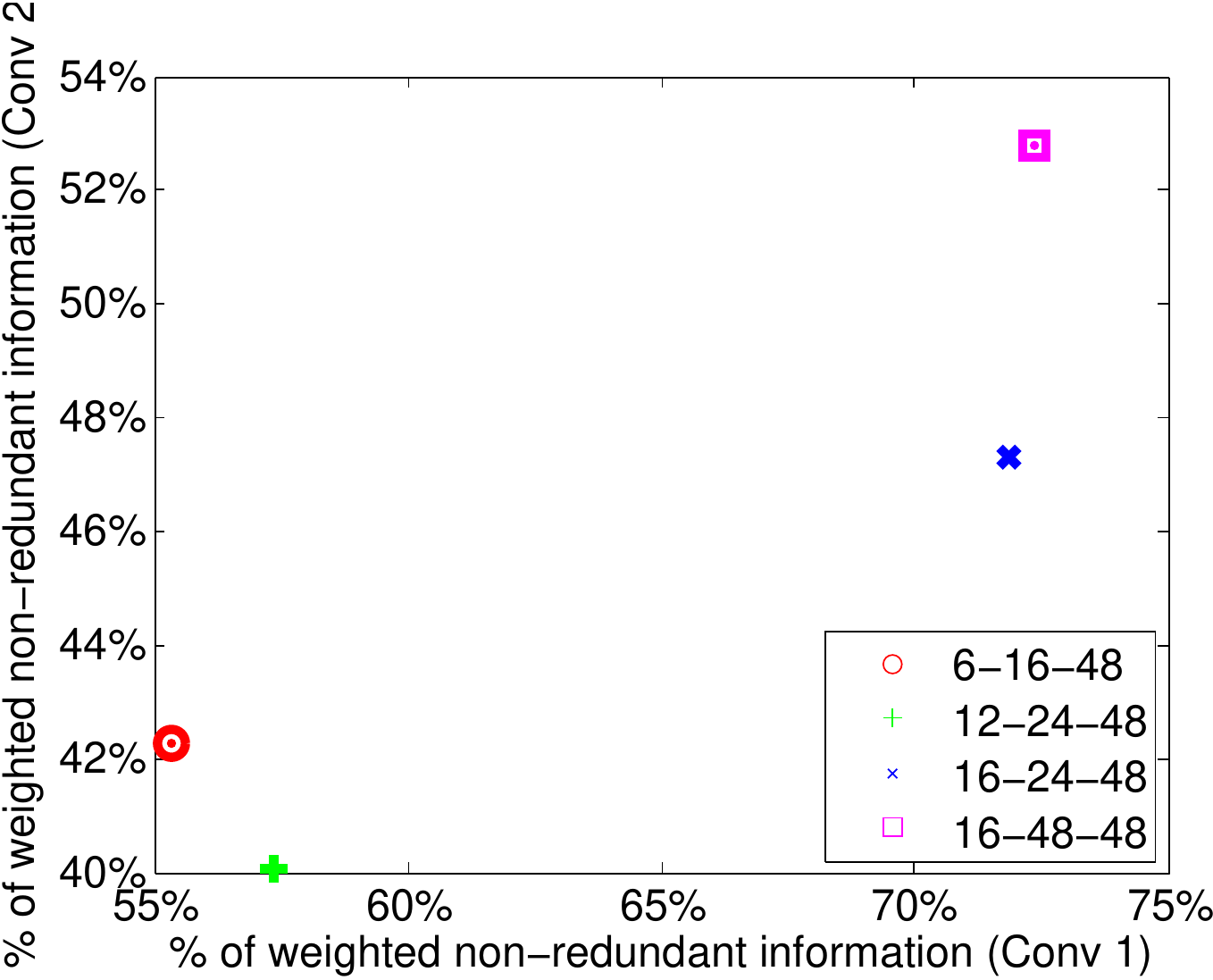}}
\subfigure[Redundancy-Synergy trade-off. The networks differ in the number of filters in Conv.~$1$ and Conv.~$3$.] {\includegraphics[width=.23\textwidth]{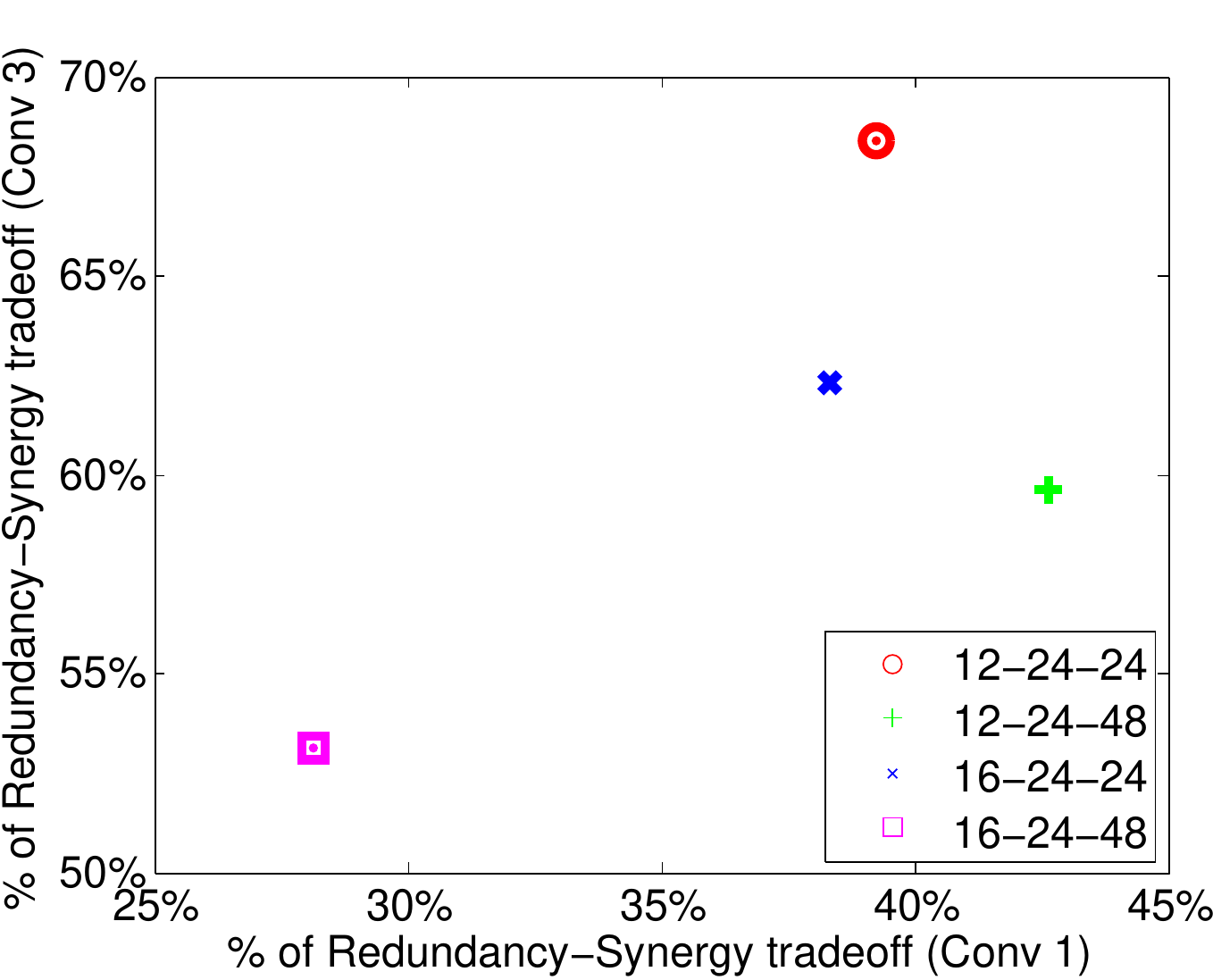}}
\subfigure[Weighted non-redundant information. The networks differ in the number of filters in Conv.~$1$ and Conv.~$3$.] {\includegraphics[width=.23\textwidth]{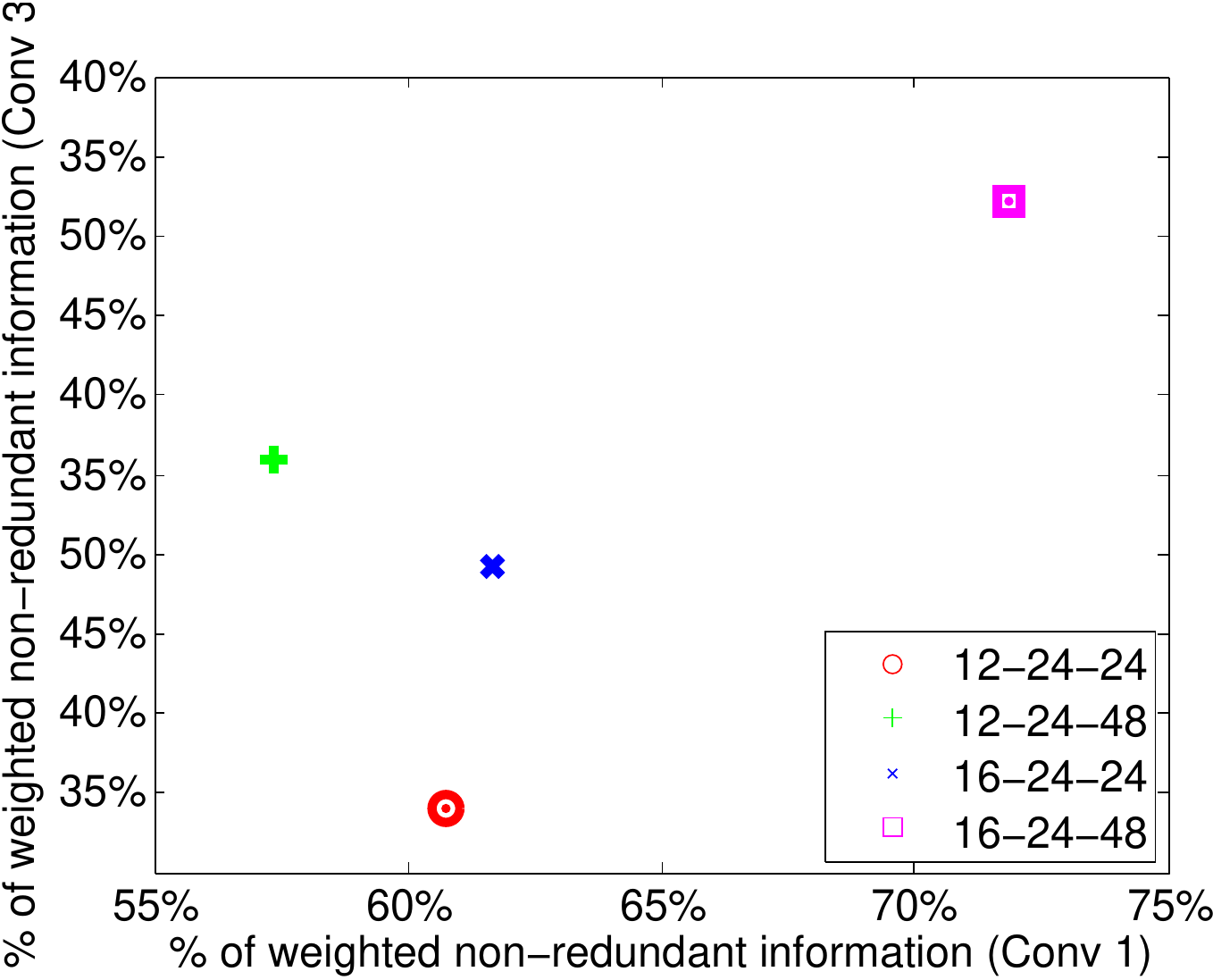}}\\
\end{tabular}
\caption{The redundancy-synergy trade-off and the weighted non-redundant information in MNIST (the first row) and Fruits~$360$ (the second row) data sets. (a) and (b) demonstrate the percentages of these two quantities with respect to different number of filters in Conv.~$1$, but $12$ filters in Conv.~$2$. (c) and (d) demonstrate the percentages of these two quantities with respect to $6$ filters in Conv.~$1$, but different number of filters in Conv.~$2$. Similarly, (e) and (f) compare these two quantities with respect to different number of filters in both Conv.~$1$ and Conv.~$2$, with $48$ filters in Conv.~$3$; whereas (e) and (f) compare these two quantities with respect to different number of filters in both Conv.~$1$ and Conv.~$3$, with $24$ filters in Conv.~$2$. In each subfigure, the topologies of all competing networks are specified in the legend.\vspace{-0.4cm}}
\label{fig:tradeoff}
\end{figure*}

We first evaluate MMI with respect to different CNN topologies in Fig.~\ref{fig:MMI_saturation}. For MNIST, we demonstrate MMI values in the first two convolutional layers (denote Conv.~$1$ and Conv.~$2$). Similarly, for Fruits 360, we demonstrate MMI values in Conv.~$1$, Conv.~$2$ and Conv.~$3$. By DPI, the maximum amount of information that each convolutional layer representation can capture is exactly the entropy of input. As can be seen, with the increase of the number of filters, the total amount of information that each layer captured also increases accordingly. However, it is interesting to see that MMI values are likely to saturate with only a few filters. For example, in Fruits $360$, with only $6$ filters in Conv.~$1$, $12$ filters in Conv.~$2$ and $16$ filters in Conv.~$3$, make MMI values to reach their maximum value $4.7281$ (i.e., the ensemble average entropy across mini-batches) in each layer. More filters will increase classification accuracy at first. However, increasing the number of filters does not guarantee that classification accuracy increases, and might even degrade performance.

We argue that this phenomenon can be explained by the percentage that the redundancy-synergy trade-off or the weighted non-redundant information accounts for the MMI in each pair of feature maps, i.e., $\frac{2}{C(C-1)}\sum_{i=1}^{C}\sum_{j=i+1}^{C} \frac{\mathbf{I}(X;T^i)+\mathbf{I}(X;T^j)-\mathbf{I}(X;\{T^i,T^j\})}{\mathbf{I}(X;\{T^i,T^j\})}$ or \\$\frac{2}{C(C-1)}\sum_{i=1}^{C}\sum_{j=i+1}^{C}\frac{2\times\mathbf{I}(X;\{T^i,T^j\})-\mathbf{I}(X;T^i)-\mathbf{I}(X;T^j)}{\mathbf{I}(X;\{T^i,T^j\})}$. In fact, by referring to Fig.~\ref{fig:tradeoff}, it is obvious that more filters can push the network towards an improved redundancy-synergy trade-off, i.e., the synergy gradually dominates in each pair of feature maps with the increase of filters. That is perhaps one of the main reasons why the increased number of filters can lead to better classification performance, even though the total multivariate mutual information stays the same. However, if we look deeper, it seems that the redundancy-synergy trade-off is always positive, which may suggest that redundancy is always larger than synergy. On the other hand, one should note that the amount of non-redundant information is always less than the MMI (redundancy is non-negative) no matter the number of filters. Therefore, it is impossible to improve the classification performance by blindly increasing the number of filters. This is because the minimum probability of classification error is upper bounded by the MMI expressed in different forms (e.g.,~\cite{hellman1970probability,sason2018arimoto}).




Having illustrated the DPIs and the redundancy-synergy trade-offs, it is easy to summarize some implications concerning the design and training of CNNs. First, as a possible application of DPI in the error backpropagation chain, we suggest to use the DPI as an indicator on where to perform the ``bypass" in the recently proposed Relay backpropagation~\cite{shen2016relay}. Second, the DPIs and the redundancy-synergy trade-off may give some guidelines on the depth and width of CNNs. Intuitively, we need multiple layers to quantify the multi-scale information contained in natural images. However, more layers will lead to severe information loss. The same dilemma applies to the number of filters in convolutional layers: a sufficient number of filters guarantees preservation of input information and the ability to learn a good redundancy-synergy trade-off. However, increasing the number of filters does not always lead to performance gain.

Admittedly, it is hard to give a concrete rule to determine the exact number of filters in one convolutional layer from the current results. We still present a possible solution to shed light on this problem. In fact, if we view each filter as an individual feature extractor, the problem of determining the optimal number of filters turns out to be seeking a stopping criterion for feature selection. Therefore, the number of filters can be determined by monitoring the value of the conditional mutual information (CMI), i.e., $\mathbf{I}(T_r;Y|T_s)$, where $T_s$ and $T_r$ denote respectively the selected and the remaining filters, and $Y$ denotes desired response.
Theoretically, $\mathbf{I}(T_r;Y|T_s)$ is monotonically decreasing if a new filter $t$ is added into $T_s$~\cite{cover2012elements}, but will never reach zero in practice~\cite{vinh2014reconsidering}. Therefore, in order to evaluate the impact of $t$ on $\mathbf{I}(T_r;Y|T_s)$, we can create a random permutation of $t$ (without permuting the corresponding $Y$), denoted $\tilde{t}$. If $\mathbf{I}(\{T_r-t\};Y|\{T_s,t\})$ is not significantly smaller than $\mathbf{I}(\{T_r-\tilde{t}\};Y|\{T_s,\tilde{t}\})$, $t$ can be discarded and the filter selection is stopped. We term this method CMI-permutation~\cite{yu2019simple}. We refer interested readers to Appendix~\ref{appendix_B} for its detailed implementation.

\begin{figure}[!htbp]
\centering
\begin{tabular}{ccc}
\subfigure[$\#$ filters in Conv.~2 of LeNet-5] {\includegraphics[width=.23\textwidth]{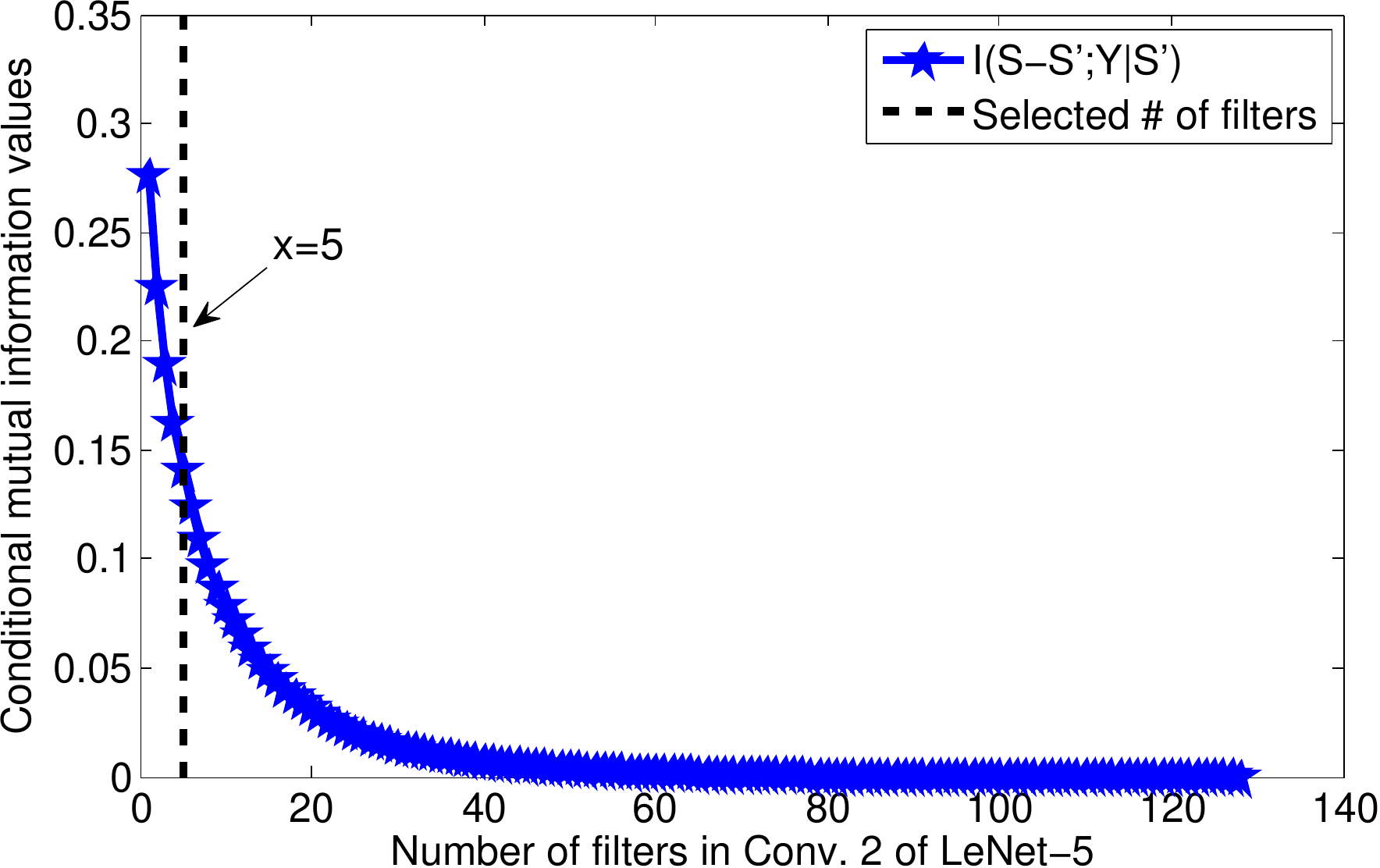}}
\subfigure[$\#$ filters in Conv5-3 of VGG-16] {\includegraphics[width=.23\textwidth]{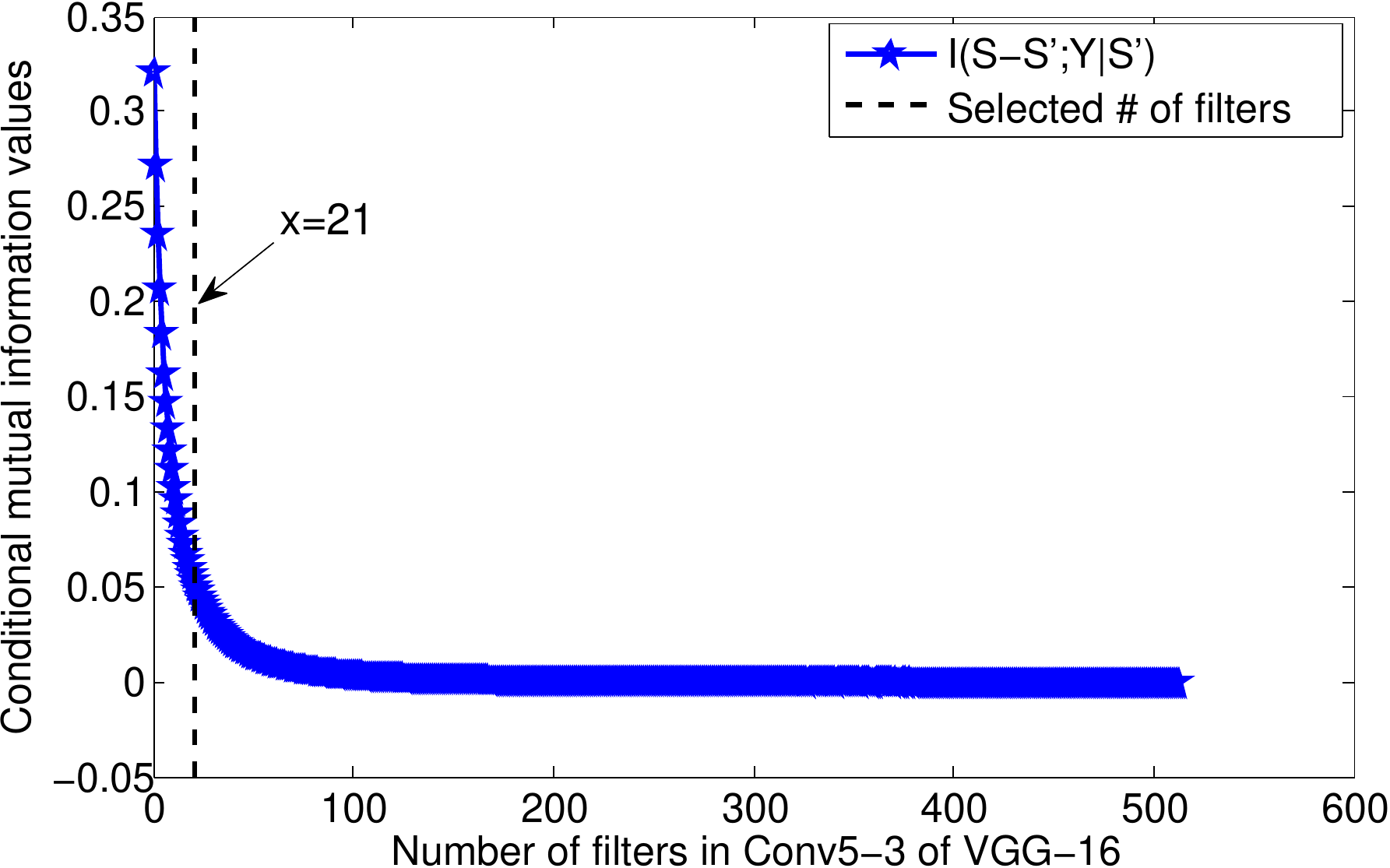}}
\end{tabular}
\caption{Determination of the number of filters in (a) Conv. $2$ of LeNet-5 trained on MNIST data set; and (b) Conv5-3 of VGG-16 trained on CIFAR-10 data set. CMI-permutation~\cite{yu2019simple} suggests $5$ filters among a total of $128$ filters in case (a), and $21$ filters among a total of $512$ filters in case (b).
\vspace{-0.4cm}}
\label{fig:number_filters}
\end{figure}

Our preliminary results shown in Fig.~\ref{fig:number_filters} suggest that CMI-permutation is likely to underestimate the number of filters. Therefore, additional design efforts are required as future work.

\vspace{-0.0cm}
\subsection{Revisiting the Information Plane (IP)} \label{sec3.4}

The behaviors of curves in the IP is currently a controversial issue. Recall the discrepancy reported by Saxe $et$ $al$.~\cite{michael2018on}, the existence of compression phase observed by Shwartz-Ziv and Tishby~\cite{shwartz2017opening} depends on the adopted nonlinearity functions: double-sided saturating nonlinearities like ``tanh" or ``sigmoid" yield a compression phase, but linear activation functions and single-sided saturating nonlinearities like the ``ReLU" do not. Interestingly, Noshad $et$ $al$.~\cite{noshad2018scalable} employed dependence graphs to estimate mutual information values and observed the compression phase even using ``ReLU" activation functions.
Similar phenomenon is also observed by Chelombiev $et$ $al$.~\cite{chelombiev2019adaptive}, in which an entropy-based adaptive binning (EBAB) estimator is developed to enable more robust mutual information estimation that adapts to hidden activity of neural networks.
On the other hand, Goldfeld $et$ $al$.~\cite{goldfeld2018estimating} argued that compression is due to layer representations clustering, but it is hard to observe the compression in large network. We disagree with this attribution of different behavior to the nonlinear activation functions. Instead, we often forget that, rarely, estimators share all the properties of the statistically defined quantities~\cite{paninski2003estimation}. Hence, variability in the displayed behavior is mostly likely attributed to different estimators\footnote{Shwartz-Ziv and Tishby~\cite{shwartz2017opening} use the basic Shannon's definition and estimate mutual information by dividing neuron activation values into $30$ equal-interval bins, whereas the base estimator used by Saxe $et$ $al$.~\cite{michael2018on} provides Kernel Density Estimator (KDE) based lower and upper bounds on the true mutual information~\cite{kolchinsky2017estimating,noshad2018scalable}.}, although this argument is rarely invoked in the literature. This is the reason we suggest that a first step before analyzing the information plane curves, is to show that the employed estimators meet the expectation of the DPI (or similar known properties of the statistical quantities). We show above that our R{\'e}nyi's entropy estimator passes this test.

\begin{figure*}[!htbp]
\centering
\begin{tabular}{ccc}
\subfigure[IP, sigmoid ($6$ filters in Conv.~1, $32$ filters in Conv.~2)] {\includegraphics[width=.23\textwidth]{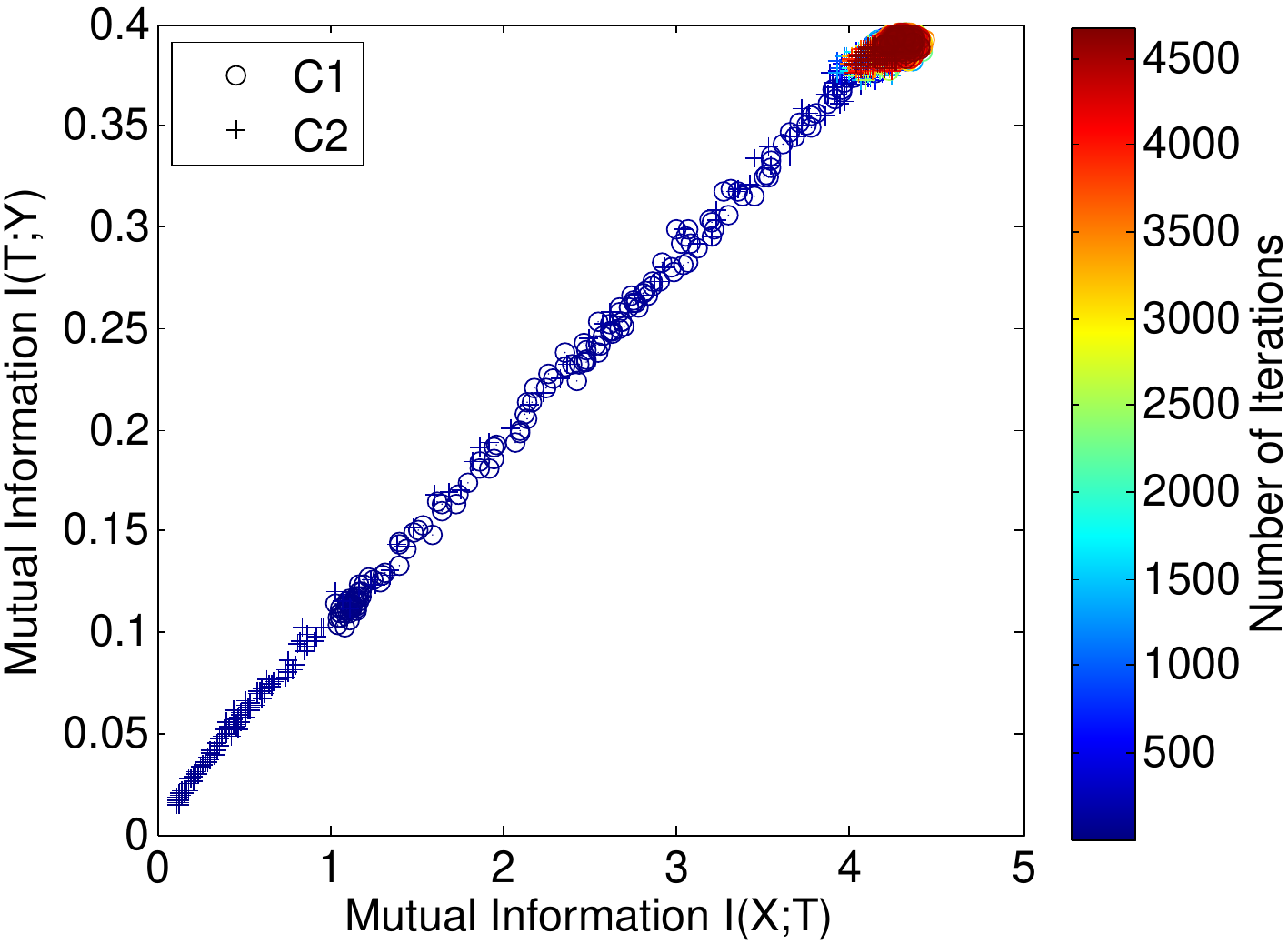}}
\subfigure[M-IP, sigmoid ($6$ filters in Conv.~1, $32$ filters in Conv.~2)] {\includegraphics[width=.23\textwidth]{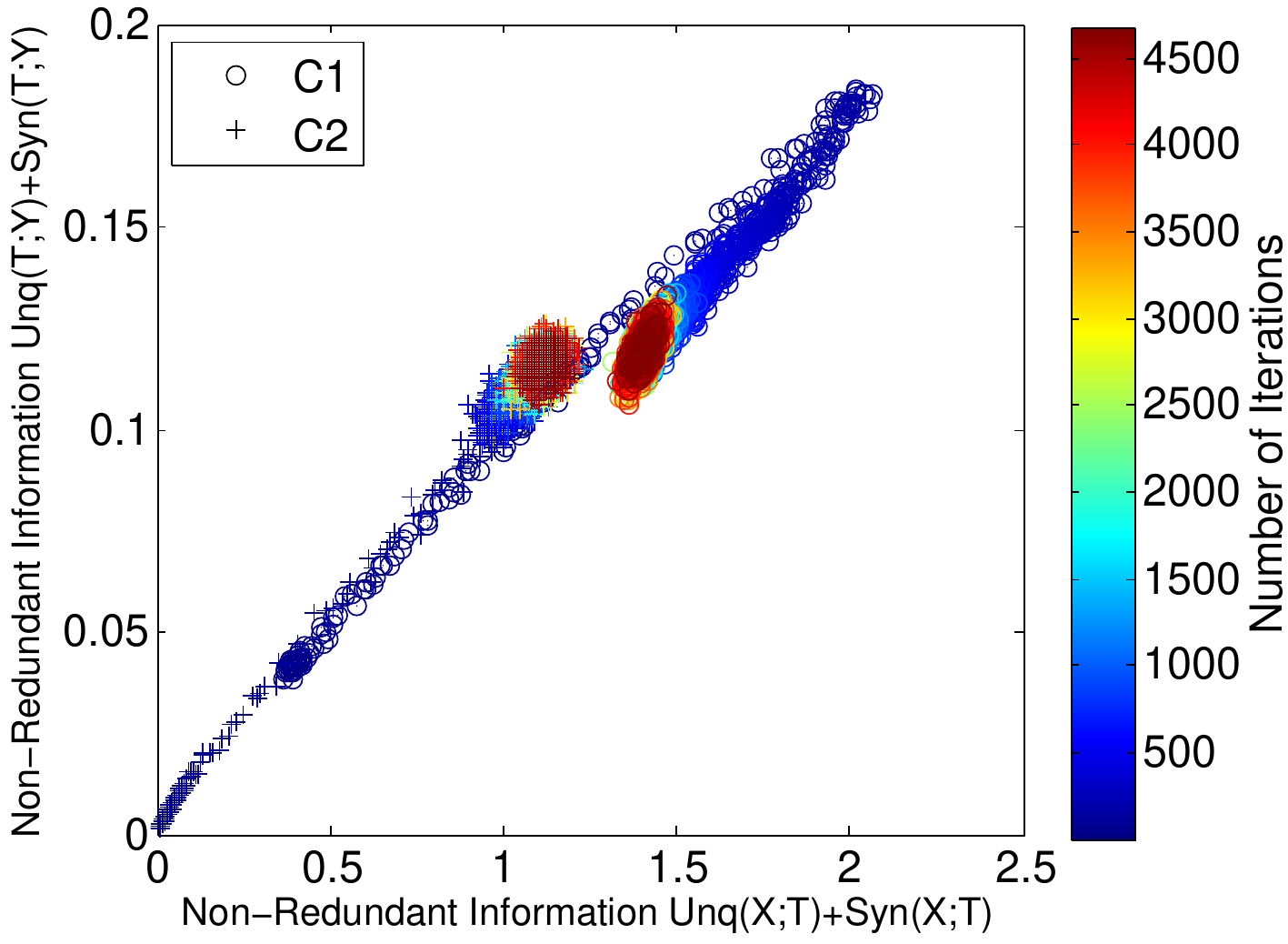}}
\subfigure[M-IP, ReLU ($6$ filters in Conv.~1, $32$ filters in Conv.~2)] {\includegraphics[width=.23\textwidth]{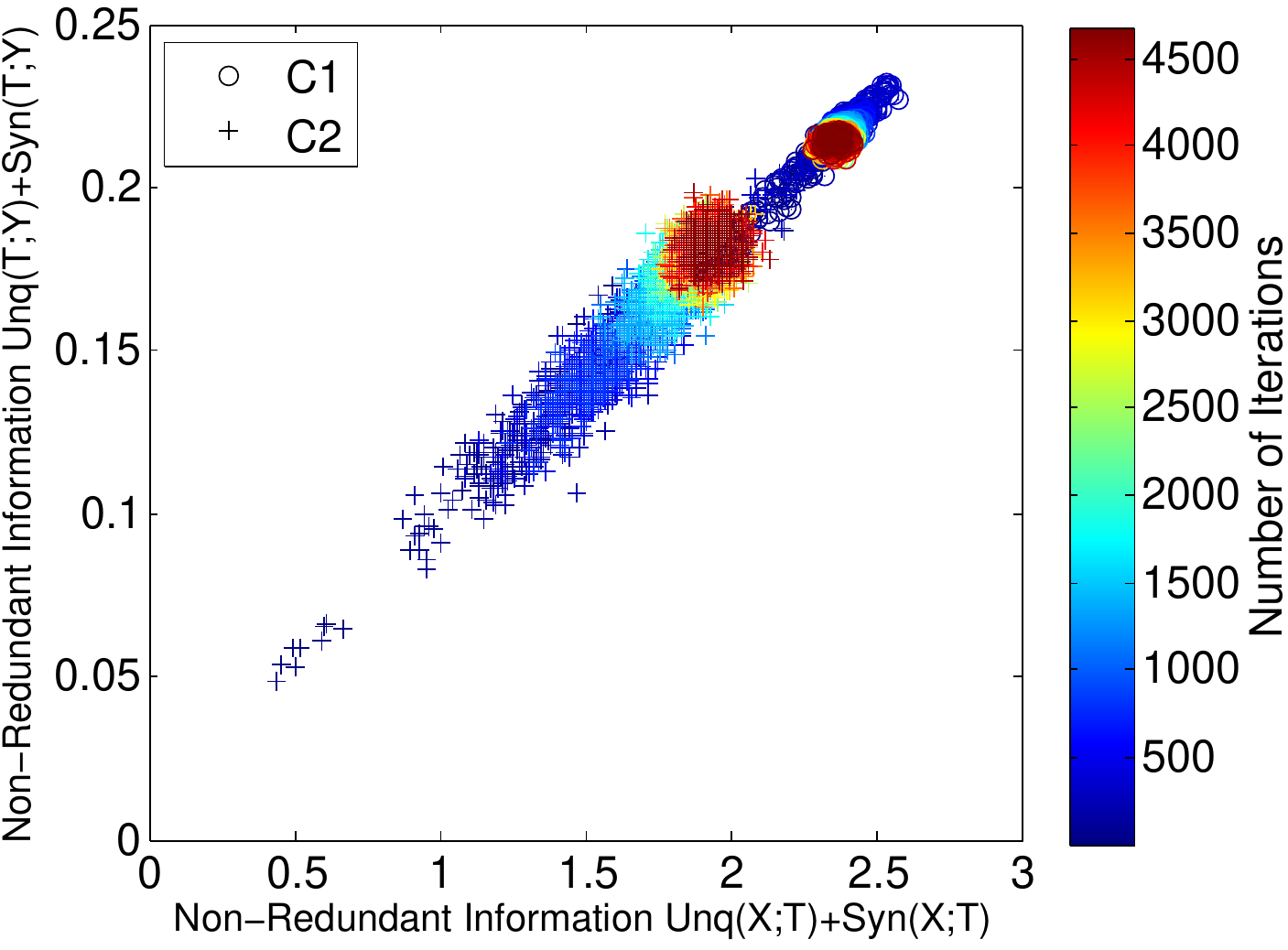}}\\
\subfigure[IP, sigmoid ($6$ filters in Conv.~1, $32$ filters in Conv.~2)] {\includegraphics[width=.23\textwidth]{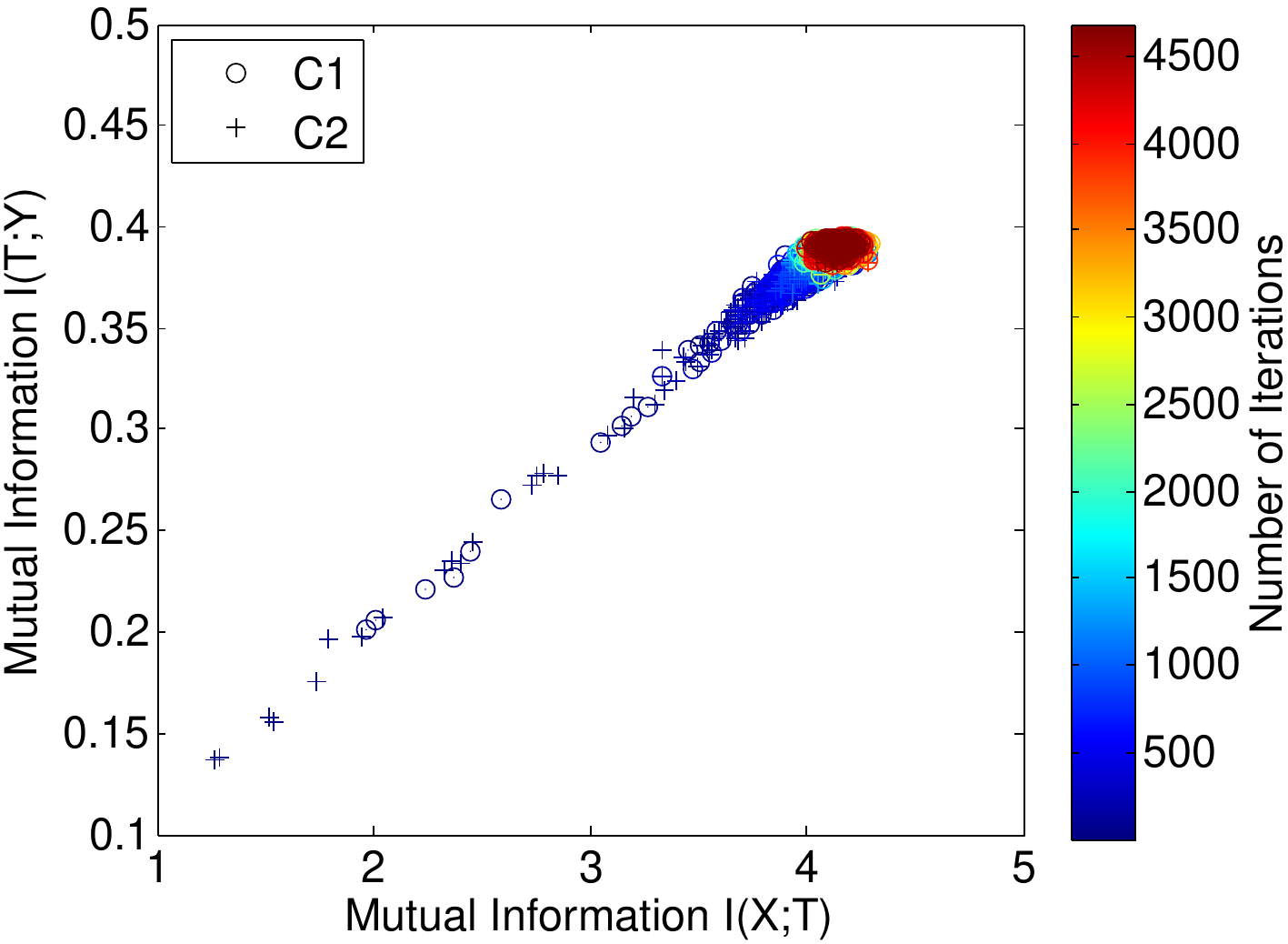}}
\subfigure[M-IP, sigmoid ($6$ filters in Conv.~1, $32$ filters in Conv.~2)] {\includegraphics[width=.23\textwidth]{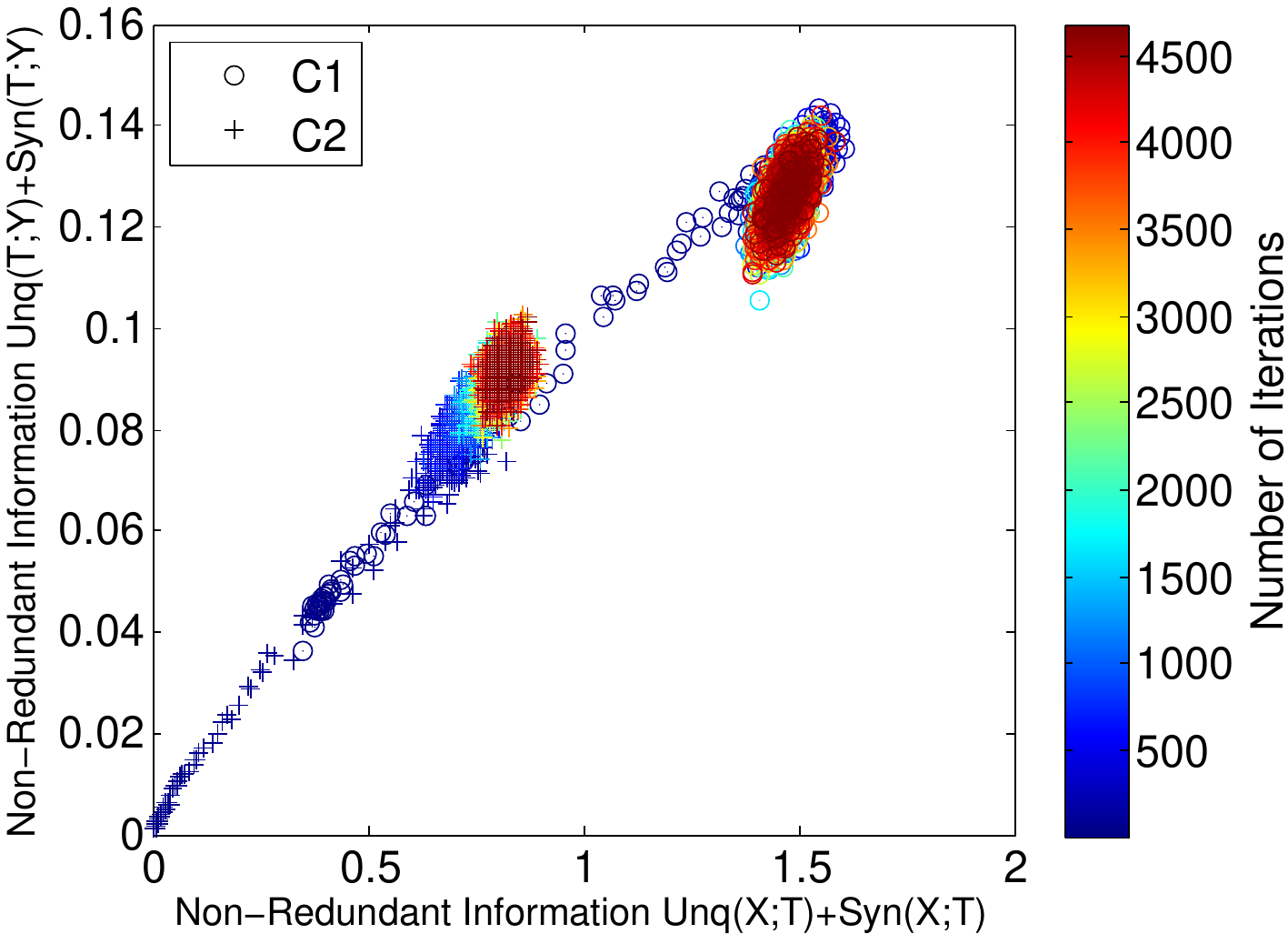}}
\subfigure[M-IP, ReLU ($6$ filters in Conv.~1, $32$ filters in Conv.~2)] {\includegraphics[width=.23\textwidth]{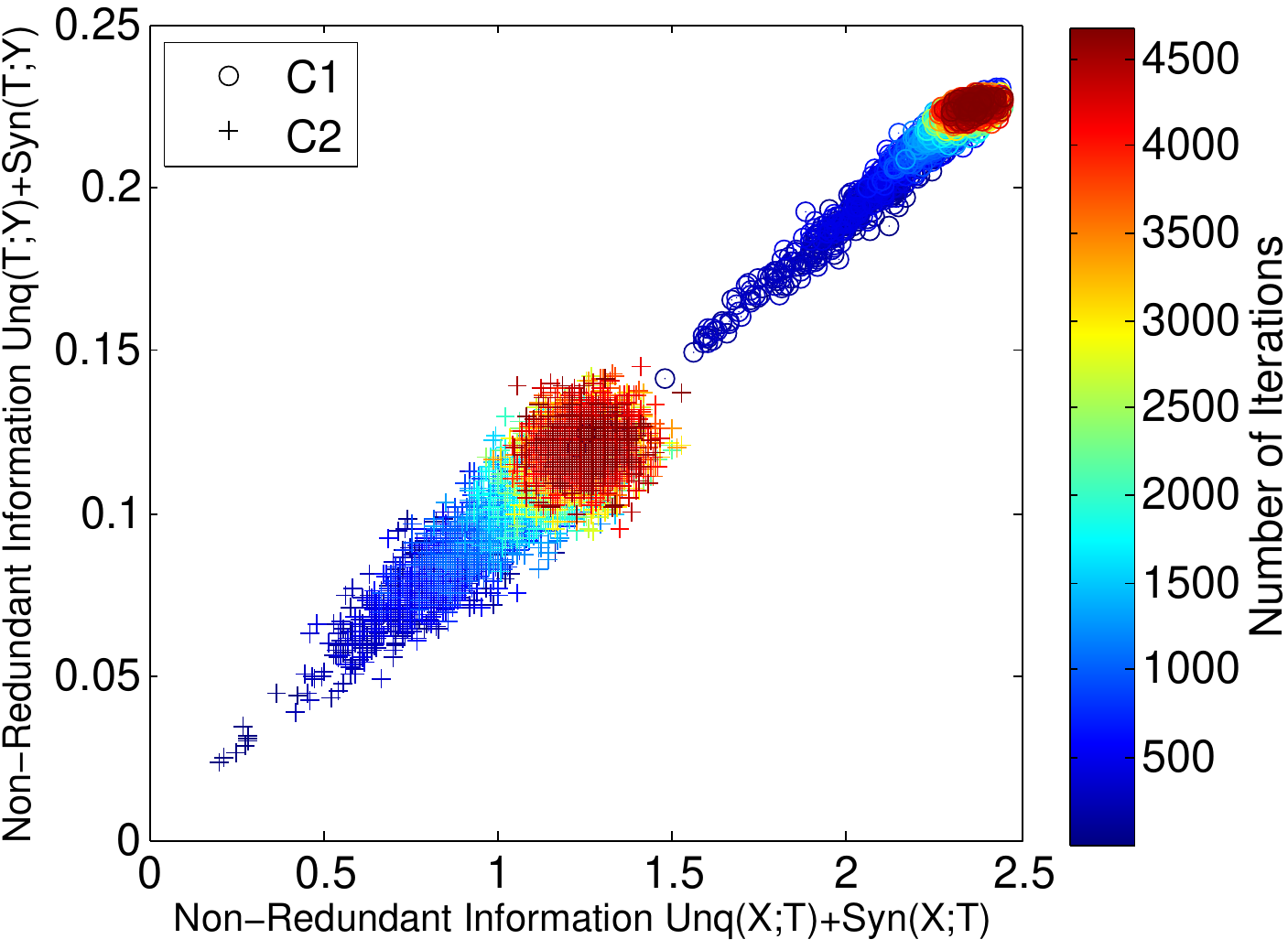}}\\
\end{tabular}
\caption{The Information Plane (IP) and the modified Information Plane (M-IP) of a LeNet-$5$ type CNN trained on MNIST (the first row) and Fashion-MNIST (the second row) data sets. The $\#$ of filters in Conv.~$1$, the $\#$ of filters in Conv.~$2$, and the adopted activation function are indicated in the subtitle of each plot. The curves in IP increase rapidly up to a point without compression (see (a) and (d)). By contrast, it is very easy to observe the compression in M-IP (see (b), (c) and (e)). Moreover, compared with ReLU, sigmoid is more likely to incur the compression (e.g., comparing (b) with (c), or (e) with (f)).\vspace{-0.3cm}}
\label{fig:IP_MNIST}
\end{figure*}



The IPs for a LeNet-5 type CNN trained on MNIST and Fashion-MNIST data sets are shown in Fig.~\ref{fig:IP_MNIST}. From the first column, both $\mathbf{I}(X;T)$ and $\mathbf{I}(T;Y)$ increase rapidly up to a certain point with the SGD iterations. This result conforms to the description in~\cite{goldfeld2018estimating}, suggesting that the behaviour of CNNs in the IP not being the same as that of the MLPs in~\cite{shwartz2017opening,michael2018on,noshad2018scalable} and our intrinsic dimensionality hypothesis in~\cite{yu2018understanding} is specific to SAEs. However, if we remove the redundancy in $\mathbf{I}(X;T)$ and $\mathbf{I}(T;Y)$, and only preserve the unique information and the synergy (i.e., substituting $\mathbf{I}(X;T)$ and $\mathbf{I}(T;Y)$ with their corresponding (average) weighted non-redundant information defined in Section~\ref{sec3.2}), it is easy to observe the compression phase in the modified IP. Moreover, it seems that ``sigmoid" is more likely to incur the compression, compared with ``ReLU", where this intensity can be attributed to the nonlinearity. Our result shed light on the discrepancy in~\cite{shwartz2017opening} and~\cite{michael2018on}, and refined the argument in~\cite{noshad2018scalable,chelombiev2019adaptive}.



\section{Conclusions and Future Work}

This brief presents a systematic method to analyze CNNs mapping and training from an information theoretic perspective. Using the multivariate extension of the matrix-based R{\'e}nyi's $\alpha$-entropy functional, we validated two data processing inequalities (DPIs) in CNNs. The introduction of partial information decomposition (PID) framework enables us to pin down the redundancy-synergy trade-off in layer representations. We also analyzed the behaviors of curves in the information plane, aiming at clarifying the debate on the existence of compression in DNNs. We close by highlighting some potential extensions of our methodology and direction of future research:

1) All the information quantities mentioned in this paper are estimated based on a vector rastering of samples, i.e., each layer input (e.g., an input image, a feature map) is first converted to a single vector before entropy or mutual information estimation. Albeit its simplicity, we distort spatial relationships amongst neighboring pixels. Therefore, a question remains on the reliable information theoretic estimation that is feasible within a tensor structure.

\begin{figure}
\includegraphics[width=0.5\textwidth]{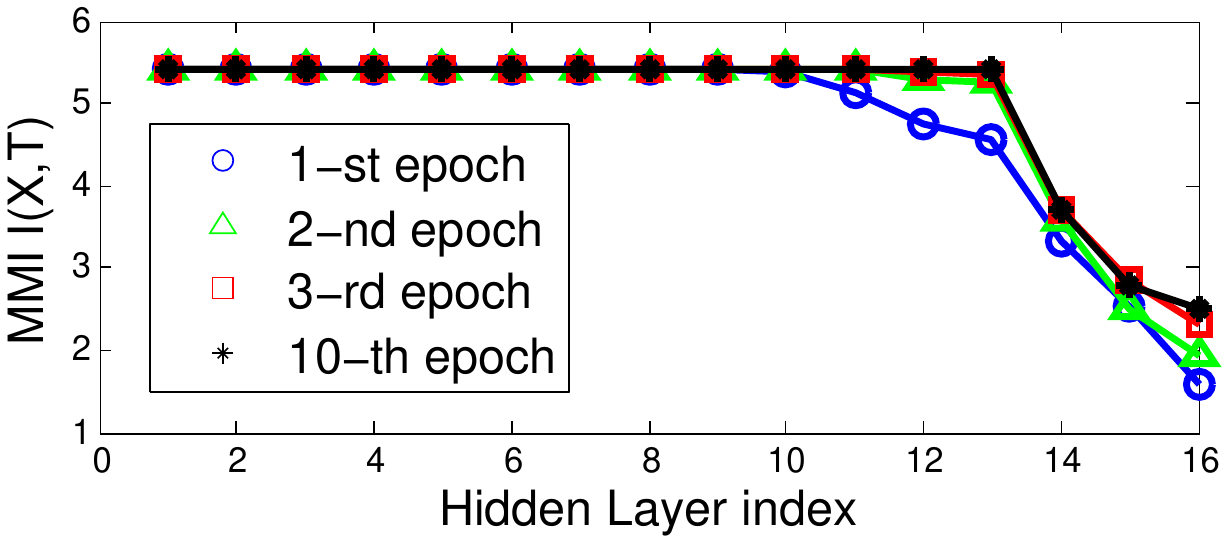}
\caption{DPI in VGG-16 on CIFAR-10: $\mathbf{I}(X,T_1)\geq\mathbf{I}(X,T_2)\geq\cdots\geq\mathbf{I}(X,T_K)$. Layer 1 to Layer 13 are convolutional layers, whereas Layer 14 to Layer 16 are fully-connected layers.\vspace{-0.5cm}} \label{fig_VGG}
\end{figure}

2) We look forward to evaluate our estimators on more complex CNN architectures, such as VGGNet~\cite{simonyan2014very} and ResNet~\cite{he2016deep}. According to our observation, it is easy to validate the DPI and the rapid increase of mutual information (in top layers) in VGG-$16$ on CIFAR-$10$ dataset~\cite{krizhevsky2009learning} (see Fig.~\ref{fig_VGG}). However, it seems that the MMI values in bottom layers are likely to saturate. The problem arises when we try to take the Hadamard product of the kernel matrices of each feature map in Eq.~(\ref{eq8}). The elements in these (normalized) kernel matrices have values between $0$ and $1$, and taking the entrywise product of, e.g., $512$ such matrices like in the convolutional layer of VGG-$16$, will tend towards a matrix with diagonal entries $1/n$ and nearly zero everywhere else. The eigenvalues of the resulting matrix will quickly have almost the same value across training epochs.

3) Our estimator described in Eqs.~(\ref{eq6}) and (\ref{eq7}) has two limitations one needs to be aware of. First, our estimators are highly efficient in the scenario of high-dimensional data, regardless of the data properties (e.g., continuous, discrete, or mixed). However, the computational complexity increases almost cubically with the number of samples $n$, because of the eigenvalue decomposition. Fortunately, it is possible to apply methods such as kernel randomization~\cite{lopez2013randomized} to reduce the burden to $\mathcal{O}(n\log(n))$. By contrast, the well-known kernel density estimator~\cite{moon1995estimation} suffers from the curse of dimensionality~\cite{nagler2016evading}, whereas the $k$-nearest neighbor estimator~\cite{kraskov2004estimating} requires exponentially many samples for accurate estimation~\cite{gao2015efficient}. Second, as we have emphasized in previous work (e.g.,~\cite{yu2018understanding,yu2018multivariate}), it is important to select an appropriate value for the kernel size $\sigma$ and the order $\alpha$ of R{\'e}nyi's entropy functional. Otherwise, spurious conclusions may happen. Reliable manners to select $\sigma$ include the Silverman's rule of thumb and $10$ to $20$ percent of the total (median) range of the Euclidean distances between all pairwise data points~\cite{shi2000normalized,jenssen2009kernel}. On the other hand, the choice of $\alpha$ is associated with the task goal. If the application requires emphasis on tails of the distribution (rare events) or multiple modalities, $\alpha$ should be less than $2$ and possibly approach to $1$ from above. $\alpha=2$ provides neutral weighting~\cite{principe2010information}. Finally, if the goal is to characterize modal behavior, $\alpha$ should be greater than $2$. This work fix $\alpha=1.01$ to approximate Shannon's definition.

\begin{wrapfigure}{r}{0.25\textwidth} 
    \centering
    \includegraphics[width=0.25\textwidth]{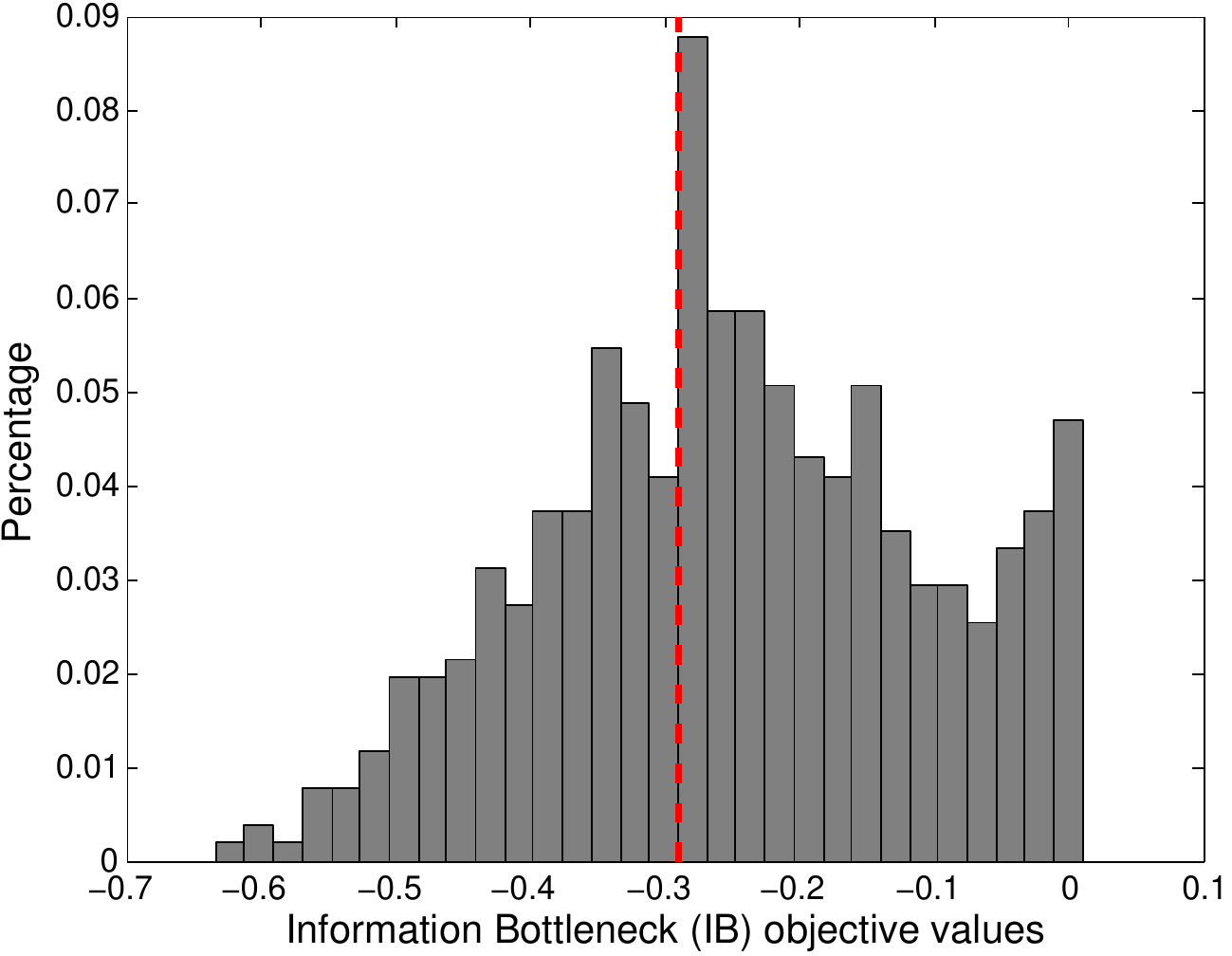}
    \caption{Information Bottleneck (IB) objective values distribution in CONV$5$-$3$ of VGG-$16$ trained on CIFAR-$10$. Nearly $55\%$ filters have IB values above the red dashed line, which imply less importance for classification.}\label{pruning_CIFAR_10}
    \end{wrapfigure}

4) Perhaps one of the most promising applications concerning our estimators is the filter-level pruning for CNNs compression, i.e., discarding whole filters that are less important~\cite{luo2017thinet}. Different statistics (e.g., the absolute weight sum~\cite{li2016pruning} or the Average Percentage of Zeros (APoZ)~\cite{hu2016network}) have been propose to measure filter importance. Moreover, a recent study~\cite{amjad2018understanding} suggests that mutual information is a reliable indicator to measure neuron (or filter) importance. Therefore, given an individual filter $T^i$, we suggest evaluating its importance by the Information Bottleneck (IB)~\cite{tishby2000information} objective $\mathbf{I}(X;T^i)-\beta\mathbf{I}(T^i;Y)$, where $X$ and $Y$ denote respectively the input batch and its corresponding desired output, and $\beta$ is a Lagrange multiplier. Intuitively, a small value of this objective indicates that $T^i$ obtains a compressed representation of $X$ that is relevant to $Y$. Therefore, the smaller the objective value, the higher importance of $T^i$. Fig.~\ref{pruning_CIFAR_10} demonstrates the IB objective ($\beta=2$) values distribution for $512$ filters in CONV$5$-$3$ layer of VGG-$16$ trained on CIFAR-$10$. This distribution looks similar to the one obtained by APoZ in~\cite{hu2016network}, both of them indicate that more than $50\%$ of filters are less important and can be discarded with negligible loss in accuracy~\cite{li2016pruning}.


%
%

\bibliographystyle{IEEEtran}
\bibliography{samplepaper}

\begin{thebibliography}{10}
\providecommand{\url}[1]{#1}
\csname url@samestyle\endcsname
\providecommand{\newblock}{\relax}
\providecommand{\bibinfo}[2]{#2}
\providecommand{\BIBentrySTDinterwordspacing}{\spaceskip=0pt\relax}
\providecommand{\BIBentryALTinterwordstretchfactor}{4}
\providecommand{\BIBentryALTinterwordspacing}{\spaceskip=\fontdimen2\font plus
\BIBentryALTinterwordstretchfactor\fontdimen3\font minus
  \fontdimen4\font\relax}
\providecommand{\BIBforeignlanguage}[2]{{%
\expandafter\ifx\csname l@#1\endcsname\relax
\typeout{** WARNING: IEEEtran.bst: No hyphenation pattern has been}%
\typeout{** loaded for the language `#1'. Using the pattern for}%
\typeout{** the default language instead.}%
\else
\language=\csname l@#1\endcsname
\fi
#2}}
\providecommand{\BIBdecl}{\relax}
\BIBdecl

\bibitem{tishby2015deep}
N.~Tishby and N.~Zaslavsky, ``Deep learning and the information bottleneck
  principle,'' in \emph{IEEE ITW}, 2015, pp. 1--5.

\bibitem{achille2017emergence}
A.~Achille and S.~Soatto, ``Emergence of invariance and disentanglement in deep
  representations,'' \emph{JMLR}, vol.~19, no.~1, pp. 1947--1980, 2018.

\bibitem{tax2017partial}
T.~Tax, P.~A. Mediano, and M.~Shanahan, ``The partial information decomposition
  of generative neural network models,'' \emph{Entropy}, vol.~19, no.~9, p.
  474, 2017.

\bibitem{shwartz2017opening}
R.~Shwartz-Ziv and N.~Tishby, ``Opening the black box of deep neural networks
  via information,'' \emph{arXiv preprint arXiv:1703.00810}, 2017.

\bibitem{michael2018on}
A.~M. Saxe \emph{et~al.}, ``On the information bottleneck theory of deep
  learning,'' in \emph{ICLR}, 2018.

\bibitem{yu2018understanding}
S.~Yu and J.~C. Principe, ``Understanding autoencoders with information
  theoretic concepts,'' \emph{Neural Networks}, vol. 117, pp. 104--123, 2019.

\bibitem{giraldo2015measures}
L.~G. Sanchez~Giraldo, M.~Rao, and J.~C. Principe, ``Measures of entropy from
  data using infinitely divisible kernels,'' \emph{IEEE Transactions on
  Information Theory}, vol.~61, no.~1, pp. 535--548, 2015.

\bibitem{camastra2016intrinsic}
F.~Camastra and A.~Staiano, ``Intrinsic dimension estimation: Advances and open
  problems,'' \emph{Information Sciences}, vol. 328, pp. 26--41, 2016.

\bibitem{brown2012conditional}
G.~Brown, A.~Pocock, M.-J. Zhao, and M.~Luj{\'a}n, ``Conditional likelihood
  maximisation: a unifying framework for information theoretic feature
  selection,'' \emph{JMLR}, vol.~13, no. Jan, pp. 27--66, 2012.

\bibitem{yu2018multivariate}
S.~Yu, L.~G. Sanchez~Giraldo, R.~Jenssen, and J.~C. Principe, ``Multivariate
  extension of matrix-based renyi's $\alpha$-order entropy functional,''
  \emph{IEEE Transactions on Pattern Analysis and Machine Intelligence}, 2019.

\bibitem{williams2010nonnegative}
P.~L. Williams and R.~D. Beer, ``Nonnegative decomposition of multivariate
  information,'' \emph{arXiv preprint arXiv:1004.2515}, 2010.

\bibitem{renyi1961measures}
A.~R{\'e}nyi, ``On measures of entropy and information,'' in \emph{Proc. of the
  4th Berkeley Sympos. on Math. Statist. and Prob.}, vol.~1, 1961, pp.
  547--561.

\bibitem{principe2010information}
J.~C. Principe, \emph{Information theoretic learning: Renyi's entropy and
  kernel perspectives}.\hskip 1em plus 0.5em minus 0.4em\relax Springer Science
  \& Business Media, 2010.

\bibitem{muller2013quantum}
M.~M{\"u}ller-Lennert, F.~Dupuis, O.~Szehr, S.~Fehr, and M.~Tomamichel, ``On
  quantum r{\'e}nyi entropies: A new generalization and some properties,''
  \emph{J. Math. Phys.}, vol.~54, no.~12, p. 122203, 2013.

\bibitem{bhatia2006infinitely}
R.~Bhatia, ``Infinitely divisible matrices,'' \emph{The American Mathematical
  Monthly}, vol. 113, no.~3, pp. 221--235, 2006.

\bibitem{yeung1991new}
R.~W. Yeung, ``A new outlook on shannon's information measures,'' \emph{IEEE
  transactions on information theory}, vol.~37, no.~3, pp. 466--474, 1991.

\bibitem{lecun1998gradient}
Y.~LeCun, L.~Bottou, Y.~Bengio, and P.~Haffner, ``Gradient-based learning
  applied to document recognition,'' \emph{Proceedings of the IEEE}, vol.~86,
  no.~11, pp. 2278--2324, 1998.

\bibitem{xiao2017fashion}
H.~Xiao, K.~Rasul, and R.~Vollgraf, ``Fashion-mnist: a novel image dataset for
  benchmarking machine learning algorithms,'' \emph{arXiv preprint
  arXiv:1708.07747}, 2017.

\bibitem{thoma2017hasyv2}
M.~Thoma, ``The hasyv2 dataset,'' \emph{arXiv preprint arXiv:1701.08380}, 2017.

\bibitem{murecsan2018fruit}
H.~Mure{\c{s}}an and M.~Oltean, ``Fruit recognition from images using deep
  learning,'' \emph{Acta Universitatis Sapientiae, Informatica}, vol.~10,
  no.~1, pp. 26--42, 2018.

\bibitem{krizhevsky2012imagenet}
A.~Krizhevsky, I.~Sutskever, and G.~E. Hinton, ``Imagenet classification with
  deep convolutional neural networks,'' in \emph{NeurIPS}, 2012, pp.
  1097--1105.

\bibitem{silverman1986density}
B.~W. Silverman, \emph{Density estimation for statistics and data
  analysis}.\hskip 1em plus 0.5em minus 0.4em\relax CRC press, 1986, vol.~26.

\bibitem{bertschinger2014quantifying}
N.~Bertschinger, J.~Rauh, E.~Olbrich, J.~Jost, and N.~Ay, ``Quantifying unique
  information,'' \emph{Entropy}, vol.~16, no.~4, pp. 2161--2183, 2014.

\bibitem{griffith2014quantifying}
V.~Griffith and C.~Koch, ``Quantifying synergistic mutual information,'' in
  \emph{Guided Self-Organization: Inception}.\hskip 1em plus 0.5em minus
  0.4em\relax Springer, 2014, pp. 159--190.

\bibitem{bell2003co}
A.~J. Bell, ``The co-information lattice,'' in \emph{Proceedings of the Fifth
  International Workshop on Independent Component Analysis and Blind Signal
  Separation: ICA}, vol. 2003, 2003.

\bibitem{timme2014synergy}
N.~Timme, W.~Alford, B.~Flecker, and J.~M. Beggs, ``Synergy, redundancy, and
  multivariate information measures: an experimentalist's perspective,''
  \emph{J. Comput. Neurosci.}, vol.~36, no.~2, pp. 119--140, 2014.

\bibitem{hellman1970probability}
M.~Hellman and J.~Raviv, ``Probability of error, equivocation, and the chernoff
  bound,'' \emph{IEEE Transactions on Information Theory}, vol.~16, no.~4, pp.
  368--372, 1970.

\bibitem{sason2018arimoto}
I.~Sason and S.~Verd{\'u}, ``Arimoto--r{\'e}nyi conditional entropy and
  bayesian $ m $-ary hypothesis testing,'' \emph{IEEE Transactions on
  Information Theory}, vol.~64, no.~1, pp. 4--25, 2018.

\bibitem{shen2016relay}
L.~Shen and Q.~Huang, ``Relay backpropagation for effective learning of deep
  convolutional neural networks,'' in \emph{ECCV}, 2016, pp. 467--482.

\bibitem{cover2012elements}
T.~M. Cover and J.~A. Thomas, \emph{Elements of information theory}.\hskip 1em
  plus 0.5em minus 0.4em\relax John Wiley \& Sons, 2012.

\bibitem{vinh2014reconsidering}
N.~X. Vinh, J.~Chan, and J.~Bailey, ``Reconsidering mutual information based
  feature selection: A statistical significance view,'' in \emph{AAAI}, 2014.

\bibitem{yu2019simple}
S.~Yu and J.~C. Pr{\'\i}ncipe, ``Simple stopping criteria for information
  theoretic feature selection,'' \emph{Entropy}, vol.~21, no.~1, p.~99, 2019.

\bibitem{noshad2018scalable}
M.~Noshad, Y.~Zeng, and A.~O. Hero, ``Scalable mutual information estimation
  using dependence graphs,'' in \emph{ICASSP}, 2019, pp. 2962--2966.

\bibitem{chelombiev2019adaptive}
I.~Chelombiev, C.~Houghton, and C.~O'Donnell, ``Adaptive estimators show
  information compression in deep neural networks,'' \emph{arXiv preprint
  arXiv:1902.09037}, 2019.

\bibitem{goldfeld2018estimating}
Z.~Goldfeld \emph{et~al.}, ``Estimating information flow in neural networks,''
  \emph{arXiv preprint arXiv:1810.05728}, 2018.

\bibitem{paninski2003estimation}
L.~Paninski, ``Estimation of entropy and mutual information,'' \emph{Neural
  computation}, vol.~15, no.~6, pp. 1191--1253, 2003.

\bibitem{kolchinsky2017estimating}
A.~Kolchinsky and B.~Tracey, ``Estimating mixture entropy with pairwise
  distances,'' \emph{Entropy}, vol.~19, no.~7, p. 361, 2017.

\bibitem{simonyan2014very}
K.~Simonyan and A.~Zisserman, ``Very deep convolutional networks for
  large-scale image recognition,'' in \emph{ICLR}, 2015.

\bibitem{he2016deep}
K.~He, X.~Zhang, S.~Ren, and J.~Sun, ``Deep residual learning for image
  recognition,'' in \emph{CVPR}, 2016, pp. 770--778.

\bibitem{krizhevsky2009learning}
A.~Krizhevsky and G.~Hinton, ``Learning multiple layers of features from tiny
  images,'' Citeseer, Tech. Rep., 2009.

\bibitem{lopez2013randomized}
D.~Lopez-Paz, P.~Hennig, and B.~Sch{\"o}lkopf, ``The randomized dependence
  coefficient,'' in \emph{NeurIPS}, 2013, pp. 1--9.

\bibitem{moon1995estimation}
Y.-I. Moon, B.~Rajagopalan, and U.~Lall, ``Estimation of mutual information
  using kernel density estimators,'' \emph{Physical Review E}, vol.~52, no.~3,
  p. 2318, 1995.

\bibitem{nagler2016evading}
T.~Nagler and C.~Czado, ``Evading the curse of dimensionality in nonparametric
  density estimation with simplified vine copulas,'' \emph{Journal of
  Multivariate Analysis}, vol. 151, pp. 69--89, 2016.

\bibitem{kraskov2004estimating}
A.~Kraskov, H.~St{\"o}gbauer, and P.~Grassberger, ``Estimating mutual
  information,'' \emph{Physical review E}, vol.~69, no.~6, p. 066138, 2004.

\bibitem{gao2015efficient}
S.~Gao, G.~Ver~Steeg, and A.~Galstyan, ``Efficient estimation of mutual
  information for strongly dependent variables,'' in \emph{AISTATS}, 2015, pp.
  277--286.

\bibitem{shi2000normalized}
J.~Shi and J.~Malik, ``Normalized cuts and image segmentation,'' \emph{IEEE
  Transactions on Pattern Analysis and Machine Intelligence}, vol.~22, no.~8,
  pp. 888--905, 2000.

\bibitem{jenssen2009kernel}
R.~Jenssen, ``Kernel entropy component analysis,'' \emph{IEEE Transactions on
  Pattern Analysis and Machine Intelligence}, vol.~32, no.~5, pp. 847--860,
  2009.

\bibitem{luo2017thinet}
J.-H. Luo, J.~Wu, and W.~Lin, ``Thinet: A filter level pruning method for deep
  neural network compression,'' in \emph{ICCV}, 2017, pp. 5058--5066.

\bibitem{li2016pruning}
H.~Li, A.~Kadav, I.~Durdanovic, H.~Samet, and H.~P. Graf, ``Pruning filters for
  efficient convnets,'' in \emph{ICLR}, 2017.

\bibitem{hu2016network}
H.~Hu, R.~Peng, Y.-W. Tai, and C.-K. Tang, ``Network trimming: A data-driven
  neuron pruning approach towards efficient deep architectures,'' \emph{arXiv
  preprint arXiv:1607.03250}, 2016.

\bibitem{amjad2018understanding}
R.~A. Amjad, K.~Liu, and B.~C. Geiger, ``Understanding individual neuron
  importance using information theory,'' \emph{arXiv preprint
  arXiv:1804.06679}, 2018.

\bibitem{tishby2000information}
N.~Tishby, F.~C. Pereira, and W.~Bialek, ``The information bottleneck method,''
  \emph{arXiv preprint physics/0004057}, 2000.

\bibitem{koller1996toward}
D.~Koller and M.~Sahami, ``Toward optimal feature selection,'' Stanford
  InfoLab, Tech. Rep., 1996.

\bibitem{yaramakala2005speculative}
S.~Yaramakala and D.~Margaritis, ``Speculative markov blanket discovery for
  optimal feature selection,'' in \emph{ICDM}, 2005, pp. 809--812.

\bibitem{vergara2014review}
J.~R. Vergara and P.~A. Est{\'e}vez, ``A review of feature selection methods
  based on mutual information,'' \emph{Neural computing and applications},
  vol.~24, no.~1, pp. 175--186, 2014.

\end{thebibliography}

\appendix
\subsection{Partial Information Diagram}\label{appendix_A}
\begin{figure}[!htbp]
\centering
\includegraphics[width=.45\textwidth]{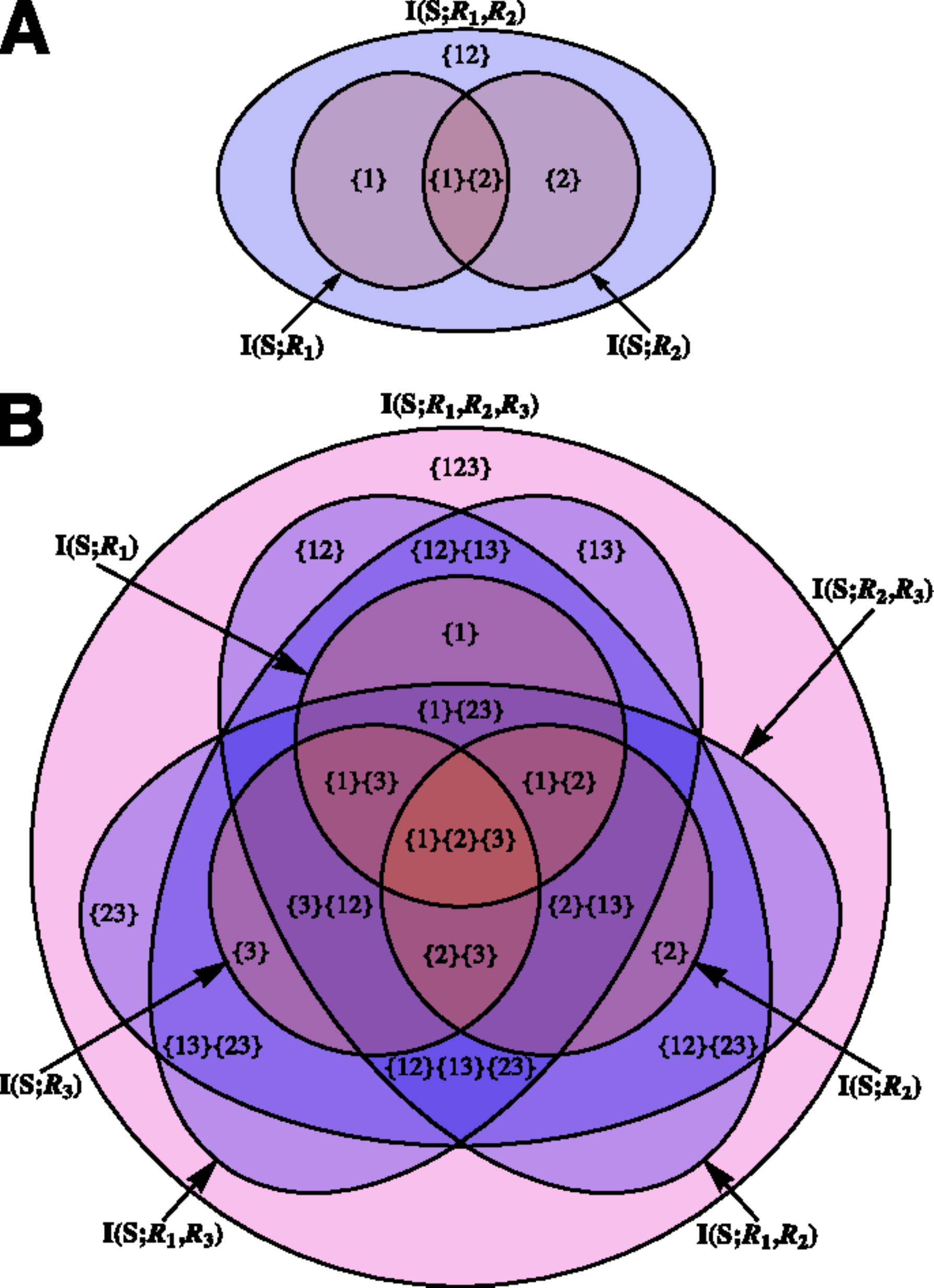}
\caption{Partial information diagrams for (a) $3$ and (b) $4$ variables. For brevity, the sets are abbreviated by the indices of their elements; that is, $\{R_1,R_2\}$ is abbreviated by $\{12\}$, and so on. Figure by Williams and Beer~\cite{williams2010nonnegative}.
\vspace{-0.0cm}}
\label{fig:PID_diagram}
\end{figure}

In this section, we briefly recall the ideas behind the partial information decomposition (PID) by Williams and Beer~\cite{williams2010nonnegative}. The PID is a framework to define information decompositions of arbitrarily many random variables, which can be characterized by a partial information (PI) diagram. The general structure of PI diagram becomes clear when we consider the decomposition for four variables (see Fig.~\ref{fig:PID_diagram}), which illustrates the way in which the total information that $R=\{R_1,R_2,R_3\}$ provides about $S$ is distributed across various combinations of sources.

Specifically, each element in $R$ can provide unique information (regions labeled $\{1\}$, $\{2\}$, and $\{3\}$), information redundantly with one other variable ($\{1\}\{2\}$, $\{1\}\{3\}$, and $\{2\}\{3\}$), or information synergistically with one other variable ($\{12\}$, $\{13\}$, and $\{23\}$). Additionally, information can be provided redundantly by all three variables ($\{1\}\{2\}\{3\}$) or provided by their three-way synergy ($\{123\}$). More interesting are the new kinds of terms representing combinations of redundancy and synergy. For instance, the regions marked $\{1\}\{23\}$, $\{2\}\{13\}$, and $\{3\}\{12\}$ represent information that is available redundantly from either one variable considered individually or the other two considered together. In general, each PI item represents the redundancy of synergies for a particular collection of sources, corresponding to one distinct way for the components of $R=\{R_1,R_2,R_3\}$ to contribute information about $S$. Unfortunately, the number of PI items grows exponentially with the increase of the number of elements in $R$. For example, if $R=\{R_1,R_2,R_3,R_4\}$, there will be $166$ individual non-negative items. Moreover, the reliable estimation of each PID term still remains a big challenge~\cite{bertschinger2014quantifying,griffith2014quantifying}.

\subsection{Determining Network Width with CMI-Permutation~\cite{yu2019simple}}\label{appendix_B}

Our method to determine the network width (i.e., the number of filters $N$) is motivated by the concept of the Markov blanket (MB)~\cite{koller1996toward,yaramakala2005speculative}. Remember that the MB $M$ of a target variable $Y$ is the smallest subset of $S$ such that $Y$ is conditional independent of the rest of the variables $S-M$, i.e., $Y\perp(S-M)|M$~\cite{vergara2014review}. From the perspective of information theory, this indicates that the conditional mutual information (CMI) $\mathbf{I}(\{S-M\};Y|M)$ is zero. Therefore, given the selected filter subset $T_s$, the remaining filter subset $T_r$, and the class labels $Y$, we can obtain a reliable estimate to $N$ by evaluating if $\mathbf{I}(T_r;Y|T_s)$. Theoretically, $\mathbf{I}(T_r;Y|T_s)$ is non-negative and monotonically decreasing with the increase of the size of $T_s$ (i.e., $|T_s|$)~\cite{cover2012elements}. But it will never approach zero in practice due to statistical variation and chance agreement between variables~\cite{vinh2014reconsidering}.

To measure how close is $\mathbf{I}(T_r;Y|T_s)$ to zero, let us select a new candidate filter $t$ in $T_r$, we quantify how $t$ affects the MB condition by creating a random permutation of $t$ (without permuting the corresponding $Y$), denoted $\tilde{t}$. If $\mathbf{I}(\{T_r-t\};Y|\{T_s,t\})$ is not significantly smaller than $\mathbf{I}(\{T_r-\tilde{t}\};Y|\{T_s,\tilde{t}\})$, $t$ can be discarded from $T_s$ and the filter selection is stopped (i.e., $N=|T_s|$). We term this method CMI-permutation~\cite{yu2019simple}. Algorithm~\ref{PermutationAlg} gives its detailed implementation, in which $\mathbbm{1}$ denotes the indicator function.

\begin{algorithm}[!htbp]
\caption{CMI-permutation}
\small
\label{PermutationAlg}
\begin{algorithmic}[1]
\Require
Selected filter subset $T_s$;
Remaining filter subset $T_r$;
Class labels $Y$;
Selected filter $t$ (in $T_r$);
Permutation number $P$;
Significance level $\alpha$.
\Ensure
\emph{decision} (stop filter selection or continue filter selection);
Network width $N$.
\State Estimate $\mathbf{I}(\{T_r-t\};Y|\{T_s,t\})$ with matrix-based R{\'e}nyi's $\alpha$-entropy functional estimator~\cite{yu2018multivariate}.
\For {$i = 1$ to $P$}
\State Randomly permute $t$ to obtain $\tilde{t}_i$.
\State Estimate $\mathbf{I}(\{T_r-\tilde{t}_i\};Y|\{T_s,\tilde{t}_i\})$ with matrix-based R{\'e}nyi's $\alpha$-entropy functional estimator~\cite{yu2018multivariate}.
\EndFor
\If {$\frac{\sum\nolimits_{i=1}^P\mathbbm{1}[\mathbf{I}(\{T_r-t\};Y|\{T_s,t\})\geq\mathbf{I}(\{T_r-\tilde{t}_i\};Y|\{T_s,\tilde{t}_i\})]}{P}\leq\alpha$}
\State \emph{decision}$\leftarrow$Continue filter selection.
\Else
\State \emph{decision}$\leftarrow$Stop filter selection.
\State $N\leftarrow|T_s|$.
\EndIf\\
\Return \emph{decision}; $N$ (if stop filter selection)
\end{algorithmic}
\end{algorithm}

\subsection{Additional Experimental Results} \label{appendix_C}

\begin{figure}[!htbp]
\centering
\begin{tabular}{ccc}
\subfigure[MMI in Fashion-MNIST] {\includegraphics[width=.23\textwidth]{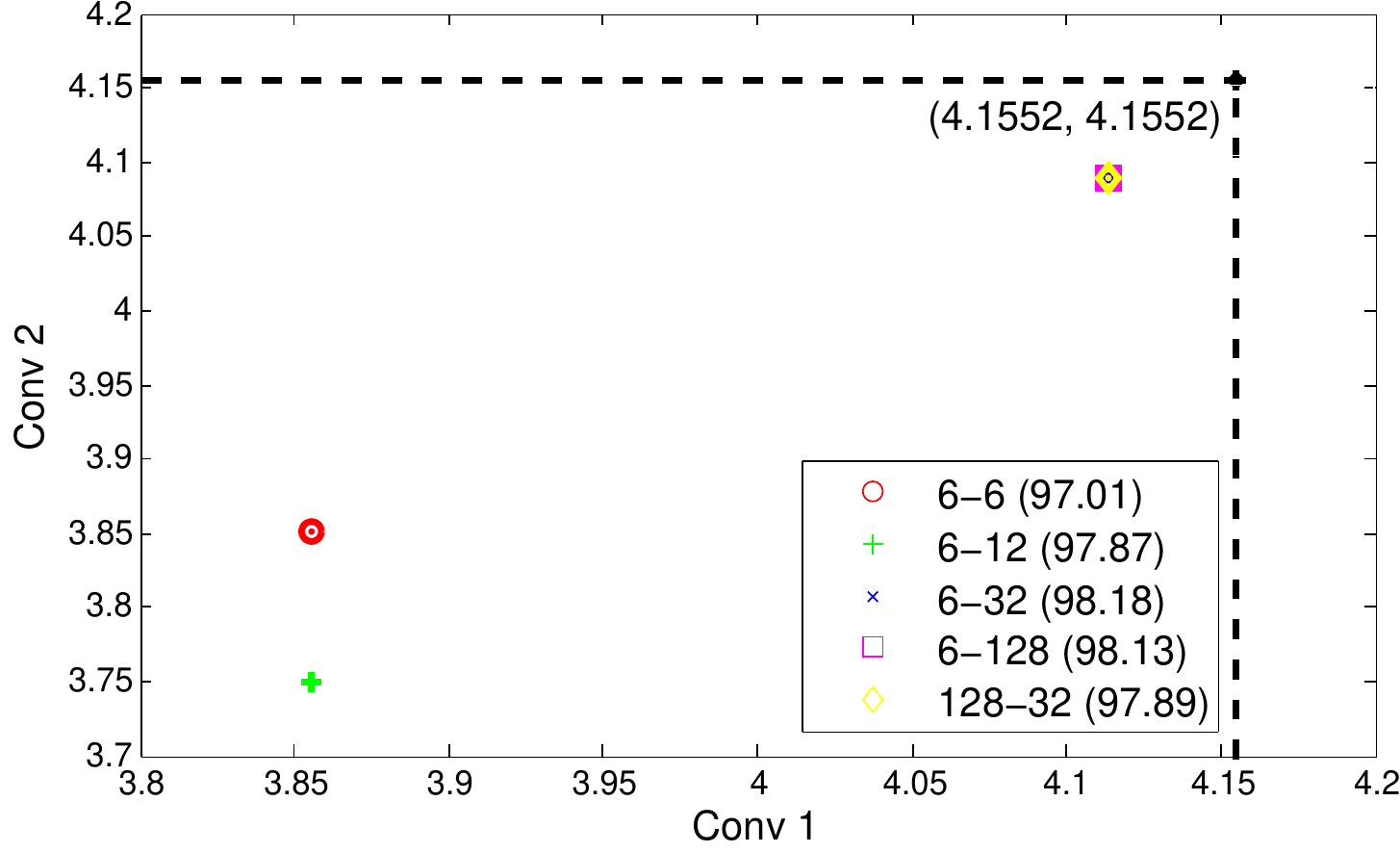}}
\subfigure[MMI in HASYv2] {\includegraphics[width=.23\textwidth]{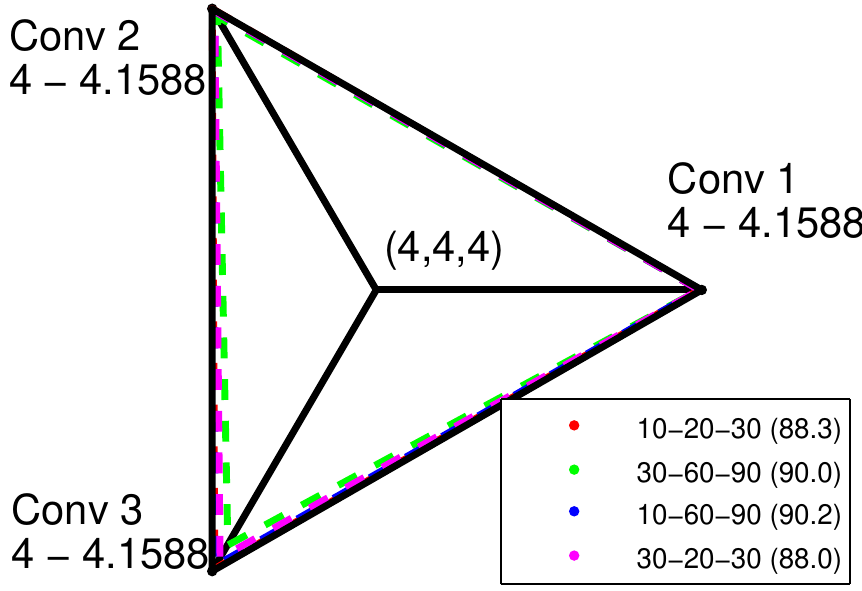}}
\end{tabular}
\caption{The MMI values in (a) Conv. $1$ and Conv. $2$ in Fashion-MNIST data set; and (b) Conv. $1$, Conv. $2$ and Conv. $3$ in Fruits~360 data set. The black line indicates the upper bound of MMI, i.e., the average mini-batch input entropy. The topologies of all competing networks are specified in the legend, in which the successive numbers indicate the number of filters in each convolutional layer. We also report their classification accuracies ($\%$) on testing set averaged over $10$ Monte-Carlo simulations in the parentheses.
\vspace{-0.0cm}}
\label{fig:MMI_saturation_supp}
\end{figure}

\begin{figure*}[!htbp]
\centering
\begin{tabular}{ccccc}
\subfigure[Redundancy-Synergy trade-off. The networks differ in the number of filters in Conv.~$1$.] {\includegraphics[width=.23\textwidth]{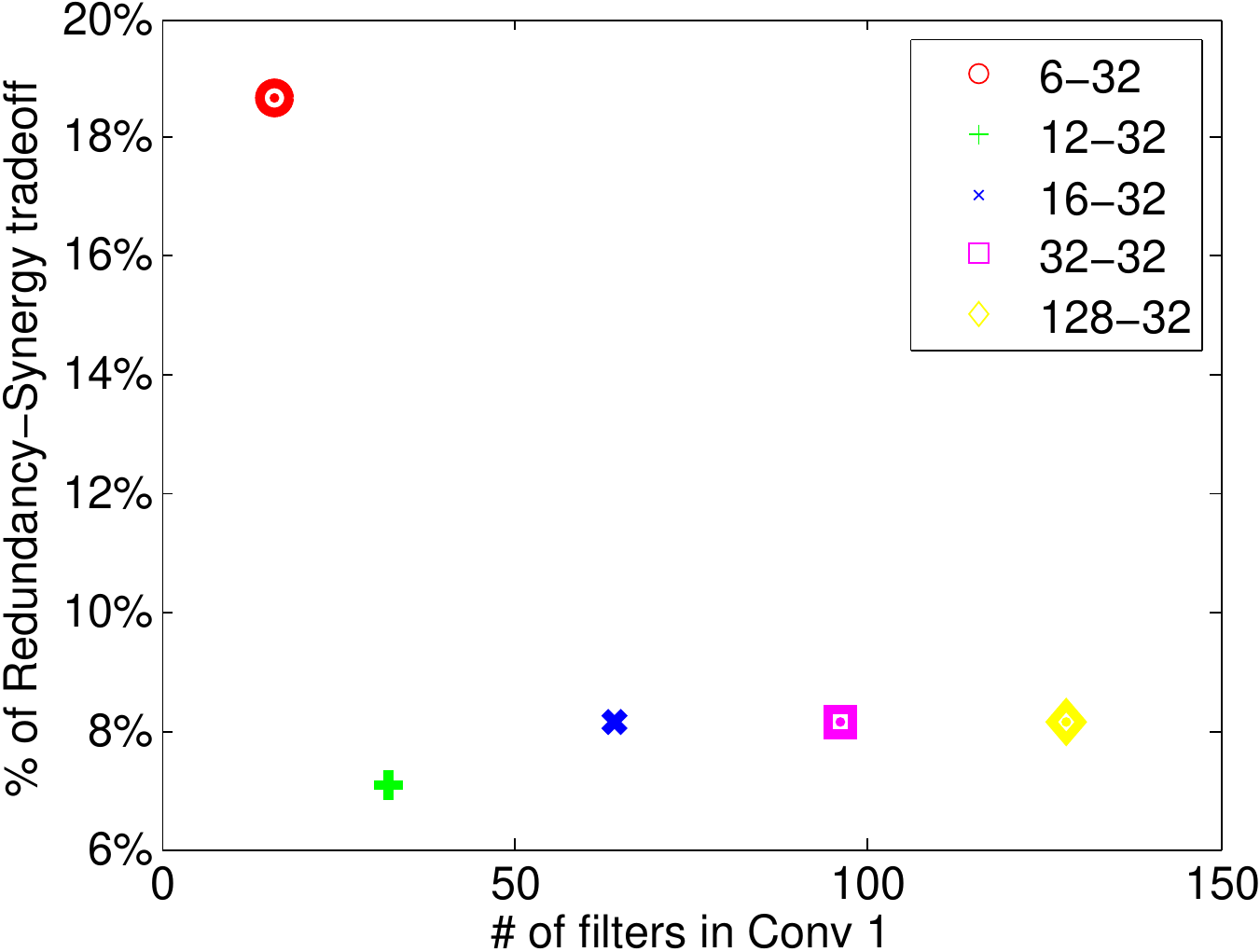}}
\subfigure[Weighted non-redundant information. The networks differ in the number of filters in Conv.~$1$.] {\includegraphics[width=.23\textwidth]{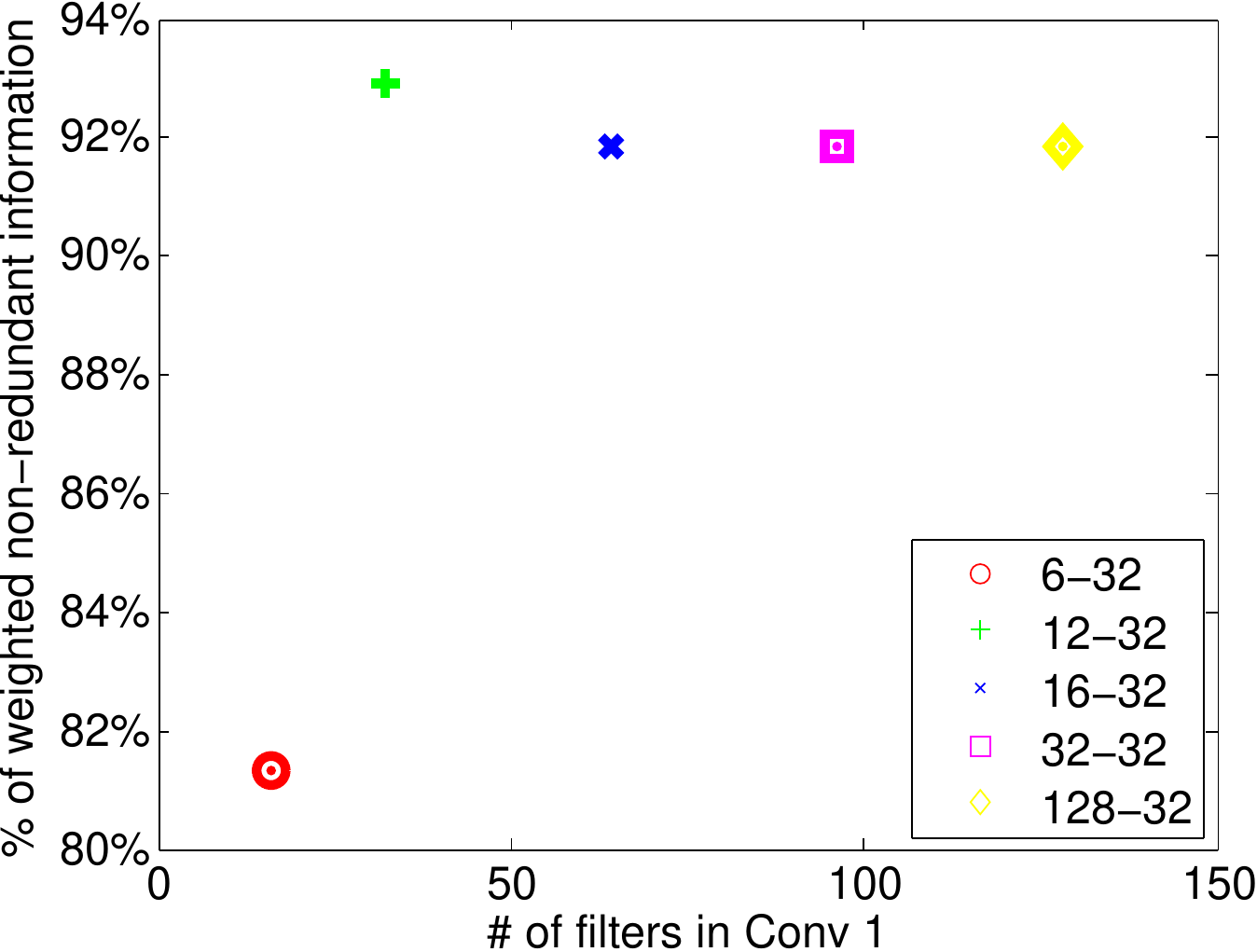}}
\subfigure[Redundancy-Synergy trade-off. The networks differ in the number of filters in Conv.~$2$.] {\includegraphics[width=.23\textwidth]{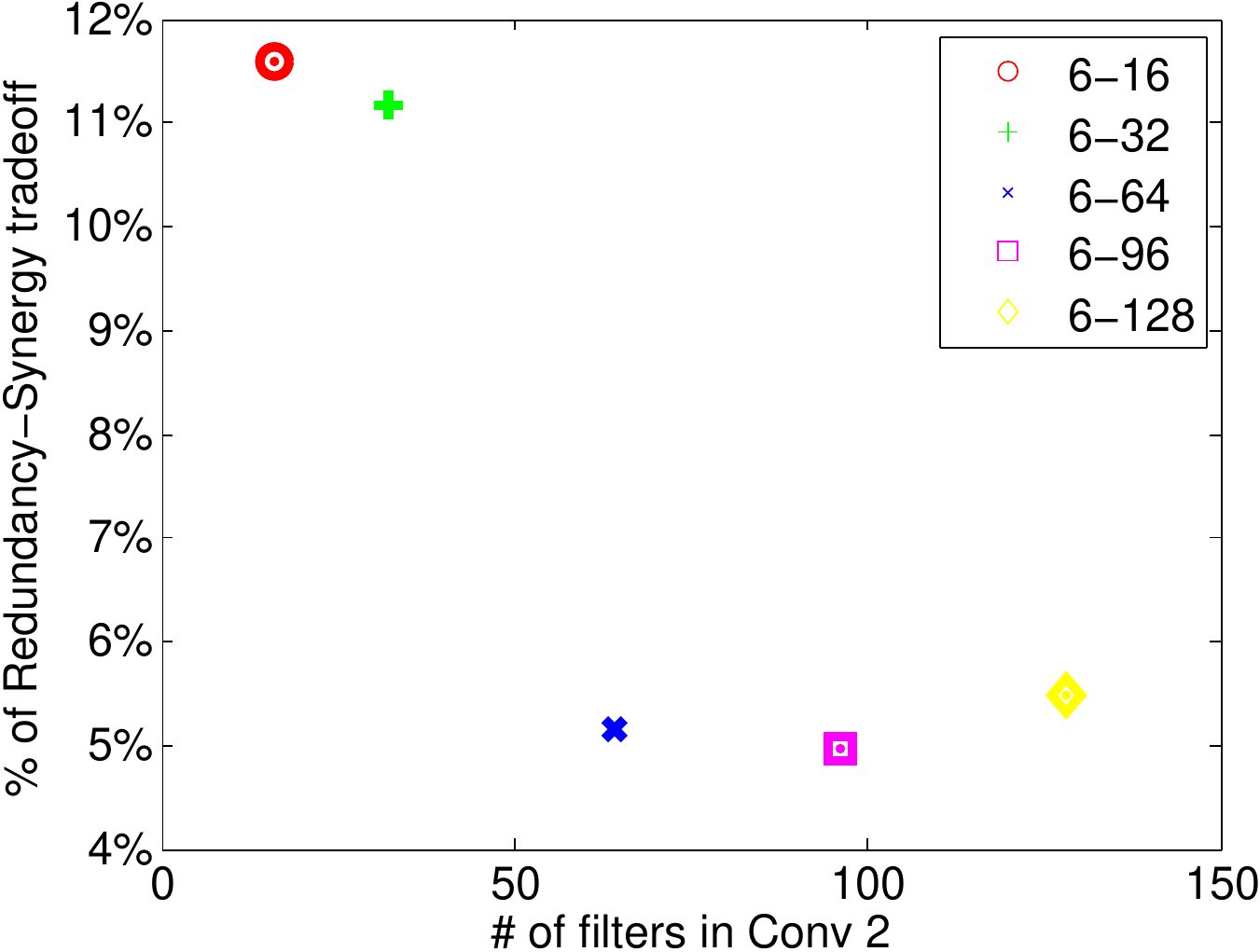}}
\subfigure[Weighted non-redundant information. The networks differ in the number of filters in Conv.~$2$.] {\includegraphics[width=.23\textwidth]{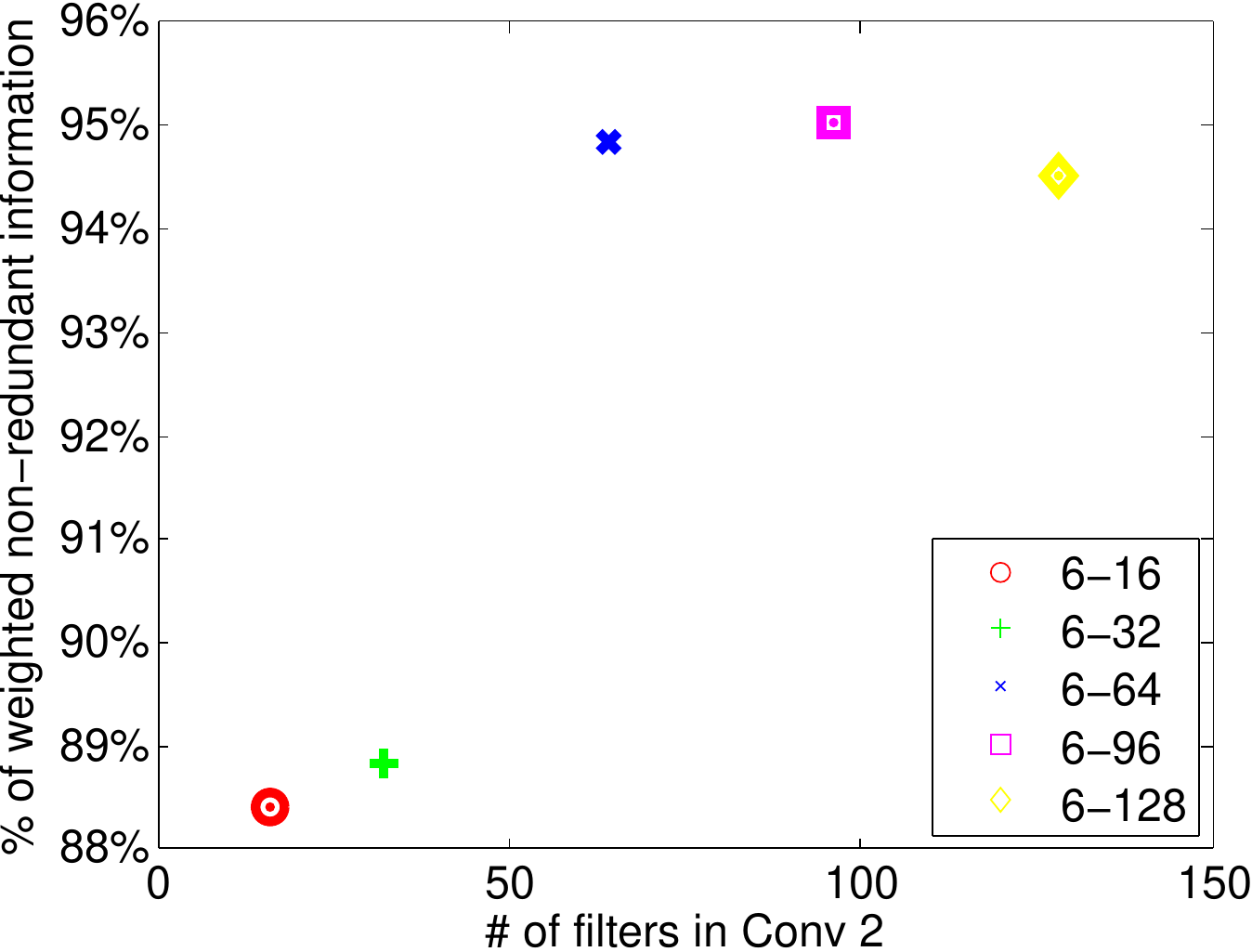}} & \\
\end{tabular}
\caption{The redundancy-synergy trade-off and the weighted non-redundant information in Fashion-MNIST data set. (a) and (b) demonstrate the percentages of these two quantities with respect to different number of filters in Conv.~$1$, but $32$ filters in Conv.~$2$. (c) and (d) demonstrate the percentages of these two quantities with respect to $6$ filters in Conv.~$1$, but different number of filters in Conv.~$2$. In each subfigure, the topologies of all competing networks are specified in the legend.\vspace{-0.0cm}}
\label{fig:tradeoff_supp}
\end{figure*}

\begin{figure*}[!htbp]
\centering
\begin{tabular}{ccc}
\subfigure[IP ($10$ filters in Conv.~1, $20$ filters in Conv.~2, $30$ filters in Conv.~3)] {\includegraphics[width=.23\textwidth]{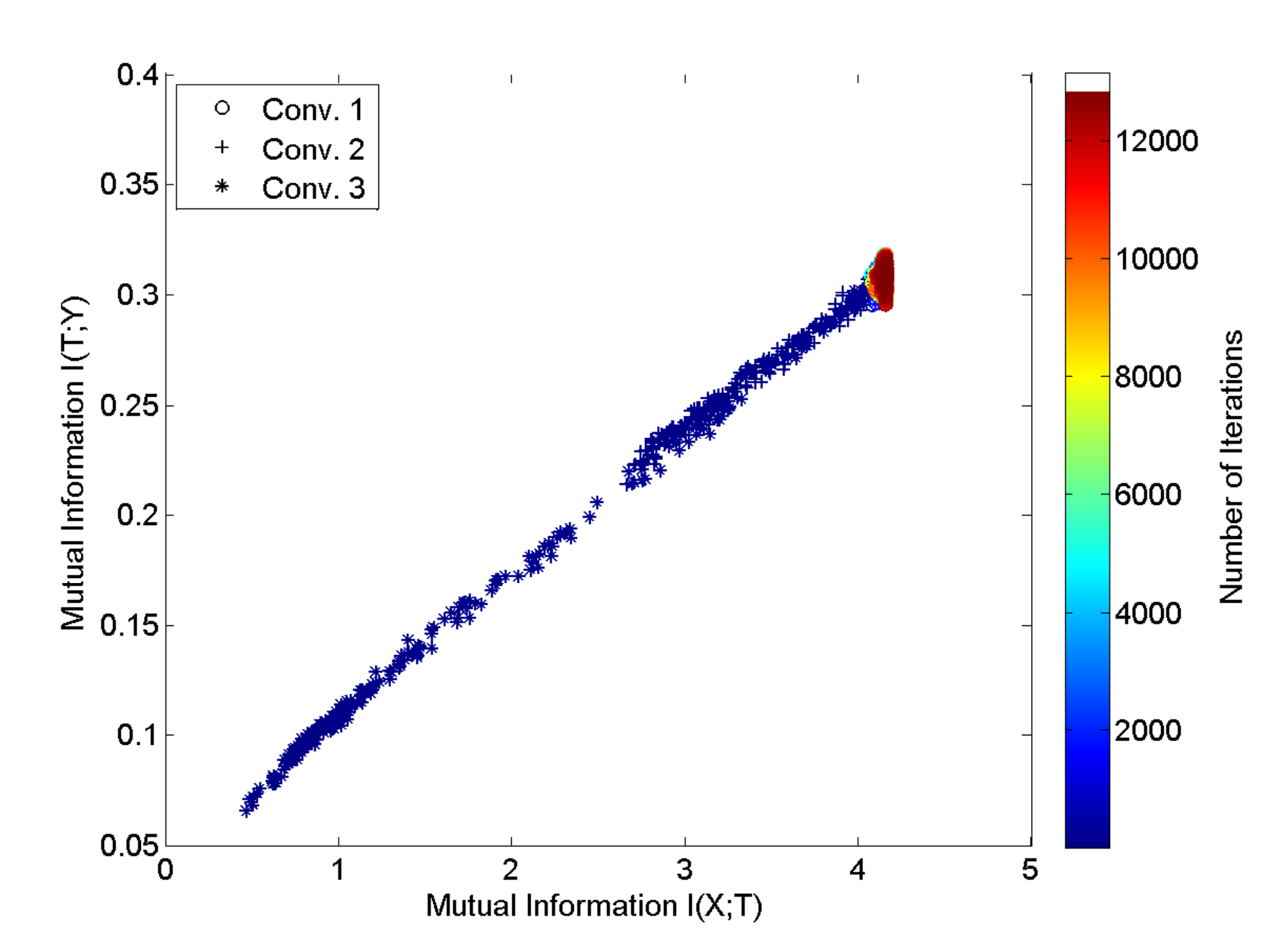}}
\subfigure[Conv.~2 in M-IP ($10$ filters in Conv.~1, $20$ filters in Conv.~2, $30$ filters in Conv.~3)] {\includegraphics[width=.23\textwidth]{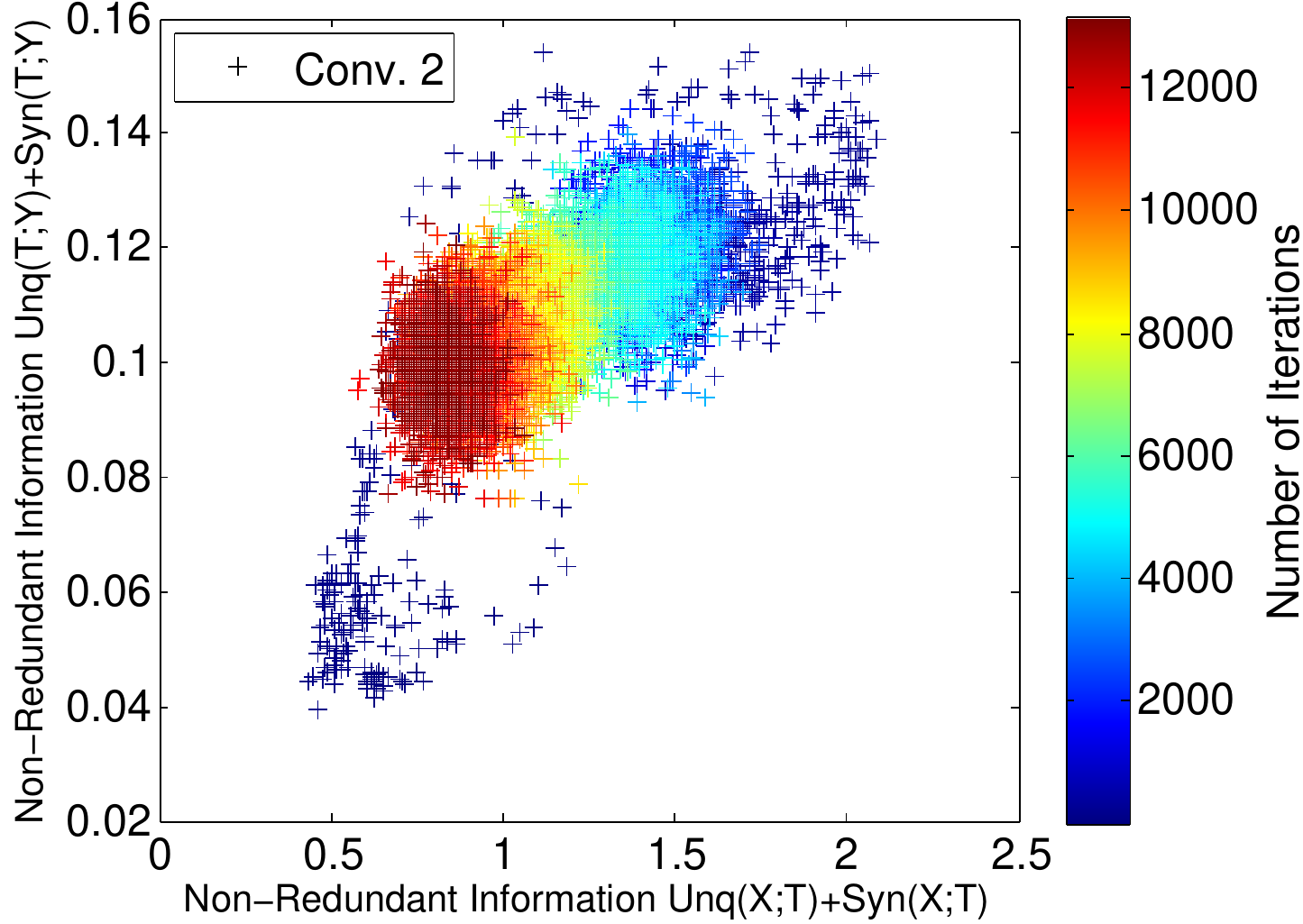}}
\subfigure[Conv.~3 in M-IP ($10$ filters in Conv.~1, $20$ filters in Conv.~2, $30$ filters in Conv.~3)] {\includegraphics[width=.23\textwidth]{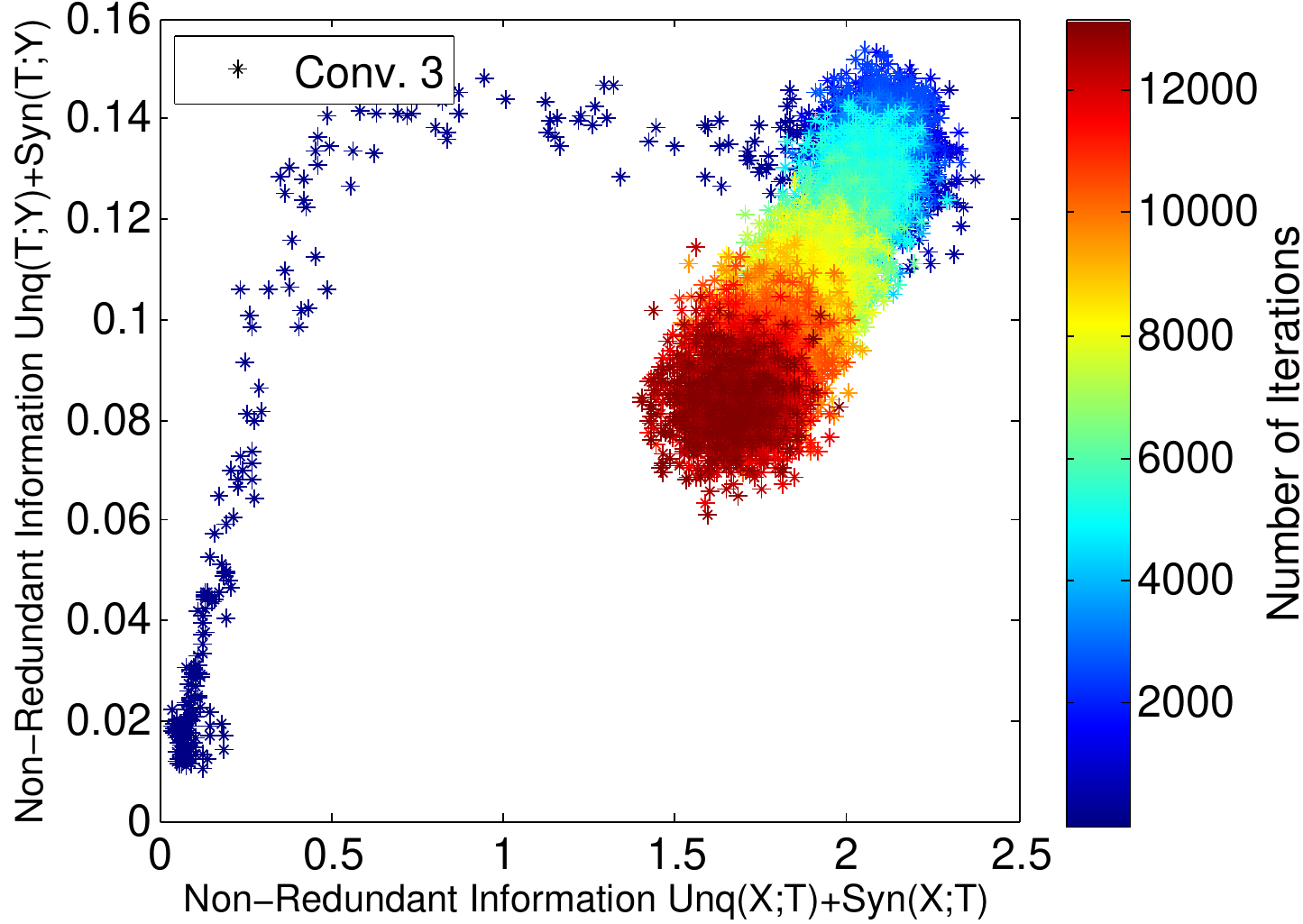}}\\
\subfigure[IP ($16$ filters in Conv.~1, $24$ filters in Conv.~2, $48$ filters in Conv.~3)] {\includegraphics[width=.23\textwidth]{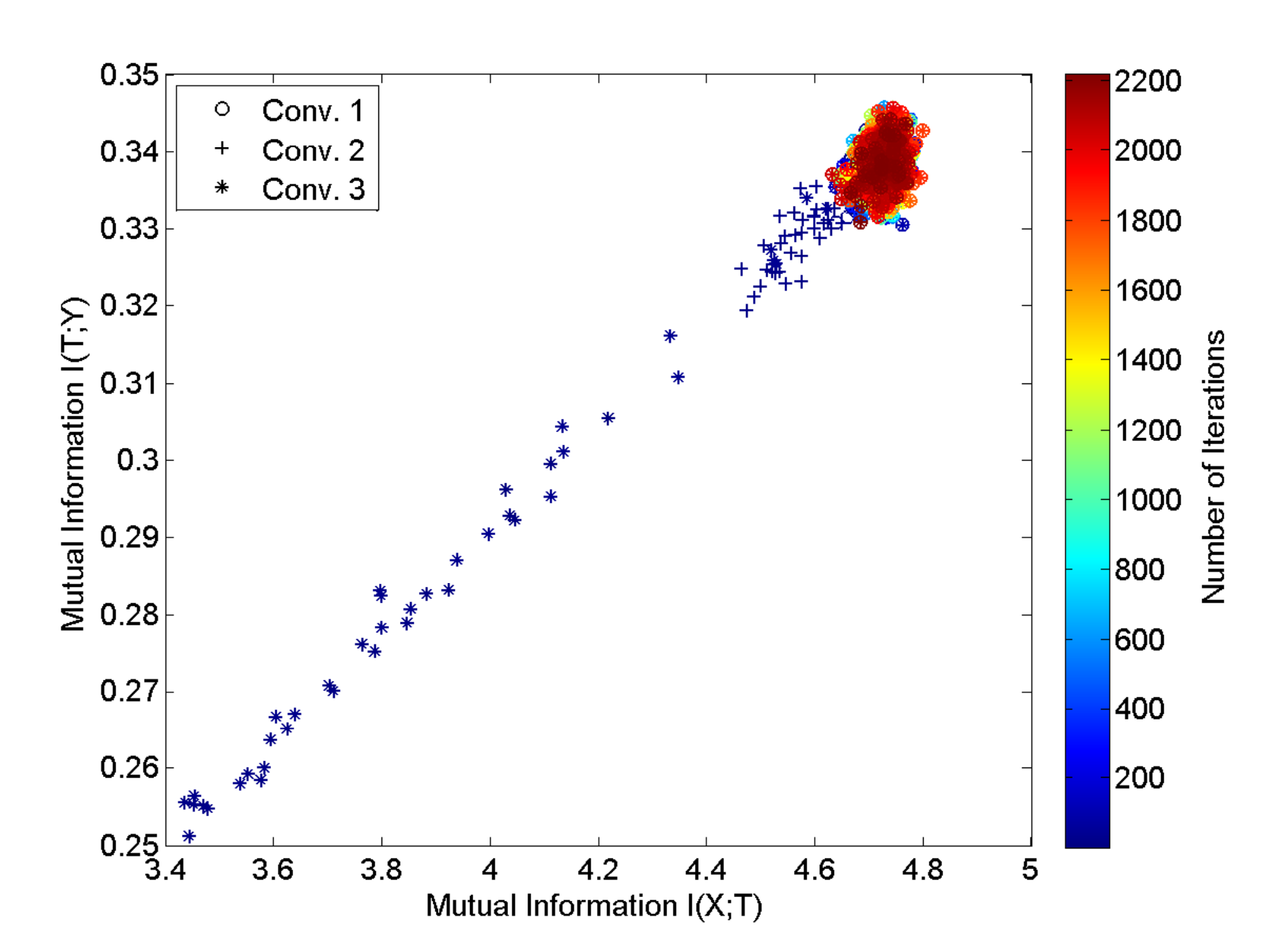}}
\subfigure[Conv.~2 in M-IP ($16$ filters in Conv.~1, $24$ filters in Conv.~2, $48$ filters in Conv.~3)] {\includegraphics[width=.23\textwidth]{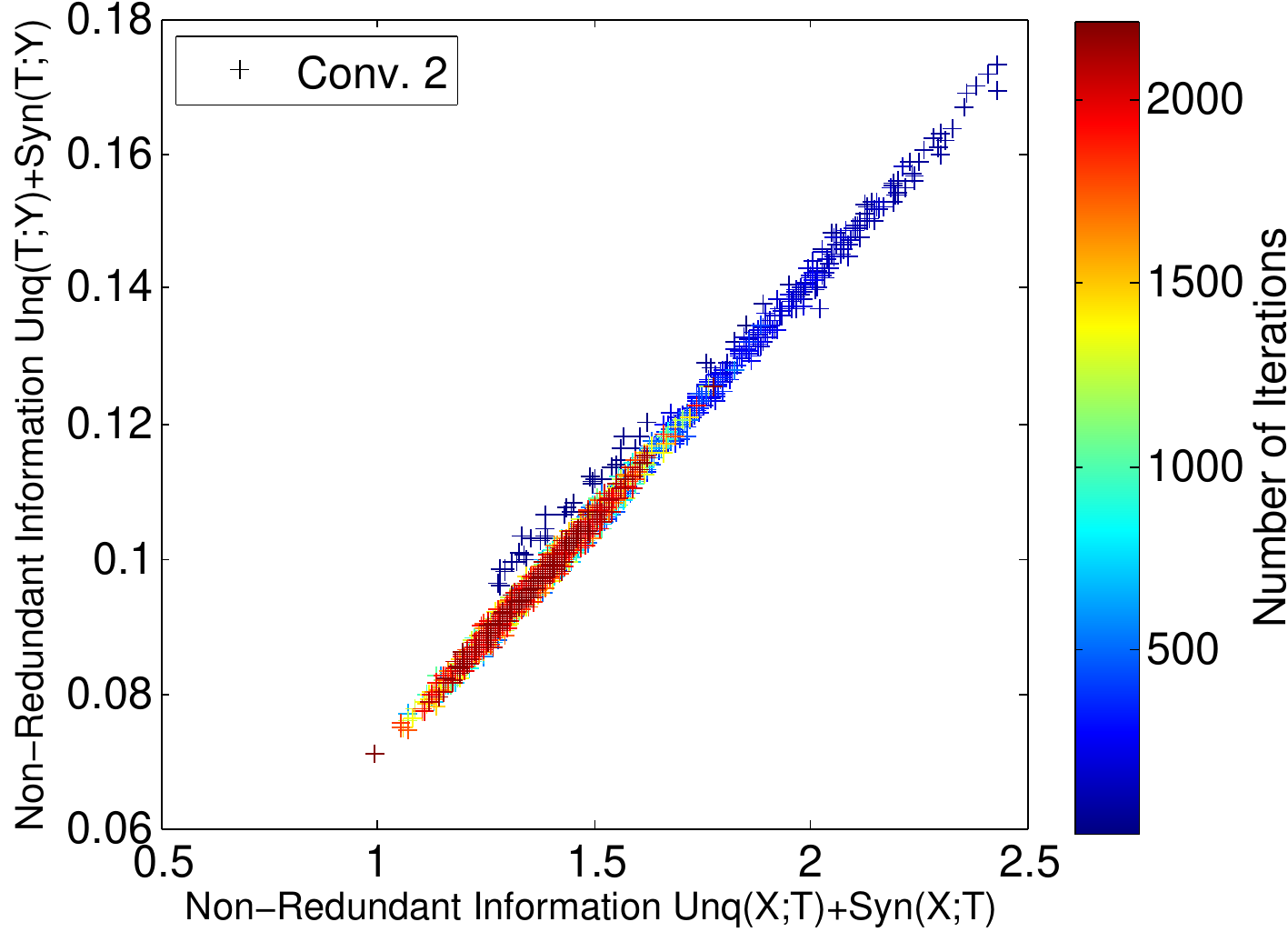}}
\subfigure[Conv.~3 in M-IP ($16$ filters in Conv.~1, $24$ filters in Conv.~2, $48$ filters in Conv.~3)] {\includegraphics[width=.23\textwidth]{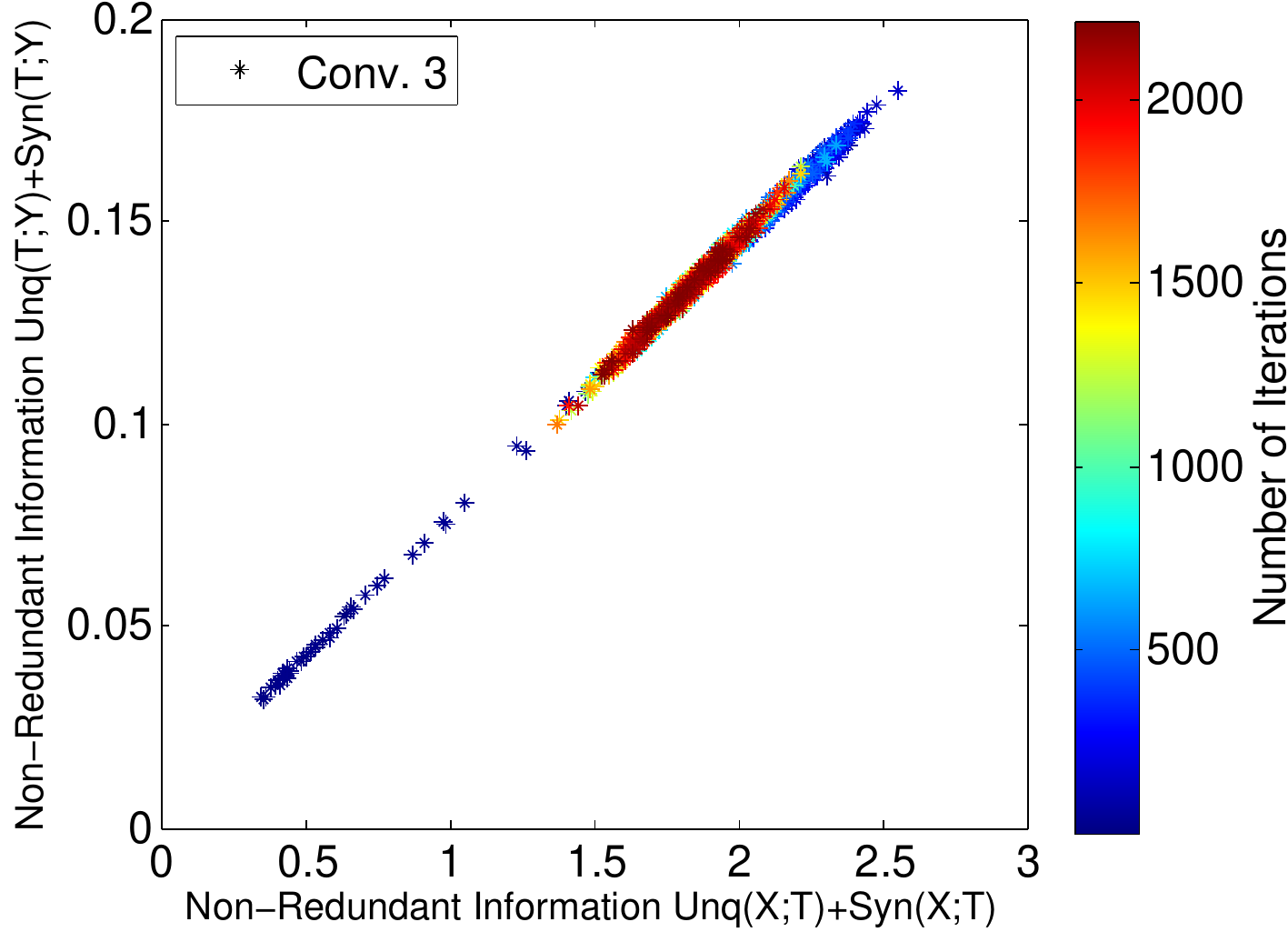}}\\
\end{tabular}
\caption{The Information Plane (IP) and the modified Information Plane (M-IP) of a smaller AlexNet type CNN trained on HASYv2 (the first row) and Fruits~$360$ (the second row) data sets. The $\#$ of filters in Conv.~$1$, the $\#$ of filters in Conv.~$2$, and the $\#$ of filters in Conv.~$3$ are indicated in the subtitle of each plot. The curves in IP increase rapidly up to a point without compression (see (a) and (d)). By contrast, it is very easy to observe the compression in M-IP (see (b), (c), (e) and (f)).\vspace{-0.0cm}}
\label{fig:IP_MNIST_supp}
\end{figure*}

We finally represent additional results to further support our arguments in the main text. Specifically, Fig.~\ref{fig:MMI_saturation_supp} evaluates multivariate mutual information (MMI) values with respect to different CNN topologies trained on Fashion-MNIST and HASYv2 data sets. Obviously, MMI values are likely to saturate with only a few filters. Moreover, increasing the number of filters does not guarantee that classification accuracy increases, and might even degrade performance.
The tendency of redundancy-synergy tradeoff and weighted non-redundant information are shown in Fig.~\ref{fig:tradeoff_supp}. Again, more filters can push the network towards an improved redundancy-synergy trade-off, i.e., the synergy gradually dominates in each pair of feature maps with the increase of number of filters.
Fig.~\ref{fig:IP_MNIST_supp} demonstrates the information plane (IP) and the modified information plane (MIP) for a smaller AlexNet~\cite{krizhevsky2012imagenet} type CNN trained on HASYv2 and Fruits~360 data sets. Although we did not observe any compression in IP, it is very easy to observe the compression phase in the MIP.



\end{document}